\documentclass[11 pt]{article}
\usepackage{style}
\title{\Large\bfseries From Set Convergence to Pointwise Convergence: Finite-Time Guarantees for Average-Reward Q-Learning with Adaptive Stepsizes}
\author{Zaiwei Chen$^*$ and Phalguni Nanda$^\dagger$\\
{\small
\textit{Edwardson School of Industrial Engineering, Purdue University}}\\ {\small$^*$\href{mailto:chen5252@purdue.edu}{\textit{chen5252@purdue.edu}}, $^\dagger$\href{mailto:nanda14@purdue.edu}{\textit{nanda14@purdue.edu}}}
}
\date{\vspace{-0.4 in}}
\begin{document}
\maketitle

\begin{abstract}
This work presents the first finite-time analysis for the last-iterate convergence of average-reward $Q$-learning with an asynchronous implementation. A key feature of the algorithm we study is the use of adaptive stepsizes, which serve as local clocks for each state-action pair. We show that, under appropriate assumptions, the iterates generated by this $Q$-learning algorithm converge at a rate of $\tilde{\mathcal{O}}(1/k)$ (in the mean-square sense) to the optimal  $Q$-function in the span seminorm. Moreover, by adding a centering step to the algorithm, we further establish pointwise mean-square convergence to the centered optimal  $Q$-function, also at a rate of $\tilde{\mathcal{O}}(1/k)$. To prove these results, we show that adaptive stepsizes are necessary, as without them, the algorithm fails to converge to the correct target. In addition, adaptive stepsizes can be interpreted as a form of implicit importance sampling that counteracts the effects of asynchronous updates. Technically, the use of adaptive stepsizes makes each $Q$-learning update depend on the entire sample history, introducing strong correlations and making the algorithm a non-Markovian stochastic approximation (SA) scheme. Our approach to overcoming this challenge involves (1) a time-inhomogeneous Markovian reformulation of non-Markovian SA, and (2) a combination of almost-sure time-varying bounds, conditioning arguments, and Markov chain concentration inequalities to break the strong correlations between the adaptive stepsizes and the iterates. The tools developed in this work are likely to be broadly applicable to the analysis of general SA algorithms with adaptive stepsizes.
\end{abstract}

\section{Introduction}\label{sec:intro}
Reinforcement Learning (RL) has become as a powerful framework for solving sequential decision-making problems, as demonstrated by its growing impact across a range of real-world applications, including autonomous robotics \cite{singh2022reinforcement}, game-playing AI \cite{silver2018general}, and the development of large language models \cite{brown2020language}. Given the promising potential of RL, establishing strong theoretical foundations to guide its practical implementation is of significant importance.

An RL problem is usually modeled as a Markov decision process (MDP) \cite{sutton2018reinforcement}, but its objective function can vary depending on the application of interest. Most recent theoretical results in RL focus on either the finite-horizon or infinite-horizon discounted-reward settings, which may not apply to real-world scenarios that require continual, long-horizon decision-making. As a concrete example, in queueing and network scheduling applications, the system operates continuously without a natural episode end, and the quality of decisions is evaluated through steady-state metrics such as long-run delay, queue length, or throughput \cite{harchol2013performance,liu2022rl,dai2022queueing}. In such settings, transient effects vanish over time, and the natural goal is to optimize the time-average performance. The infinite-horizon average-reward MDP formulation captures this steady-state objective \cite{mahadevan1996average}, making it particularly well-suited for continual learning applications such as manufacturing, inventory control, and queueing and networking, among others. However, the absence of discounting introduces unique challenges for both algorithm design and analysis. For example, the associated Bellman operator is no longer a norm-contractive mapping \cite{puterman2014markov}, and sample-based updates introduce additional complexities into the algorithmic structure \cite{abounadi2001learning}.

These challenges are particularly evident in $Q$-learning \cite{watkins1992q}, one of the most classical and practically impactful RL algorithms. Due to its popularity and its role as a major milestone in RL \cite{mnih2015human}, substantial efforts have been dedicated to providing theoretical guarantees, especially in terms of convergence rates. In the discounted setting, the first finite-time analysis of $Q$-learning was conducted in the early 2000s \cite{even2003learning}, followed by a series of works over the past two decades that eventually led to almost matching upper and lower bounds \cite{azar2012dynamic,li2024q}. In contrast, in the average-reward setting, due to the aforementioned challenges, existing results are largely limited to asymptotic convergence \cite{abounadi2001learning,wan2024convergence,kara2023q,yang2024relative,yu2024asynchronous,suttle2021reinforcement,wan2021learning} and regret analysis \cite{jaksch2010near,agrawal2024optimistic,wei2020model,zhang2023sharper}. Regarding finite-time analysis, even for $Q$-learning with synchronous updates (which requires a generative model for i.i.d. sampling), results have only appeared very recently \cite{zhang2021finite,bravo2024stochastic,jin2024feasible}. See Section \ref{subsec:literature} for a more detailed literature review. To the best of our knowledge, no existing work provides a finite-time analysis for the last-iterate convergence of $Q$-learning with asynchronous updates based on a single trajectory of Markovian samples.

\subsection{Main Contributions}

In this work, we provide the first principled study on the finite-time analysis of average-reward $Q$-learning. Specifically, we make the following contributions.

\paragraph{Finite-Time Analysis for the Last-Iterate Convergence.} We study two average-reward $Q$-learning algorithms. The first one (cf.\ Algorithm~\ref{alg:QL-set}) represents the most natural extension of $Q$-learning from the discounted setting, where the only modification, besides setting the discount factor to one, is the use of stepsizes $\alpha_k(s,a)$ that adapt to individual state–action pairs $(s,a)$. In particular, the adaptive stepsize $\alpha_k(s,a)$ is inversely proportional to the number of visits to the state–action pair $(s,a)$, thereby serving as a local clock for each pair. We establish finite-time convergence bounds for this $Q$-learning algorithm, showing that $\mathbb{E}[\mysp(Q_k - Q^*)^2] \leq \tilde{\mathcal{O}}(1/k)$, where $\mysp(\cdot)$ denotes the span seminorm, $Q_k$ is the $k$-th iterate, and $Q^*$ is the optimal  $Q$-function (cf.\ Theorem~\ref{thm:QL-set}). Alternatively, this result can be interpreted as convergence to the set $\mathcal{Q} = \{Q^* + c e \mid c \in \mathbb{R}\}$, where $e$ denotes the all-ones vector, with a rate of $\tilde{\mathcal{O}}(1/k)$ in mean-square distance. Building upon Algorithm~\ref{alg:QL-set}, we further propose that by adding an additional centering step, the resulting algorithm (cf.\ Algorithm~\ref{alg:QL-pointwise}) achieves pointwise convergence to the centered optimal  $Q$-function, denoted by $\tilde{Q}^*$, also with a rate of $\tilde{\mathcal{O}}(1/k)$; that is, $\mathbb{E}[\|Q_k - \tilde{Q}^*\|_\infty^2] \leq \tilde{\mathcal{O}}(1/k)$ (cf.\ Theorem~\ref{thm:QL-pointwise}).

\paragraph{The Necessity of Using Adaptive Stepsizes.} Since the use of adaptive stepsizes is the only major difference compared with the existing finite-time analyses of discounted $Q$-learning,\footnote{Although adaptive stepsizes have been used in the existing asymptotic analyses of discounted $Q$-learning \cite{tsitsiklis1994asynchronous,abounadi2001learning}, they are not necessary in that setting, which stands in stark contrast to average-reward $Q$-learning. Consequently, they have been less explored in the context of finite-time analysis.} we conduct a thorough investigation into both the necessity and the underlying reasons for using adaptive stepsizes in average-reward $Q$-learning. Specifically, we show that if universal stepsizes are used (e.g., $\alpha_k(s,a) = 1/k$ for all $(s,a)$), the algorithm is, in general, guaranteed \emph{not} to converge to any point in the desired set $\mathcal{Q} = \{Q^* + c e \mid c \in \mathbb{R}\}$ (cf.\ Proposition~\ref{prop:universal-convergence}). We then identify that the adaptive stepsizes can be interpreted as a form of importance sampling that counteracts the effect of asynchronous updates based on a single sample trajectory, where the importance sampling ratios are estimated using empirical frequencies derived from the historical number of visits to each state–action pair. Building on this insight, we further develop variants of $Q$-learning with alternative forms of adaptive stepsizes and numerically demonstrate that the resulting algorithms achieve the expected performance.

\paragraph{Technical Contributions.} The use of adaptive stepsizes in $Q$-learning causes each update to depend on the entire history of visited state–action pairs, introducing strong correlations and making the algorithm a non-Markovian stochastic approximation (SA). This poses significant challenges for the finite-time analysis, which have not been addressed in the existing literature. To overcome this challenge, we first reformulate the algorithm as a Markovian SA by incorporating the empirical frequencies of visited state–action pairs into the stochastic process that drives the algorithm. However, the resulting Markovian noise is time-inhomogeneous and does not exhibit the desired geometric mixing property typically used to handle correlations between the iterates and the noise. To resolve this issue, we further develop an approach that begins by showing that the $Q$-learning iterates, while not uniformly bounded by a constant—a fundamental challenge compared with the discounted setting \cite{gosavi2006boundedness}—nonetheless satisfy an almost-sure time-varying bound that grows at most logarithmically with $k$ (cf.\ Proposition~\ref{prop:log-growth}). We then leverage this result, along with conditioning arguments and Markov chain concentration inequalities, to break the correlation. The technical tools developed in this work are likely to be of broad interest for studying general SA algorithms with adaptive stepsizes.

\subsection{Related Literature}\label{subsec:literature}
In this section, we discuss related literature on average-reward $Q$-learning, discounted $Q$-learning, and Markovian SA.

\paragraph{Average-Reward Q-Learning.} The first provably convergent algorithms for average-reward $Q$-learning are the relative value iteration (RVI) $Q$-learning and stochastic shortest path (SSP) $Q$-learning algorithms \cite{abounadi2001learning,abounadi2002stochastic}. Since then, many variants have been proposed, including differential $Q$-learning \cite{wan2021learning,wan2021average}, dynamic horizon $Q$-learning \cite{jin2024feasible}, among others. The asymptotic convergence of RVI $Q$-learning and its variants was established in \cite{abounadi2001learning} by leveraging results from general asynchronous SA \cite{borkar1998asynchronous}, and was later extended under relaxed assumptions \cite{abounadi2002stochastic,wan2024convergence,yu2024asynchronous} and to MDPs with continuous state spaces \cite{yang2024relative,kara2023q}. In contrast, non-asymptotic results are much more limited. Specifically, for $Q$-learning with a synchronous implementation (which requires a generative model for i.i.d.\ sampling), finite-time analysis has been conducted in \cite{zhang2021finite,chen2025non,zhang2021finite} by modeling the algorithm as a seminorm SA with martingale difference noise, and in \cite{jin2024feasible} by using discounted $Q$-learning with a dynamically increasing horizon as a gradual approximation to average-reward $Q$-learning.
A related but distinct line of work designs online $Q$-learning-based algorithms that aim to balance the exploration–exploitation trade-off, with performance measured in terms of regret; see \cite{agrawal2024optimistic,zhang2023sharper,wei2020model} and the references therein. However, since the performance metrics are fundamentally different (regret versus last-iterate convergence in norms or seminorms), these results are not directly comparable. 

\paragraph{Discounted Q-Learning.} The celebrated $Q$-learning algorithm for discounted MDPs was first proposed in \cite{watkins1992q}, and its almost sure convergence was established through various approaches in \cite{tsitsiklis1994asynchronous,jaakkola1994convergence,borkar2000ode}. The first finite-time analysis of $Q$-learning (for both synchronous and asynchronous implementations) was performed in \cite{even2003learning}. Since then, a sequence of works has aimed to provide refined characterizations of its convergence rate, including mean-square bounds \cite{beck2012error,beck2013improved,chen2020finite,chen2024lyapunov,wainwright2019stochastic,lee2024final} and high-probability concentration bounds \cite{qu2020finite,li2020sample,wainwright2019variance,li2021tightening,li2024q}. Variants of $Q$-learning, including Zap $Q$-learning, $Q$-learning with Polyak averaging, and $Q$-learning with Richardson - Romberg extrapolation, have been proposed and studied in \cite{devraj2017zap}, \cite{li2023statistical}, and \cite{zhang2024constant}, respectively. To overcome the curse of dimensionality, $Q$-learning is often implemented with function approximation in practice. There is a rich body of literature dedicated to providing theoretical guarantees for $Q$-learning with function approximation \cite{chen2019finitesample,chen2023target,melo2008analysis,zhang2021breaking,meyn2024projected,meyn2024projected}. 

Compared with $Q$-learning in the discounted setting, the finite-time analysis of average-reward $Q$-learning is significantly more challenging due to the following reasons:
(1) the lack of discounting makes the Bellman operator non-contractive in any norm,
(2) the iterates are not uniformly bounded, and
(3) the use of adaptive stepsizes (which is necessary, as will be illustrated in Section \ref{sec:viewpoint}) introduces strong correlations and makes the algorithm a non-Markovian SA. As a result, none of the existing approaches for discounted $Q$-learning are applicable here.

\paragraph{Stochastic Approximation.} 
At a high level, $Q$-learning can be viewed as iteratively solving the Bellman equation, a fixed-point equation, using the SA method \cite{robbins1951stochastic}. The asymptotic convergence of SA for solving fixed-point equations has been established under fairly general assumptions in \cite{borkar2000ode,lauand2024revisiting,benveniste2012adaptive,borkar_SA,kushner2012stochastic}, among many others.  For finite-time analysis, the properties of the fixed-point operator and the nature of the noise sequence play crucial roles. In particular, when the operator is linear or contractive with respect to some norm, and the noise process is either i.i.d., or a martingale difference sequence, or forms a uniformly ergodic Markov chain, there is a rich body of literature establishing mean-square and high-probability bounds \cite{srikant2019finite,bhandari2018finite,qu2020finite,chen2024lyapunov,wainwright2019stochastic,mou2021optimal,mou2020linear,chen2025concentration,durmus2021tight,qian2024almost,huo2023bias}. Beyond these standard settings, finite-time analysis of SA with seminorm contractive operators and non-expansive operators have been developed recently in \cite{chen2025non} and \cite{blaser2024asymptotic,bravo2024stochastic}, respectively. 

A main feature of the average-reward $Q$-learning algorithms studied in this work is that, as SA algorithms, they are inherently non-Markovian due to the use of adaptive stepsizes that depend on the entire history of visited state–action pairs. Through a novel reformulation, we cast the algorithm as a Markovian SA; however, the resulting Markovian noise is time-inhomogeneous and thus does not exhibit geometric mixing. This presents a unique challenge that has not been addressed in the existing finite-time analyses of SA.

\section{From Average-Reward RL to the Seminorm Bellman Equation}\label{sec:preliminaries}
In this section, we first provide background on average-reward RL and then introduce the seminorm Bellman equation, which plays a central role in motivating both the algorithm design and the analysis of $Q$-learning.

\subsection{Background on Average-Reward RL}\label{subsec:background}
Consider an infinite-horizon, average-reward MDP defined by the tuple $(\mathcal{S}, \mathcal{A}, \mathcal{P}, \mathcal{R})$ \cite{puterman2014markov}, where $\mathcal{S}$ and $\mathcal{A}$ denote the finite state and action spaces, respectively. The transition probabilities are given by $\mathcal{P} = \{p(s' | s, a)\}_{(s,a,s') \in \mathcal{S} \times \mathcal{A} \times \mathcal{S}}$, where $p(s' | s, a)$ denotes the probability of transitioning to state $s'$ after taking action $a$ in state $s$. The stage-wise reward function is denoted by $\mathcal{R}: \mathcal{S} \times \mathcal{A} \to [-1,1]$. The transition probabilities and the reward function are unknown to the agent, but the agent can interact with the environment by taking actions, observing transitions, and receiving rewards.

Given a policy $\pi: \mathcal{S} \to \Delta(\mathcal{A})$ (where $\Delta(\mathcal{A})$ denotes the set of probability distributions supported on $\mathcal{A}$), the average reward $r^\pi\in\mathbb{R}^{|\mathcal{S}|}$ is defined as $r^\pi(s)=\lim_{K\to\infty}\mathbb{E}_\pi[\frac{1}{K}\sum_{k=1}^{K}\mathcal{R}(S_k,A_k)\mid S_1=s]$ for all $s\in\mathcal{S}$,
where the expectation $\mathbb{E}_\pi[\ \cdot \ ]$ is taken over the randomness in the trajectory generated by the policy $\pi$. It is shown in standard MDP theory \cite{puterman2014markov} that if the Markov chain induced by the policy $\pi$ has a single recurrent class, the limit exists and is independent of the initial state. In this work, we operate under this setting. Consequently, we slightly abuse notation and use $r^\pi$ to denote the scalar average reward associated with policy $\pi$. The goal is to find an optimal policy $\pi^*$ that maximizes the average reward.

Define the action-value function, also known as the $Q$-function, $Q^*\in\mathbb{R}^{|\mathcal{S}||\mathcal{A}|}$ as $Q^*(s,a) = \mathbb{E}_{\pi^*} [\sum_{k=1}^\infty (\mathcal{R}(S_k, A_k) - r^*) \mid S_1 = s, A_1 = a ]$ for all $(s,a)\in\mathcal{S}\times \mathcal{A}$,
where $r^*\in\mathbb{R}$ is the optimal average reward. It is known that any policy $\pi$ that satisfies $\pi(s) \in \argmax_{a \in \mathcal{A}} Q^*(s,a)$ for all state $s\in\mathcal{S}$ is an optimal policy \cite{puterman2014markov}. Therefore, the problem reduces to finding $Q^*$, which leads to the Bellman equation. Let $\mathcal{H}: \mathbb{R}^{|\mathcal{S}||\mathcal{A}|} \to \mathbb{R}^{|\mathcal{S}||\mathcal{A}|}$ denote the Bellman operator defined as $[\mathcal{H}(Q)](s,a) = \mathcal{R}(s,a) + \mathbb{E} [ \max_{a' \in \mathcal{A}} Q(S_2, a') \mid  S_1 = s, A_1 = a ]$
for all $(s,a)$ and $Q \in \mathbb{R}^{|\mathcal{S}||\mathcal{A}|}$. Then, the Bellman equation is given by
\begin{align}\label{eq:Bellman}
    \mathcal{H}(Q) - Q = r^* e.
\end{align}
It is well known that $Q^*$ is a solution to Equation \eqref{eq:Bellman} \cite{puterman2014markov}. However, such a solution is not unique. To see this, observe that for any $c \in \mathbb{R}$, it follows directly from the definition of $\mathcal{H}(\cdot)$ that $\mathcal{H}(Q^* + c e)-(Q^* + c e)=\mathcal{H}(Q^*)+ce-Q^*-ce=\mathcal{H}(Q^*)-Q^*=r^*e$,
implying that $Q^*+ce$ is also a solution to Equation \eqref{eq:Bellman}. Fortunately, for the purpose of finding an optimal policy, errors in the direction of the all-ones vector have no impact on the result, because $\pi^*(s) \in \argmax_{a \in \mathcal{A}} Q^*(s,a) = \argmax_{a \in \mathcal{A}} (Q^*(s,a) + c)$ for all $c\in\mathbb{R}$.
Therefore, it suffices to find any point in the set $\mathcal{Q} := \{ Q^* + c e \mid c \in \mathbb{R} \}$.

\subsection{The Seminorm Bellman Equation}\label{subsec:span-seminorm}
Due to the lack of discounting, the Bellman operator $\mathcal{H}(\cdot)$ is not a contraction mapping under any norm. However, it has been shown in \cite{puterman2014markov} that the operator $\mathcal{H}(\cdot)$ can be a contraction mapping with respect to the span seminorm under certain additional assumptions on the underlying stochastic model (to be discussed shortly). Since this property forms the foundation of the $Q$-learning algorithm we will present, we begin by formally introducing the span seminorm and discussing its properties.

\begin{definition}
    The span seminorm is defined as 
    $\mysp(x)=(\max_ix_i-\min_jx_j)/2$.
\end{definition}

Similar to a norm, the span seminorm is non-negative and satisfies the triangle inequality: $\mysp(x + y) \leq \mysp(x) + \mysp(y)$ for all $x,y \in \mathbb{R}^d$; and absolute homogeneity: $\mysp(\alpha x) = |\alpha|\, \mysp(x)$ for all $\alpha \in \mathbb{R}$ and $x \in \mathbb{R}^d$ \cite[Section 6.6.1]{puterman2014markov}. However, unlike a norm, $\mysp(x) = 0$ does not imply $x = 0$. In fact, the set $\{x \in \mathbb{R}^d \mid  \mysp(x) = 0\}$ is called the kernel of the span seminorm and is denoted by $\ker(\mysp)$. Since $\mysp(x) = 0$ if and only if $\max_i x_i = \min_j x_j$, in which case all entries of $x$ must be identical, we have $\ker(\mysp) = \{ c e \mid c \in \mathbb{R} \}$. Another important property of the span seminorm is that $\mysp(x)$ can be interpreted as the distance from $x$ to the linear subspace $\{ce| c\in\mathbb{R}\}$ with respect to $\|\cdot\|_\infty$. Since this result will be used frequently throughout the paper, we formally state it in the following lemma. The proof is provided in Appendix \ref{pf:le:span_sup_projection}.

\begin{lemma}\label{le:span_sup_projection}
    For any $x \in \mathbb{R}^d$, we have $\argmin_{c \in \mathbb{R}} \|x - c e\|_\infty=(\max_i x_i + \min_j x_j)/2$. As a result, the span seminorm of $x$ can be equivalently written as $\mysp(x) = \min_{c \in \mathbb{R}} \|x - c e\|_\infty$.
\end{lemma}

With $\mysp(\cdot)$ properly introduced, we next state our assumption on the Bellman operator $\mathcal{H}(\cdot)$.

\begin{assumption}\label{as:seminorm-contraction}
The operator $\mathcal{H}(\cdot)$ is a contraction mapping with respect to $\mysp(\cdot)$; that is, there exists $\beta \in (0,1)$ such that $\mysp(\mathcal{H}(Q_1) - \mathcal{H}(Q_2)) \leq \beta\, \mysp(Q_1 - Q_2)$ for all $Q_1, Q_2 \in \mathbb{R}^{|\mathcal{S}||\mathcal{A}|}$.
\end{assumption}

A sufficient condition for Assumption \ref{as:seminorm-contraction} to hold is
\begin{align}\label{as:alternative}
    \max_{(s,a),(s',a')} \|p(\cdot \mid s,a) - p(\cdot \mid s',a')\|_{\text{TV}} < 1,
\end{align}
where $\|\cdot\|_{\text{TV}}$ denotes the total variation distance. In this case, Assumption \ref{as:seminorm-contraction} is satisfied with $\beta=\max_{(s,a),(s',a')} \|p(\cdot \mid s,a) - p(\cdot \mid s',a')\|_{\text{TV}}$. This condition is adopted from standard MDP theory textbooks \cite{puterman2014markov}, where it was used to study the convergence of relative value iteration \cite[Proposition 6.6.1 and Theorem 6.6.2]{puterman2014markov}. 
Since the goal of this work is to establish the convergence rate for $Q$-learning in the span seminorm contraction setting, we do not attempt to relax Assumption \ref{as:seminorm-contraction}. Nevertheless, we include a detailed discussion in Section \ref{sec:discussion} on potential approaches for analyzing $Q$-learning when Assumption \ref{as:seminorm-contraction} is not satisfied.

Under Assumption~\ref{as:seminorm-contraction}, we have the following result, which shows that the Bellman equation~(\ref{eq:Bellman}) can be equivalently expressed as a fixed-point equation under the span seminorm.

\begin{lemma}\label{le:solution_same}
    Under Assumption \ref{as:seminorm-contraction}, we have 
    \begin{align*}
        \{Q^*+ce\mid c\in\mathbb{R}\}=\{Q \mid \mathcal{H}(Q) - Q = r^* e\}=\{Q \mid \mysp(\mathcal{H}(Q) - Q) = 0\}.
    \end{align*}
\end{lemma}

The proof of Lemma \ref{le:solution_same} is presented in Appendix \ref{ap:le:solution_same}. As a result of Lemma \ref{le:solution_same},
the seminorm fixed-point equation
\begin{align}\label{eq:seminorm-FPE}
    \mysp(\mathcal{H}(Q) - Q) = 0
\end{align}
and the Bellman equation (\ref{eq:Bellman}) are equivalent in the sense that they have the same set of solutions: $\mathcal{Q} = \{Q^* + c e \mid c \in \mathbb{R}\}$. Therefore, it suffices to find a solution to Equation (\ref{eq:seminorm-FPE}) in order to compute an optimal policy. For this reason, we shall refer to Equation (\ref{eq:seminorm-FPE}) as the seminorm Bellman equation, or simply the Bellman equation.

To solve the Bellman equation (\ref{eq:seminorm-FPE}), we introduce two $Q$-learning algorithms with finite-time convergence guarantees in the next two sections. The first represents the most natural form of $Q$-learning and guarantees convergence to the optimal  $Q$-function $Q^*$ in $\mysp(\cdot)$, or equivalently, convergence to the set $\mathcal{Q}=\{Q^*+ce\mid c\in\mathbb{R}\}$ with respect to $\|\cdot\|_\infty$ (cf. Lemma \ref{le:span_sup_projection}). The second ensures pointwise convergence to the centered optimal  $Q$-function.

\section{Q-Learning with Set Convergence}\label{sec:QL-set}
This section presents the algorithm and finite-time analysis of $Q$-learning with set convergence. 

\subsection{Algorithm}\label{subsec:algo:QL-set}
Our first $Q$-learning algorithm is represented in Algorithm \ref{alg:QL-set}. Note that Algorithm \ref{alg:QL-set} is surprisingly simple and represents the most natural extension of $Q$-learning in the discounted setting \cite{watkins1992q}. In fact, the only modification (aside from setting the discount factor to one) is the use of adaptive stepsizes of the form $\alpha_k(s,a) = \alpha / (N_k(s,a) + h)$, where the tunable parameter $\alpha > 0$ plays a key role in determining the convergence rate, and the parameter $h > 0$ ensures that $\alpha_k(s,a) \in (0,1)$. Importantly, the stepsize $\alpha_k(s,a)$ depends on the specific state-action pair through the counter $N_k(s,a)$ and is therefore not universal. Throughout, we refer to such stepsizes as \emph{adaptive stepsizes}. Although adaptive stepsizes of this form have been used in the existing asymptotic analysis of $Q$-learning in both the discounted and average-reward settings \cite{tsitsiklis1994asynchronous,abounadi2001learning}, they have been less explored in the context of finite-time analysis. 
The necessity and theoretical motivation behind this choice will be discussed in Section \ref{sec:viewpoint}.

\begin{algorithm}[ht]\caption{$Q$-Learning with Set Convergence}\label{alg:QL-set}
	\begin{algorithmic}[1]
		\STATE \textbf{Input:} Initializations $Q_1\in\mathbb{R}^{|\mathcal{S}||\mathcal{A}|}$, $S_1\in\mathcal{S}$, and a behavior policy $\pi$.
		\FOR{$k=1,2,\cdots,$}
        \STATE Take $A_k\sim \pi(\cdot\mid S_k)$, observe $S_{k+1}\sim p(\cdot\mid S_k,A_k)$, and receive $\mathcal{R}(S_k,A_k)$.
        \STATE Compute the temporal difference: $\delta_k=R(S_k, A_k) + \max_{a'} Q_k(S_{k+1}, a') - Q_k(S_k, A_k)$.
        \STATE Update the $Q$-function: $Q_{k+1}(s,a) = Q_k(s,a) + \alpha_k(s,a)\mathbf{1}_{\{(s,a) = (S_k,A_k)\}}\delta_k$ for all $(s,a)$,
where $\alpha_k(s,a) = \alpha / (N_k(s,a) + h)$, and $N_k(s,a) := \sum_{i=1}^k \mathbf{1}_{\{(S_i, A_i) = (s,a)\}}$ denotes the total number of visits to the state-action pair $(s, a)$ up to the $k$-th iteration.
		\ENDFOR
        \STATE \textbf{Output:} $\{Q_k\}_{k\geq 1}$
	\end{algorithmic}
\end{algorithm}

In the existing literature, the algorithm most closely related to Algorithm \ref{alg:QL-set} is RVI $Q$-learning \cite{abounadi2001learning}. In RVI $Q$-learning, the temporal difference is defined as $\delta_k = R(S_k, A_k) + \max_{a'} Q_k(S_{k+1}, a') - Q_k(S_k, A_k) - f(Q_k)$,
where $f(\cdot)$ is a Lipschitz function satisfying $f(e) = 1$ and $f(Q + c e) = f(Q) + c$ for any $c \in \mathbb{R}$. Subtracting the additional term $f(Q_k)$ in the temporal difference ensures pointwise almost sure convergence to a particular solution $Q$ of Equation \eqref{eq:Bellman} that satisfies $f(Q) = r^*$ \cite{abounadi2001learning}. In contrast, Algorithm \ref{alg:QL-set} does not include such a normalization step, as it guarantees convergence to the set $\mathcal{Q} = \{Q^* + c e \mid c \in \mathbb{R}\}$ (which will be shown shortly). From a theoretical standpoint, such a set convergence is sufficient for computing an optimal policy.

\subsection{Finite-Time Analysis}
To present the convergence rate of Algorithm \ref{alg:QL-set}, we first state our assumption regarding the behavior policy.
\begin{assumption}\label{as:MC}
    The behavior policy satisfies $\pi(a|s)>0$ for all $(s,a)$, and the Markov chain $\{S_k\}$ induced by the behavior policy $\pi$ is irreducible and aperiodic.
\end{assumption}

Assumption \ref{as:MC} is standard in the existing studies of both value-based and policy-based RL algorithms \cite{tsitsiklis1994asynchronous,tsitsiklis1999average,srikant2019finite,bhandari2018finite,li2020sample,li2024stochastic}. Intuitively, it guarantees that all state-action pairs are visited infinitely often during learning. Specifically, under Assumption \ref{as:MC}, the Markov chain $\{S_k\}$ induced by the behavior policy has a unique stationary distribution, denoted by $\mu\in\Delta(\mathcal{S})$, which satisfies $\min_{s\in\mathcal{S}}\mu(s)>0$ \cite{levin2017markov}. Moreover, there exist $C>1$ and $\rho\in (0,1)$ such that $\max_{s\in\mathcal{S}}\|p_\pi^k(S_k=\cdot| S_1=s)-\mu(\cdot)\|_{\text{TV}}\leq C\rho^{k-1}$ for all $k\geq 1$,
where $p_\pi$ denotes the transition kernel of the Markov chain $\{S_k\}$ induced by $\pi$ \cite{levin2017markov}. To aid in the statement of our main theorem, we introduce the following notation. Let 
\begin{align*}
    \tau_k=\,&\min\left\{t:C\rho^{t-1}\leq \frac{\alpha}{k+h}\right\},\;\;
    b_k=\mysp(Q_1)+\alpha|\mathcal{S}||\mathcal{A}|\log \left(\frac{\lceil (k-1) / (|\mathcal{S}||\mathcal{A}|) \rceil+h}{h}\right),\\    
    m_k=\,&b_k+\mysp(Q^*),
\end{align*}
where $\lceil x \rceil$ returns the smallest integer greater than or equal to $x$. Note that $\tau_k$, $b_k$, and $m_k$ all grow at most logarithmically in $k$. Let $D_{\min}=\min_{s,a}\mu(s)\pi(a|s)$, which is strictly positive under Assumption \ref{as:MC}. 

\begin{theorem}\label{thm:QL-set}
    Consider $\{Q_k\}$ generated by Algorithm \ref{alg:QL-set}. Suppose that Assumptions \ref{as:seminorm-contraction} and \ref{as:MC} are satisfied. Then, there exists $K>0$ such that for any $k\in \{1,2,\cdots,K\}$, we have $\mysp(Q_k-Q^*)\leq b_k+\mysp(Q^*)$ almost surely (a.s.), and for any $k\geq K+1$, we have 
    \begin{align*}
        \mathbb{E}[\mysp(Q_k-Q^*)^2]\leq B_k:=\begin{dcases}
            3m_K^2\left(\frac{K+h}{k+h}\right)^{\frac{\alpha(1-\beta)}{2}}+\frac{C_1\tau_k (m_k+1)^2}{(k+h)^{\frac{\alpha(1-\beta)}{2}}},&\text{ if }\alpha(1-\beta)<2,\\
            3m_K^2\left(\frac{K+h}{k+h}\right)+\frac{C_2\tau_k (m_k+1)^2\log(k+h)}{(k+h)},&\text{ if }\alpha(1-\beta)=2,\\
            3m_K^2\left(\frac{K+h}{k+h}\right)^{\frac{\alpha(1-\beta)}{2}}+\frac{C_1\tau_k (m_k+1)^2}{(k+h)},&\text{ if }\alpha(1-\beta)>2.
        \end{dcases}
    \end{align*}
    Here, $C_1$ and $C_2$ are problem-dependent constants defined as
    \begin{align*}       C_1=\frac{c_1\alpha^2|\mathcal{S}||\mathcal{A}|\log(|\mathcal{S}||\mathcal{A}|)}{(1-\rho)(1-\beta) D_{\min}^2\min(1,\alpha)|2-(1-\beta)\alpha|},\;\text{and}\;C_2=\frac{c_2|\mathcal{S}||\mathcal{A}|\log(|\mathcal{S}||\mathcal{A}|)}{(1-\rho)(1-\beta)^3 D_{\min}^2},
    \end{align*}
    where $c_1$ and $c_2$ are absolute constants.
\end{theorem}

The proof of Theorem \ref{thm:QL-set} is presented in Section \ref{sec:proof}. Due to the presence of Markovian noise and adaptive/stochastic stepsizes, the convergence bound does not hold from the initial iteration. Specifically, prior to iteration $K$, we establish an almost-sure bound that grows at most logarithmically with $k$. After iteration $K$, the ``averaging'' effect becomes dominant, and the mean-square error begins to decay, with the rate of convergence depending critically on the choice of the constant $\alpha$ in Algorithm \ref{alg:QL-set}. In particular, if $\alpha$ is below the threshold $2/(1 - \beta)$, the convergence rate is $\smash{\mathcal{O}(k^{-\alpha(1 - \beta)/2})}$, which can be arbitrarily slow. Conversely, if $\alpha$ exceeds the threshold, the convergence rate improves to the optimal $\mathcal{O}(1/k)$ up to logarithmic factors. A qualitatively similar phenomenon has been observed in norm-contractive SA algorithms \cite{chen2024lyapunov}, linear SA algorithms \cite{bhandari2018finite}, and stochastic gradient descent/ascent algorithms \cite{lan2020first}.

Based on Theorem \ref{thm:QL-set}, we have the following corollary for the sample complexity.
\begin{corollary}\label{co:sc}
    Given $\epsilon>0$, to achieve $\mathbb{E}[\mysp(Q_k-Q^*)]\leq \epsilon$ with Algorithm \ref{alg:QL-set}, the sample complexity is $\tilde{\mathcal{O}}(|\mathcal{S}|^{-3}|\mathcal{A}|^{-3}D_{\min}^{-2}(1-\beta)^{-5}\epsilon^{-2})$
\end{corollary}

The proof of Corollary \ref{co:sc} is provided in Appendix \ref{pf:co:sc}. Notably, the dependence on the desired accuracy level is $\tilde{\mathcal{O}}(\epsilon^{-2})$, which is unimprovable in general \cite{jin2021towards}. While we make the dependence on the size of the state–action space and the seminorm contraction factor explicit, these terms are unlikely to be optimal in light of existing information‐theoretic lower bounds \cite{jin2021towards,wangoptimal}. It is worth noting, however, that the lower bound in \cite{jin2021towards,wangoptimal} was derived under a generative model that provides i.i.d. samples, whereas our analysis considers the more challenging Markovian sampling setting. Despite this discrepancy, tightening the dependencies on $|\mathcal{S}||\mathcal{A}|$, $1/(1-\beta)$, and $D_{\min}$, whether through refined analysis or improved algorithmic techniques such as Polyak averaging or variance reduction, remains an important direction for future research.

\section{Q-Learning with Pointwise Convergence}\label{sec:QL-pointwise}
Although $\{Q_k\}$ generated by Algorithm \ref{alg:QL-set} achieves mean-square convergence to $Q^*$ in $\mysp(\cdot)$, due to using the span seminorm as the performance metric, the error $Q_k - Q^*$ can still diverge in the direction of the all-ones vector, which corresponds to the kernel space of $\mysp(\cdot)$. To see this clearly, consider the following simple example.

\begin{example}\label{example}
Let $\mathcal{M}$ be an average-reward MDP consists of only one state-action pair $(s,a)$ with $\mathcal{R}(s,a) = 1$. In this case, Algorithm \ref{alg:QL-set} reduces to the following deterministic update equation:
\begin{align*}
    Q_{k+1}(s,a) =Q_k(s,a)+\frac{\alpha}{k + h}=\cdots=Q_1(s,a)+\sum_{i=1}^k\frac{\alpha}{i + h},\quad \forall\,k \geq 1,
\end{align*}
which implies $\lim_{k \to \infty} Q_k(s,a) = \infty$ because the harmonic series diverges.   
\end{example}

Although the divergence illustrated in Example \ref{example} has no theoretical impact on identifying an optimal policy because there is only one policy (choosing action $a$ at state $s$ with probability one), it is nevertheless preferable in practice to avoid any form of divergence when designing algorithms. This motivates our second $Q$-learning algorithm, which achieves pointwise convergence to the centered optimal  $Q$-function.

Our algorithm design is motivated by the following observation. Let $x, y \in \mathbb{R}^d$ be two arbitrary vectors. It is known that $\mysp(x - y) = 0$ does not, in general, imply $\|x - y\|_\infty = 0$, or even that $\|x - y\|_\infty$ is bounded, since $x - y$ may lie in the direction of the all-ones vector $e$. However, if both $x$ and $y$ are \emph{centered} in the sense that $\max_i x_i + \min_j x_j = 0$ and $\max_i y_i + \min_j y_j = 0$, or equivalently, $\|x\|_\infty = \mysp(x)$ and $\|y\|_\infty = \mysp(y)$ (cf.~Lemma~\ref{le:span_sup_projection}), then the quantities $\mysp(x-y)$ and $\|x-y\|_\infty$ differ only by a constant multiplicative factor. This observation is formalized in the following lemma, whose proof is provided in Appendix~\ref{pf:le:span_sup_equivalence}.

\begin{lemma}\label{le:span_sup_equivalence}
    Let $x,y\in\mathbb{R}^d$ be such that $\|x\|_\infty=\mysp(x)$ and $\|y\|_\infty=\mysp(y)$. Then, we have
    \begin{align*}
        \mysp(x-y)\leq \|x-y\|_\infty\leq 2\mysp(x-y).
    \end{align*}
\end{lemma}

Returning to the algorithm design and starting with Algorithm \ref{alg:QL-set}, in light of Lemma \ref{le:span_sup_equivalence}, as long as we can ensure that $Q_k$ is centered, the combination of Theorem \ref{thm:QL-set} and Lemma \ref{le:span_sup_equivalence} guarantees the mean-square convergence of $Q_k$ to the centered optimal  $Q$-function in the $\|\cdot\|_\infty$ norm. That is, $Q_k$ converges to an element $\tilde{Q}^*$ in the set $\mathcal{Q} = \{Q^* + c e \mid c \in \mathbb{R}\}$ satisfying $\max_{s,a} \tilde{Q}^*(s,a) + \min_{s,a} \tilde{Q}^*(s,a) = 0$. This motivates the design of Algorithm \ref{alg:QL-pointwise}.

\begin{algorithm}[ht]\caption{$Q$-Learning with Pointwise Convergence}\label{alg:QL-pointwise}
	\begin{algorithmic}[1]
		\STATE \textbf{Input:} Initializations $Q_1\in\mathbb{R}^{|\mathcal{S}||\mathcal{A}|}$, $S_1\in\mathcal{S}$, and a behavior policy $\pi$.
		\FOR{$k=1,2,\cdots,$}
        \STATE Take $A_k\sim \pi(\cdot\mid S_k)$, observe $S_{k+1}\sim p(\cdot\mid S_k,A_k)$, and receive $\mathcal{R}(S_k,A_k)$.
        \STATE Compute the temporal difference: $\delta_k=R(S_k, A_k) + \max_{a'} Q_k(S_{k+1}, a') - Q_k(S_k, A_k)$.
        \STATE Update the $Q$-function: $\tilde{Q}_{k+1}(s,a) = Q_k(s,a) + \alpha_k(s,a)\mathbf{1}_{\{(s,a) = (S_k,A_k)\}}\delta_k$ for all $(s,a)$.
        \STATE Centering the iterate for pointwise convergence: $Q_{k+1}=\tilde{Q}_{k+1}-g(\tilde{Q}_{k+1})e$,
        where $g(Q) = [\max_{s,a} Q(s,a) + \min_{s',a'} Q(s',a')]/2$.
		\ENDFOR
        \STATE \textbf{Output:} $\{Q_k\}_{k\geq 1}$
	\end{algorithmic}
\end{algorithm}

Compared with Algorithm \ref{alg:QL-set}, the only difference is the additional step in Line 6 of Algorithm \ref{alg:QL-pointwise}. To illustrate the purpose of this step, recall from Lemma \ref{le:span_sup_projection} that an important property of the span seminorm $\mysp(Q)$ is that it can be interpreted as the distance from $Q$ to $\ker(\mysp)$ with respect to $\|\cdot\|_\infty$, i.e., $\mysp(Q) = \min_{c \in \mathbb{R}} \|Q - c e\|_\infty$. Moreover, we have $\arg\min_{c \in \mathbb{R}} \|Q - c e\|_\infty = [\max_{s,a}Q(s,a)+\min_{s,a}Q(s,a)]/2= g(Q)$. Therefore, for all $k \geq 1$, we have by Line 6 of Algorithm \ref{alg:QL-pointwise} that $\|Q_k\|_\infty = \|\tilde{Q}_k - g(\tilde{Q}_k) e\|_\infty = \mysp(\tilde{Q}_k) = \mysp(Q_k)$,
where the last equality follows from $\tilde{Q}_k - Q_k \in \ker(\mysp)$. This chain of equalities implies that $Q_k$ is always centered. As a result, the sequence $\{Q_k\}$ generated by Algorithm \ref{alg:QL-pointwise} converges (in the mean-square sense) in $\|\cdot\|_\infty$ to the centered optimal  $Q$-function $\tilde{Q}^* = Q^* - g(Q^*) e$. The explicit convergence rate is characterized in the following theorem, whose proof is presented in Section \ref{sec:proof}.

\begin{theorem}\label{thm:QL-pointwise}
    Consider $\{Q_k\}$ generated by Algorithm \ref{alg:QL-pointwise}. Suppose that Assumptions \ref{as:seminorm-contraction} and \ref{as:MC} are satisfied. Then, for any $k\in \{1,2,\cdots,K\}$, we have $\|Q_k-\tilde{Q}^*\|_\infty\leq b_k+\mysp(Q^*)$ a.s., and for any $k\geq K+1$, we have $\mathbb{E}[\|Q_k-\tilde{Q}^*\|_\infty^2]\leq 4B_k$, where $K$, $b_k$, and $B_k$ were defined in Theorem \ref{thm:QL-set}. As a result, to achieve $\mathbb{E}[\|Q_k-\tilde{Q}^*\|_\infty]\leq \epsilon$, the sample complexity is $\tilde{\mathcal{O}} (|\mathcal{S}|^3|\mathcal{A}|^3D_{\min}^{-2}(1-\beta)^{-5}\epsilon^{-2})$.
\end{theorem}

Note that while the convergence bounds are the same as those in Theorem \ref{thm:QL-set}, the guarantees are stronger, as we now establish pointwise mean-square convergence to the centered optimal $Q$-function $\tilde{Q}^*$, rather than convergence to the set $\mathcal{Q} = \{Q^* + c e \mid c \in \mathbb{R}\}$.

We end this section with a side remark. As discussed in Section \ref{subsec:algo:QL-set}, the closest existing algorithm to ours is RVI $Q$-learning \cite{abounadi2001learning}. In RVI $Q$-learning, pointwise convergence is achieved by subtracting a Lipschitz function $f(Q_k)$ in the temporal difference. In contrast, for Algorithm \ref{alg:QL-pointwise}, pointwise convergence is achieved through a different algorithmic design: the centering step.

\section{The Necessity of Using Adaptive Stepsizes}\label{sec:viewpoint}
As discussed in the previous section, the most important feature of Algorithms \ref{alg:QL-set} and \ref{alg:QL-pointwise} is their use of adaptive stepsizes. Theoretically breaking the strong correlation between the adaptive stepsizes and the iterates $Q_k$ will be our primary technical innovation in the proof of Theorems \ref{thm:QL-set} and \ref{thm:QL-pointwise}. However, before delving into the proofs, it is important to first answer the following questions:

\begin{enumerate}[(1)]
    \item Is the use of adaptive stepsizes necessary? In particular, how do these algorithms behave if we instead use universal stepsizes (e.g., $\alpha_k(s,a) = 1/k$ for all $(s,a)$)?
    \item If adaptive stepsizes are indeed necessary, how should we interpret their role?
    \item Once their role is characterized, can we design adaptive stepsizes beyond $\alpha_k(s,a) = \alpha/(N_k(s,a) + h)$ to achieve different convergence behaviors?
\end{enumerate}

In this section, we provide clear answers to these questions. The insights developed in this section will play a central role in guiding our proof of Theorems \ref{thm:QL-set} and \ref{thm:QL-pointwise} in Section \ref{sec:proof}.

\subsection{Q-Learning with Universal Stepsizes: Provable Convergence to the Wrong Target}\label{subsec:challenge}
Since Algorithm \ref{alg:QL-set} represents the most natural form of $Q$-learning, we use it as an example to show that if one uses universal stepsizes, the algorithm fails to converge in $\mysp(\cdot)$ to $Q^*$. To this end, we first present the main update equation for $Q$-learning with universal stepsizes in the following:
\begin{align}\label{algo:illustration}
    Q_{k+1}(s,a) =
    Q_k(s,a) + \alpha_k\mathbf{1}_{\{(S_k,A_k)=(s,a)\}}\left(R(S_k, A_k) + \max_{a'} Q_k(S_{k+1}, a') - Q_k(S_k, A_k)\right)
\end{align}
for all $(s,a)$, where the asynchronous nature of the update is captured by the indicator function $\mathbf{1}_{\{(S_k,A_k)=(s,a)\}}$. Here, the stepsize $\alpha_k$ does not depend on the state-action pairs. 

The $Q$-learning algorithm described in Equation (\ref{algo:illustration}) takes the typical form of a Markovian SA algorithm. To show that it does not converge to $Q^*$ even in $\mysp(\cdot)$, we begin with the reformulation and identify the target equation it aims to solve.  Let $Y_k=(S_k,A_k,S_{k+1})$ for all $k\geq 1$. Note that $\{Y_k\}$ forms a Markov chain, with state space $\mathcal{Y}=\mathcal{S}\times \mathcal{A}\times \mathcal{S}$. In addition, under Assumption \ref{as:MC}, the Markov chain $\{Y_k\}$ admits a unique stationary distribution $\nu\in \Delta(\mathcal{Y})$ satisfying $\nu(s,a,s')=\mu(s)\pi(a|s)p(s'|s,a)$ for all $(s,a,s')\in\mathcal{Y}$. Let $G:\mathbb{R}^{|\mathcal{S}||\mathcal{A}|}\times \mathcal{Y}\to \mathbb{R}^{|\mathcal{S}||\mathcal{A}|}$ be an operator defined such that given inputs $Q\in \mathbb{R}^{|\mathcal{S}||\mathcal{A}|}$ and $y=(s_0,a_0,s_1)\in\mathcal{Y}$, the $(s,a)$-th entry of the output is given by
\begin{align*}
    [G(Q,y)](s,a)=\mathbf{1}_{\{(s_0,a_0)=(s,a)\}}\big(\mathcal{R}(s_0,a_0)+\max_{a'\in\mathcal{A}}Q(s_1,a')-Q(s_0,a_0)\big)+Q(s,a).
\end{align*}
With $\{Y_k\}$ and $G(\cdot)$ defined above, Equation (\ref{algo:illustration}) can be formulated as a Markovian SA:
\begin{align}\label{algo:illustration-reformulation}
    Q_{k+1}=Q_k+\alpha_k(G(Q_k,Y_k)-Q_k).
\end{align}
Let $\Bar{\mathcal{H}}:\mathbb{R}^{|\mathcal{S}||\mathcal{A}|}\to \mathbb{R}^{|\mathcal{S}||\mathcal{A}|}$ be the ``expected'' operator defined as:
\begin{align}\label{def:asynchronousBE}
    \Bar{\mathcal{H}}(Q):=\mathbb{E}_{Y\sim \nu}[G(Q,Y)]=[(I-D)+D\mathcal{H}](Q),\quad \forall\,Q\in\mathbb{R}^{|\mathcal{S}||\mathcal{A}|},
\end{align}
where $\mathcal{H}(\cdot)$ is the Bellman operator, and $D$ is an $|\mathcal{S}||\mathcal{A}| \times |\mathcal{S}||\mathcal{A}|$ diagonal matrix with diagonal entries $\{\mu(s)\pi(a | s)\}_{(s,a) \in \mathcal{S} \times \mathcal{A}}$. In other words, the $(s,a)$-th diagonal entry of $D$, denoted by $D(s,a)$, corresponds to the probability of visiting the state-action pair $(s,a)$ under the stationary distribution of the Markov chain $\{(S_k, A_k)\}$ induced by the policy $\pi$. 

Inspired by \cite{chen2024lyapunov} (which studies $Q$-learning in the discounted setting), we refer to $\Bar{\mathcal{H}}(\cdot)$ as the \emph{asynchronous Bellman operator}. The reason is that for each $(s,a)$, the output $[\Bar{\mathcal{H}}(Q)](s,a)$, according to its definition, can be viewed as the expectation of a random variable that takes $[\mathcal{H}(Q)](s,a)$ with probability $D(s, a)$ and takes $Q(s,a)$ with probability $1-D(s,a)$, capturing the asynchronous nature of $Q$-learning. The following lemma shows that $\Bar{\mathcal{H}}(\cdot)$ is also a contraction mapping with respect to the span seminorm.

\begin{lemma}\label{le:ABO-contraction}
    Under Assumptions \ref{as:seminorm-contraction} and \ref{as:MC}, the asynchronous Bellman operator $\bar{\mathcal{H}}(\cdot)$ is a contraction mapping with respect to $\mysp(\cdot)$, with contraction factor $\bar{\beta}:=1-(1-\beta)D_{\min}$.
\end{lemma}

The proof of Lemma \ref{le:ABO-contraction} is presented in Appendix \ref{pf:le:ABO-contraction}.
In light of Lemma \ref{le:ABO-contraction}, we identify that the Markovian SA algorithm described in Equation (\ref{algo:illustration-reformulation}) solves the seminorm fixed-point equation
\begin{align}\label{eq:asynchronous_BE}
    \mysp(\Bar{\mathcal{H}}(Q) - Q) = 0,
\end{align}
which we refer to as the \emph{asynchronous Bellman equation}, in contrast to the original Bellman equation (\ref{eq:seminorm-FPE}). Therefore, applying recent results on Markovian SA with seminorm contractive operators \cite{chen2025non}, we show in the following proposition that $Q$-learning with universal stepsizes, as described in Equation (\ref{algo:illustration}), provably achieves mean-square convergence in $\mysp(\cdot)$ to a \textit{wrong target}. See Appendix \ref{pf:prop:universal-convergence} for its proof.

\begin{proposition}\label{prop:universal-convergence}
Suppose that Assumptions \ref{as:seminorm-contraction} and \ref{as:MC} are satisfied. Then, we have
\begin{align}\label{eq:uniform}
    \begin{dcases}
        \{Q \mid  \mysp(\Bar{\mathcal{H}}(Q)-Q)=0\}=\{Q \mid  \mysp(\mathcal{H}(Q)-Q)=0\},&\text{ if }D=I/(|\mathcal{S}||\mathcal{A}|),\\
        \{Q \mid  \mysp(\Bar{\mathcal{H}}(Q)-Q)=0\}\cap \{Q \mid  \mysp(\mathcal{H}(Q)-Q)=0\}=\emptyset,&\text{ otherwise}.
    \end{dcases}
\end{align}
Moreover, when $\alpha_k=\alpha/(k+h)$ with appropriately chosen $\alpha$ and $h$, the $Q$-learning algorithm described in Equation (\ref{algo:illustration}) achieves $\mathbb{E}[\mysp(Q_k-\Bar{Q}^*)^2]\leq \tilde{\mathcal{O}}(1/k)$,
where $\bar{Q}^*$ is a particular solution to the asynchronous Bellman equation $\mysp(\Bar{\mathcal{H}}(Q)-Q)=0$.
\end{proposition}

Note that the sets of solutions to the asynchronous Bellman equation and the original Bellman equation are completely disjoint, except in the special case where $D = I / (|\mathcal{S}||\mathcal{A}|)$, which corresponds to the stationary distribution of the Markov chain $\{(S_k, A_k)\}$ induced by the behavior policy being uniform. Proposition \ref{prop:universal-convergence} reveals a critical issue: \textit{whenever $D \neq I / (|\mathcal{S}||\mathcal{A}|)$, $Q$-learning with universal stepsizes is guaranteed not to converge to any point in the desired solution set $\mathcal{Q} = \{Q^* + c e \mid c \in \mathbb{R}\}$ of the original Bellman equation $\mysp(\mathcal{H}(Q) - Q) = 0$.} In Appendix \ref{sec:simulations}, we conduct a sequence of numerical simulations to further verify Proposition \ref{prop:universal-convergence}.

In summary, the fundamental issue for $Q$-learning with universal stepsizes in the average-reward setting is that the combination of a seminorm contraction mapping and asynchronous updates makes the set of solutions to the asynchronous Bellman equation $\mysp(\Bar{\mathcal{H}}(Q)-Q)=0$ completely different compared to that of the original Bellman equation $\mysp(\mathcal{H}(Q)-Q)=0$.

\paragraph{The Discounted Setting.} 
One might ask: why is discounted $Q$-learning able to achieve provable convergence to the optimal $Q$-function with universal stepsizes? To illustrate, consider the $\gamma$-discounted MDP that shares the same transition kernel and reward function as the average-reward MDP studied in this work. Let $\smash{Q^*_\gamma}$ be the optimal $Q$-function, and let $\mathcal{H}_\gamma(\cdot)$ denote the Bellman operator, which is a $\gamma$-contraction mapping with respect to $\|\cdot\|_\infty$ \cite{puterman2014markov,sutton2018reinforcement,bertsekas1996neuro}.

Following the same line of reasoning, discounted $Q$-learning (with asynchronous updates) can be formulated as a Markovian SA algorithm for solving the fixed-point equation $\bar{\mathcal{H}}_\gamma(Q) = Q$, where $\bar{\mathcal{H}}_\gamma(\cdot)$ is the \emph{asynchronous Bellman operator} in the discounted setting, defined as $\bar{\mathcal{H}}_\gamma(Q) = [(I - D) + D \mathcal{H}_\gamma](Q)$.
It has been shown in the existing literature \cite{chen2024lyapunov,borkar_SA} that, under Assumption \ref{as:MC}, the asynchronous Bellman operator $\bar{\mathcal{H}}_\gamma(\cdot)$ maintains the following two important properties of the original Bellman operator $\mathcal{H}_\gamma(\cdot)$: (1) $\bar{\mathcal{H}}_\gamma(\cdot)$ is a contraction mapping with respect to $\|\cdot\|_\infty$, and (2) the asynchronous Bellman equation $\bar{\mathcal{H}}_\gamma(Q) = Q$ admits a unique solution $Q_\gamma^*$, which is the optimal $Q$-function in the discounted setting. Consequently, discounted $Q$-learning with universal stepsizes converges to $Q_\gamma^*$ by standard results on Markovian SA with norm-contractive operators \cite{borkar2000ode,benveniste2012adaptive}.

\subsection{Q-Learning with Adaptive Stepsizes: Implicit Importance Sampling to the Rescue}\label{subsec:importance-sampling}

In view of Proposition~\ref{prop:universal-convergence} and the definition of the asynchronous Bellman operator in~\eqref{def:asynchronousBE}, suppose that $(S_k, A_k, S_{k+1})$ were sampled from the distribution $\nu' \in \Delta(\mathcal{Y})$ defined by $\nu'(y) = p(s' \mid s, a)/(|\mathcal{S}||\mathcal{A}|)$ for all $y = (s, a, s') \in \mathcal{Y}$. Equivalently, $\nu'$ is induced by sampling $S_k \sim \mathrm{Unif}(\mathcal{S})$, $A_k \sim \mathrm{Unif}(\mathcal{A})$, and $S_{k+1} \sim p(\cdot \mid S_k, A_k)$. In this case, we would have $D = I/(|\mathcal{S}||\mathcal{A}|)$, and the asynchronous Bellman equation would share the same set of solutions as the original Bellman equation. However, when following the trajectory of the Markov chain $\{Y_k = (S_k, A_k, S_{k+1})\}$, even asymptotically, we receive samples only from its stationary distribution $\nu$, given by $\nu(y) = \mu(s)\pi(a \mid s)p(s' \mid s, a)$ for any $y = (s, a, s') \in \mathcal{Y}$. In this view, the natural fix for $Q$-learning with universal stepsizes is \emph{importance sampling}, which is a technique widely used in statistics \cite{kloek1978bayesian}, rare-event simulation \cite{bucklew2004introduction}, and off-policy RL \cite{espeholt2018impala}. Specifically, when direct sampling from a target distribution is difficult, importance sampling enables the estimation of expectations with respect to the target distribution by instead drawing samples from a different, more accessible distribution, referred to as the behavior distribution. The key idea is to reweight the samples drawn from the behavior distribution according to the likelihood ratio between the target and behavior distributions.

In view of Equation~(\ref{algo:illustration}), to perform importance sampling with $\nu'(\cdot)$ as the target distribution and $\nu(\cdot)$ as the behavior distribution, the algorithm becomes
\begin{align}
    Q_{k+1}(s, a) 
    =\,&
    Q_k(s, a) + \alpha_k \frac{\mathbf{1}_{\{(S_k, A_k) = (s, a)\}} \nu'(s,a,s')}{\nu(s,a,s')}\delta_k \nonumber\\
    =\,&Q_k(s, a) + \alpha_k \frac{\mathbf{1}_{\{(S_k, A_k) = (s, a)\}}}{|\mathcal{S}||\mathcal{A}|D(s,a)}
    \delta_k \nonumber\\
    =\,&Q_k(s, a) + \tilde{\alpha}_k \frac{\mathbf{1}_{\{(S_k, A_k) = (s, a)\}}}{D(s,a)}
    \delta_k,\label{algo:illustration2}
\end{align}
for all $(s,a) \in \mathcal{S} \times \mathcal{A}$,  
where the last equality follows by absorbing the $1/(|\mathcal{S}||\mathcal{A}|)$ factor into the stepsizes, i.e., by redefining $\tilde{\alpha}_k = \alpha_k / (|\mathcal{S}||\mathcal{A}|)$.

While the algorithm described in Equation (\ref{algo:illustration2}) seems promising, it is impractical because we do not have access to the stationary distribution of the Markov chain $\{(S_k,A_k)\}$ induced by $\pi$. A natural way to obtain an estimate of $D(s,a)$ is by using the empirical frequency. Specifically, recall that we have denoted $N_k(s, a)$ as the total number of visits to state-action pair $(s, a)$ up to the $k$-th iteration. Then, the estimator $N_k(s, a)/k$, or more generally, 
\begin{align}\label{eq:IS}
    D_k(s, a):=\frac{N_k(s, a) + h}{k+h},\quad \forall\,h\geq 0,
\end{align}
is an asymptotically unbiased estimator of $D(s,a)$ because under Assumption \ref{as:MC} \cite{meyn2012markov}. Substituting the estimator of $D(s, a)$ presented in Equation (\ref{eq:IS}) into Equation (\ref{algo:illustration2}), when the universal stepsize is set to $\tilde{\alpha}_k = \alpha / (k + h)$, we obtain
\begin{align}
    Q_{k+1}(s, a) 
    =\;&
    Q_k(s, a) + \frac{\tilde{\alpha}_k\mathbf{1}_{\{(S_k, A_k) = (s, a)\}}}{D_k(s,a)}\delta_k\label{eq:right-viewpoint}\\
    =\;&
    Q_k(s, a) + \frac{\alpha \mathbf{1}_{\{(S_k, A_k) = (s, a)\}}}{N_k(s, a) + h}\delta_k\nonumber\\
    =\;&
    Q_k(s, a) + \alpha_k(s,a) \mathbf{1}_{\{(S_k, A_k) = (s, a)\}}\delta_k\nonumber
\end{align}
for all $(s,a)$.
Note that this is exactly the main update equation of the $Q$-learning algorithm presented in Algorithm \ref{alg:QL-set}, which uses adaptive stepsizes of the form $\alpha_k(s, a) = \alpha / (N_k(s, a) + h)$.

In summary, adaptive stepsizes in $Q$-learning (which is necessary, as illustrated in Section \ref{subsec:challenge}) can be viewed as a form of implicit importance sampling that counteracts the effect of asynchronous updates, where the importance sampling weights are effectively constructed from the empirical frequency of visits to each state-action pair.

\subsection{Other Choices of Adaptive Stepsizes}\label{subsec:adaptive_stepsizes}

Based on the insights developed in previous sections, we now discuss other choices of adaptive stepsizes. To motivate this topic, note that for SA algorithms with universal stepsizes (such as norm-contractive SA \cite{chen2024lyapunov}, linear SA with a Hurwitz matrix \cite{srikant2019finite,bhandari2018finite}, or SGD with a smooth and strongly convex objective \cite{bottou2018optimization,lan2020first}), the stepsize can be flexibly chosen as $\alpha_k = \alpha / (k + h)^z$ with $z \in [0,1]$. Different choices of $z$ lead to distinct convergence behaviors. Specifically, when using constant stepsizes, i.e., $\alpha_k \equiv \alpha$ (corresponding to $z = 0$), the algorithm does not achieve pointwise convergence but instead converges \textit{geometrically} to a ball centered at the desired limit, with the radius of the ball proportional to the stepsize. When using diminishing stepsizes of the form $\alpha_k = \alpha / (k + h)$, the convergence behavior is qualitatively similar to that established in this work, achieving an $\mathcal{O}(1/k)$ rate of convergence when $\alpha$ exceeds a certain threshold. When using polynomially decaying stepsizes, i.e., $\alpha_k = \alpha / (k + h)^z$ for $z \in (0,1)$, the algorithm converges to the desired limit at a rate of $\mathcal{O}(1/k^z)$, which, while suboptimal, is more robust since it does not depend sensitively on the choice of $\alpha$, in contrast to the $\alpha_k = \alpha / (k + h)$ setting. A natural question is whether all these convergence behaviors can be achieved through different schedules of adaptive stepsizes in the setting of average-reward $Q$-learning.

To answer this question, we start with Equation~(\ref{eq:right-viewpoint}), which, as illustrated earlier, represents our way of viewing average-reward $Q$-learning. In particular, adaptive stepsizes can be interpreted as a combination of universal stepsizes and importance-sampling factors, where the latter are estimated using the empirical frequencies $\{D_k(s,a)\}_{(s,a)\in\mathcal{S}\times\mathcal{A}}$.

\begin{itemize}
    \item Setting $\tilde{\alpha}_k \equiv \alpha$ in Equation~(\ref{eq:right-viewpoint}) yields
\begin{align*}
    Q_{k+1}(s,a) = Q_k(s,a) + \frac{\alpha (k + h)}{N_k(s,a) + h} \mathbf{1}_{\{(S_k, A_k) = (s,a)\}} \delta_k,
\end{align*}
so the adaptive stepsize can be interpreted as $\alpha_k(s,a) = [\alpha (k + h)] / (N_k(s,a) + h)$. We expect this to produce convergence behavior analogous to norm-contractive SA with constant universal stepsizes, namely geometric convergence to a ball with radius proportional to $\alpha$ \cite{chen2024lyapunov}.
\item Setting $\tilde{\alpha}_k = \alpha / (k + h)^z$ in Equation~(\ref{eq:right-viewpoint}) leads to
\begin{align*}
    Q_{k+1}(s,a) = Q_k(s,a) + \frac{\alpha (k + h)^{1 - z}}{N_k(s,a) + h} \mathbf{1}_{\{(S_k, A_k) = (s,a)\}} \delta_k,
\end{align*}
so the adaptive stepsize is $\alpha_k(s,a) = [\alpha (k + h)^{1 - z}] / (N_k(s,a) + h)$. In this case, we expect convergence behavior similar to norm-contractive SA with polynomial stepsize $\alpha_k = \alpha / (k + h)^z$ for $z\in (0,1)$, robustly achieving an $\mathcal{O}(1/k^z)$ rate independent of $\alpha$ \cite{chen2024lyapunov}.
\end{itemize}

In Appendix~\ref{sec:simulations}, we conduct numerical simulations to verify the conjectured behavior of average-reward $Q$-learning under the above choices of adaptive stepsizes. A rigorous theoretical analysis of these behaviors is an interesting direction for future work.

\section{Proof of Theorem \ref{thm:QL-set} and Theorem \ref{thm:QL-pointwise}}\label{sec:proof}
To prove both Theorem \ref{thm:QL-set} and Theorem \ref{thm:QL-pointwise} in one shot, we present the analysis for a general class of $Q$-learning algorithms (cf. Algorithm \ref{alg:QL-generic}) that covers both cases. Note that Algorithm~\ref{alg:QL-set} corresponds to the case where $Q_{k+1} - \tilde{Q}_{k+1} = 0 \in \text{ker}(\mysp)$ in Equation~(\ref{alg:main2}). Algorithm~\ref{alg:QL-pointwise} corresponds to the case where $Q_{k+1} - \tilde{Q}_{k+1} = -g(\tilde{Q}_{k+1})e \in \text{ker}(\mysp)$. Both are special cases of Algorithm~\ref{alg:QL-generic}. Moreover, because Equation~(\ref{alg:main2}) does not define a unique update rule, it offers flexibility for future algorithm design with provable finite-time guarantees.

\begin{algorithm}[ht]\caption{A Generic Class of $Q$-Learning Algorithms}\label{alg:QL-generic}
	\begin{algorithmic}[1]
		\STATE \textbf{Input:} Initializations $Q_1\in\mathbb{R}^{|\mathcal{S}||\mathcal{A}|}$, $S_1\in\mathcal{S}$, and a behavior policy $\pi$.
		\FOR{$k=1,2,\cdots,$}
        \STATE Take $A_k\sim \pi(\cdot\mid S_k)$, observe $S_{k+1}\sim p(\cdot\mid S_k,A_k)$, and receive $\mathcal{R}(S_k,A_k)$.
        \STATE Compute the temporal difference: $\delta_k=R(S_k, A_k) + \max_{a'} Q_k(S_{k+1}, a') - Q_k(S_k, A_k)$.
        \STATE Update the $Q$-function: 
        \begin{align}
        &\tilde{Q}_{k+1}(s,a)=Q_k(s,a)+\alpha_k(s,a)\mathbf{1}_{\{(s,a)=(S_k,A_k)\}}\delta_k,\quad \forall\,(s,a),\label{alg:main1}\\
        &Q_{k+1}-\tilde{Q}_{k+1}\in\text{ker}(\mysp),\label{alg:main2}
        \end{align}
        where $\alpha_k(s,a)=\alpha/(N_k(s,a)+h)$.
		\ENDFOR
        \STATE \textbf{Output:} $\{Q_k\}_{k\geq 1}$
	\end{algorithmic}
\end{algorithm}

In the rest of this section, we show that Theorem \ref{thm:QL-set} holds for Algorithm \ref{alg:QL-generic}, which directly implies its validity for Algorithms \ref{alg:QL-set} and \ref{alg:QL-pointwise}. Moreover, combining Theorem \ref{thm:QL-set} with Lemma \ref{le:span_sup_equivalence} immediately yields Theorem \ref{thm:QL-pointwise}.

\subsection{A Markovian Reformulation of Non-Markovian Stochastic Approximation}
Although the underlying stochastic process $\{(S_k,A_k)\}$ that drives Algorithm \ref{alg:QL-generic} is a Markov chain, since the $k$-th iteration of Algorithm \ref{alg:QL-generic} depends on the entire history of the sample trajectory $(S_1,A_1,S_2,A_2,\cdots,S_k,A_k)$ through the use of adaptive stepsizes, the algorithm, as an SA, is non-Markovian. This leads to the first step of the proof, where we reformulate the algorithm as a (time-inhomogeneous) Markovian SA. 

In light of the discussion in Section \ref{subsec:importance-sampling}, especially Equation (\ref{eq:right-viewpoint}), the $k$-th iteration of Algorithm \ref{alg:QL-generic} depends deterministically on the following random variables: the current iterate $Q_k$, the $k$-th transition $(S_k,A_k,S_{k+1})$, and the empirical frequency matrix $D_k$, which is an $|\mathcal{S}||\mathcal{A}|$-by-$|\mathcal{S}||\mathcal{A}|$ diagonal matrix with diagonal entries $\{(N_k(s,a)+h)/(k+h)\}_{(s,a)\in\mathcal{S}\times \mathcal{A}}$. This motivates us to define a stochastic process $\{Z_k\}$ as $Z_k=(D_k,S_k,A_k,S_{k+1})$ for all $k\geq 1$, whose state space is denoted by $\mathcal{Z}$.  To see that $\{Z_k\}$ forms a time-inhomogeneous Markov chain, given $Z_k=(D_k,S_k,A_k,S_{k+1})$, consider the distribution of $Z_{k+1}=(D_{k+1},S_{k+1},A_{k+1},S_{k+2})$. Since $S_{k+1}$ is given in $Z_k$, $A_{k+1}\sim \pi(\cdot|S_{k+1})$, $S_{k+2}\sim p(\cdot|S_{k+1},A_{k+1})$, and
\begin{align}
D_{k+1}(s,a) =\,&\frac{N_{k}(s,a)+\mathbf{1}_{\{(S_{k+1},A_{k+1})=(s,a)\}}+h}{k+1+h}\nonumber\\
=\,&\frac{(k+h)D_k(s,a)+\mathbf{1}_{\{(S_{k+1},A_{k+1})=(s,a)\}}+h}{k+1+h},\quad \forall\,(s,a),\label{eq:D_k-update}
\end{align}
which is a deterministic function of $k$,  $(S_{k+1},A_{k+1})$, and $D_k$, the stochastic process $\{Z_k\}$ is a time-inhomogeneous Markov chain, where the time inhomogeneity arises from the fact that the transition of $D_k$ depends on $k$.

Although $\{Z_k\}$ is time-inhomogeneous, it admits a unique limiting distribution $\mu_z$ satisfying $\smash{\mu_z(\Tilde{D},s_0,a_0,s_1)=\mathbf{1}_{\{\Tilde{D}=D\}}\mu(s)\pi(a_0|s_0)p(s_1|s_0,a_0)}$, which follows from Assumption \ref{as:MC} and the strong law of large numbers for functions of Markov chains \cite{meyn2012markov}. However, since it is well known that the convergence rate from $D_k$ to $D$ is $\tilde{\mathcal{O}}(k^{-1/2})$ (measured in $\mathbb{E}[\|D_k-D\|_2]$), which is sublinear, the Markov chain $\{Z_k\}$ does not exhibit the geometric mixing property typically used in the existing study of Markovian SA \cite{srikant2019finite,bertsekas1996neuro}.

With $Z_k$ defined above, to reformulate Algorithm \ref{alg:QL-generic} as a Markovian SA, let $F:\mathbb{R}^{|\mathcal{S}||\mathcal{A}|}\times \mathcal{Z}\to \mathbb{R}^{|\mathcal{S}||\mathcal{A}|}$ be an operator defined such that given input arguments $Q\in\mathbb{R}^{|\mathcal{S}||\mathcal{A}|}$ and $z=(\smash{\Tilde{D}},s_0,a_0,s_1)\in\mathcal{Z}$, the $(s,a)$-th entry of the output of the operator is given by
    \begin{align}\label{def:operator_F}
        [F(Q,z)](s,a)=\frac{\mathbf{1}_{\{(s_0,a_0)=(s,a)\}}}{\Tilde{D}(s,a)}\big(\mathcal{R}(s_0,a_0)+\max_{a'\in\mathcal{A}}Q(s_1,a')-Q(s_0,a_0)\big)+Q(s,a).
    \end{align}
    Using the definition of $Z_k$ and $F(\cdot)$, the main update equations (\ref{alg:main1}) and (\ref{alg:main2}) from Algorithm \ref{alg:QL-generic} can be jointly written as
\begin{align}\label{eq:sa_equivalent}
    Q_{k+1}-Q_k-\alpha_k (F(Q_k,Z_k)-Q_k)\in\text{ker}(\mysp),
\end{align}
where $\alpha_k=\alpha/(k+h)$. The properties of the operator $F(\cdot)$, along with its connections to the Bellman operator $\mathcal{H}(\cdot)$ are summarized in the following lemma, whose proof is presented in Appendix \ref{pf:le:F_properties}. 

\begin{lemma}\label{le:F_properties}
The following properties hold regarding the operator $F(\cdot)$.
    \begin{enumerate}[(1)]
        \item For any $Q_1,Q_2$, $\Tilde{D}$, and $y=(s_0,a_0,s_1)\in\mathcal{Y}$, we have
        \begin{align*}
            \mysp(F(Q_1,\Tilde{D},y)-F(Q_2,\Tilde{D},y))\leq \frac{2}{\Tilde{D}(s_0,a_0)}\mysp(Q_1-Q_2).
        \end{align*}
        \item For any $Q,\Tilde{D}$, and $y=(s_0,a_0,s_1)\in\mathcal{Y}$, we have
        \begin{align*}
            \mysp(F(Q,\Tilde{D},y))\leq \frac{2}{\Tilde{D}(s_0,a_0)}(\mysp(Q)+1).
        \end{align*}
        \item For any $Q$, we have $\mathbb{E}_{Y\sim \nu}[F(Q,D,Y)]=\mathcal{H}(Q)$.
    \end{enumerate}
\end{lemma}

Among the properties stated in Lemma \ref{le:F_properties}, Parts (1) and (2) present the Lipschitz continuity of $F(\cdot)$ and its at-most-affine growth with respect to the estimated $Q$-function. However, we point out that the Lipschitz constant is inherently random in our analysis, as it depends on the input argument $\tilde{D}$, which corresponds to the empirical frequency matrix $D_k$. Lemma \ref{le:F_properties} (3) further states that the operator $F(\cdot)$ is an asymptotically unbiased estimator of the Bellman operator $\mathcal{H}(\cdot)$, thereby justifying that Equation (\ref{eq:sa_equivalent}) represents an SA algorithm for solving the Bellman equation (\ref{eq:seminorm-FPE}). 

Since the Bellman operator $\mathcal{H}(\cdot)$ will also frequently appear in our analysis, we summarize its properties in the following lemma. See Appendix \ref{pf:le:H_properties} for the proof.

\begin{lemma}\label{le:H_properties}
The following properties hold regarding the Bellman operator $\mathcal{H}(\cdot)$.
\begin{enumerate}[(1)]
    \item For any $Q_1,Q_2$, we have $\mysp(\mathcal{H}(Q_1)-\mathcal{H}(Q_2))\leq \mysp(Q_1-Q_2)$.
        \item For any $Q$, we have $\mysp(\mathcal{H}(Q))\leq \mysp(Q)+1$.
\end{enumerate}
\end{lemma}
\begin{remark}
    The proof of Lemma \ref{le:H_properties} also implies $\|\mathcal{H}(Q_1)-\mathcal{H}(Q_2)\|_\infty\leq \|Q_1-Q_2\|_\infty$ for any $Q_1$ and $Q_2$. This suggests that a possible direction for relaxing the seminorm contraction mapping assumption (cf. Assumption \ref{as:seminorm-contraction}) is to consider (time-inhomogeneous) Markovian SA under non-expansive operators. We will revisit this direction in more detail in Section \ref{sec:conclusion}.
\end{remark}

\subsection{A Lyapunov Framework for the Analysis}\label{subsec:Lyapunov}
After reformulating Algorithm \ref{alg:QL-generic} in the form of Equation (\ref{eq:sa_equivalent}), we will use a Lyapunov-drift approach to perform the finite-time analysis. Inspired by \cite{chen2024lyapunov,zhang2024prelimit,chen2025non,chandak2025finite}, we construct the Lyapunov function as the generalized Moreau envelope, defined as the informal convolution between the square of the span seminorm and the square of the $\ell_q$-norm: $M_{q,\theta}(Q) := \min_{u \in \mathbb{R}^{|\mathcal{S}||\mathcal{A}|}} \{\frac{1}{2}\mysp^2(u) + \frac{1}{2\theta}\|Q - u\|_q^2\}$ for all $Q\in\mathbb{R}^{|\mathcal{S}||\mathcal{A}|}$,
where both $q\geq 1$ and $\theta>0$ are tunable parameters yet to be chosen. For simplicity of notation, we will write $M(\cdot)$ for $M_{q,\theta}(\cdot)$ throughout the rest of the proof. Properties of this type of Lyapunov function have been thoroughly investigated in \cite{chen2024lyapunov, chen2025non}, and are restated in the following lemma for the special case of the span seminorm, for completeness. Let $\ell_q=(|\mathcal{S}||\mathcal{A}|)^{-1/q}$ and $u_q=1$. Note that we have $\ell_q\|Q\|_q\leq \|Q\|_\infty\leq u_q\|Q\|_q$ for any $Q$.

\begin{lemma}[Proposition 4.1 from \cite{chen2025non}]\label{le:Moreau}
The following properties hold:
\begin{enumerate}[(1)]
\item The function $M(\cdot)$ is convex, differentiable, and satisfies
\begin{align*}
    M(Q_2) \leq M(Q_1) + \langle \nabla M(Q_1), Q_2 - Q_1 \rangle + \frac{L}{2} \mysp(Q_2 - Q_1)^2, \quad \forall\, Q_1, Q_2 \in \mathbb{R}^d,
\end{align*}
where $L = (q-1)/(\ell_q^2 \theta)$.  
\item There exists a seminorm $p_m(\cdot)$, which satisfies $p_m(Q) = \min_{c \in \mathbb{R}} \|Q - c e\|_m$ for some norm $\|\cdot\|_m$, such that $M(Q) = p_m(Q)^2/2$. 
\item It holds that $\ell_m p_m(Q) \leq \mysp(Q) \leq u_m p_m(Q)$ for all $Q \in \mathbb{R}^{|\mathcal{S}||\mathcal{A}|}$, where $\ell_m = (1 + \theta \ell_q^2)^{1/2}$ and $u_m = (1 + \theta u_q^2)^{1/2}$.
\item It holds for all $Q \in \mathbb{R}^{|\mathcal{S}||\mathcal{A}|}$ and $c \in \mathbb{R}$ that $\langle \nabla M(Q), c e \rangle = 0$.
\item It holds for all $Q_1, Q_2, Q_3 \in \mathbb{R}^{|\mathcal{S}||\mathcal{A}|}$ that $\langle \nabla M(Q_1) - \nabla M(Q_2), Q_3 \rangle \leq L \mysp(Q_1 - Q_2) \mysp(Q_3)$.
\end{enumerate}
\end{lemma}

Lemma \ref{le:Moreau} establishes several key properties of $M(\cdot)$. Specifically, Part (1) states that $M(\cdot)$ is a smooth function with respect to the span seminorm. Part (2) states that $M(\cdot)$ itself can be written as the square of a seminorm with kernel space $\{c e \mid c \in \mathbb{R}\}$. Part (3) states that $M(\cdot)$ can approximate the seminorm-square function $\mysp(Q)^2/2$ arbitrarily closely, since $\lim_{\theta \rightarrow 0} u_m/\ell_m = 1$. This property, together with Part (1), implies that $M(Q)$ is a smooth approximation of $\mysp(Q^2)/2$. Part (4) states that the gradient of $M(\cdot)$ is always orthogonal to $\text{ker}(\mysp)$. Finally, Part (5) follows as a consequence of Parts (1) and (4).

Using the smoothness of $M(\cdot)$ (cf. Lemma \ref{le:Moreau} (1)) together with the reformulated update equation (\ref{eq:sa_equivalent}), we have for all $k\geq 1$ that
\begin{align}
    \!\!\!\mathbb{E}[M(Q_{k+1}\!-\!Q^*)]
    \leq  \,&\mathbb{E}[M(Q_k-Q^*)]\!+\!\mathbb{E}[\langle \nabla M(Q_k-Q^*),Q_{k+1}-Q_k\rangle]\!+\!\frac{L}{2}\mathbb{E}[\mysp(Q_{k+1}-Q_k)^2]\nonumber\\
    =\,&\mathbb{E}[M(Q_k-Q^*)]\!+\!\alpha_k\mathbb{E}[\langle \nabla M(Q_k-Q^*),F(Q_k,Z_k)-Q_k\rangle]\nonumber\\
    &+\frac{L\alpha_k^2}{2}\mathbb{E}[\mysp(F(Q_k,Z_k)-Q_k)^2]\nonumber\\
    =\,&  \mathbb{E}[M(Q_k-Q^*)]+\alpha_k \underbrace{\mathbb{E}[\langle\nabla M(Q_k-Q^*),\mathcal{H}(Q_k)-Q_k\rangle]}_{:=T_1}\nonumber\\
    &+\alpha_k  \underbrace{\mathbb{E}[\langle\nabla M(Q_k-Q^*),F(Q_k,D,Y_k)-\mathcal{H}(Q_k)\rangle]}_{:=T_2}\nonumber\\
    &+\alpha_k  \underbrace{\mathbb{E}[\langle\nabla M(Q_k-Q^*),F(Q_k,D_k,Y_k)-F(Q_k,D,Y_k)\rangle]}_{:=T_3}\nonumber\\
    &+\frac{L\alpha_k^2}{2}\underbrace{\mathbb{E}[\mysp( F(Q_k,Z_k)-Q_k)^2]}_{:=T_4},\label{eq:decomposition_illustration}
\end{align}
where $Y_k=(S_k,A_k,S_{k+1})$ and $D$ is the $|\mathcal{S}||\mathcal{A}|$ by $|\mathcal{S}||\mathcal{A}|$ diagonal matrix with diagonal entries $\{\mu(s)\pi(a|s)\}_{(s,a)\in\mathcal{S}\times \mathcal{A}}$. Recall that $\{Y_k\}$ forms a time-homogeneous Markov chain, with state space denoted by $\mathcal{Y}$. Moreover, under Assumption \ref{as:MC}, the Markov chain $\{Y_k\}$ admits a unique stationary distribution $\nu \in \Delta(\mathcal{Y})$, which satisfies $\nu(s,a,s') = \mu(s)\pi(a|s)p(s'|s,a)$ for all $y=(s,a,s') \in \mathcal{Y}$.

In view of Equation (\ref{eq:decomposition_illustration}), it remains to bound the terms $T_1$, $T_2$, $T_3$, and $T_4$. We begin with the term $T_1$, which can be viewed as the ``deterministic'' part of Algorithm \ref{alg:QL-generic} because $F(\cdot, Z_k)$ is an asymptotically unbiased estimator of $\mathcal{H}(\cdot)$. We show in the following lemma that the term $T_1$ provides a negative drift. The proof of Lemma \ref{le:T_1} is presented in Appendix \ref{pf:le:T_1}.

\begin{lemma}\label{le:T_1}
    It holds for all $k\geq 1$ that $T_1\leq -2\phi_1 \mathbb{E}[M(Q_k - Q^*)]$,
    where $\phi_1=1-\beta u_m/\ell_m$.
\end{lemma}

Since $\lim_{\theta \rightarrow 0} u_{m}/\ell_m = 1$ (see Lemma \ref{le:Moreau} (3)) and $\beta \in (0,1)$, we can make $\phi_1$ strictly positive (thereby ensuring a negative drift) by choosing $\theta$ appropriately.

Moving to the terms $T_2$, $T_3$, and $T_4$ in Equation (\ref{eq:decomposition_illustration}), the term $T_2$ accounts for the error due to the Markovian noise $\{Y_k\}$, the term $T_3$ accounts for the error in estimating the matrix $D$ using the empirical frequency matrix $D_k$, and the term $T_4$ arises due to the fact that Equation (\ref{eq:sa_equivalent}) is a discrete-time algorithm. To show that all of them are dominated by the negative drift provided by the term $T_1$, the analysis is different from and significantly more challenging than the existing literature on Markovian SA. Specifically, the main challenge in bounding these terms lies in handling the correlation among the iterate $Q_k$, the empirical frequency matrix $D_k$, and the Markov chain $\{Y_k\}$. In the existing literature, if the underlying noise sequence is i.i.d. or forms a uniformly ergodic Markov chain, such correlations can be handled using either an approach based on conditioning and mixing time \cite{srikant2019finite} or a more recently developed approach based on the Poisson equation \cite{chandak2023concentration,haque2023tight,haque2024stochastic,nanda2025Qlearning}. Unfortunately, neither approach is applicable here due to the fact that $\{Z_k = (D_k, Y_k)\}$ is time-inhomogeneous and lacks geometric mixing.

\subsection{Breaking the Correlation}
Our approach to breaking the correlation relies on a combination of almost-sure time-varying bounds, conditioning arguments, and concentration inequalities for Markov chains. To illustrate this approach, we use the term $T_4$ as an example. Before proceeding, we note that among the error terms $T_2$, $T_3$, and $T_4$, the term $T_4$ is the easiest to handle. Specifically, since the term $T_4$ is multiplied by $\alpha_k^2$ in Equation (\ref{eq:decomposition_illustration}), it suffices to show that $T_4 = \tilde{\mathcal{O}}(1)$ for it to be dominated by the negative drift (cf. Lemma \ref{le:T_1}). In contrast, for the terms $T_2$ and $T_3$, we need to establish that they are $o(1)$, which is more challenging.

According to the definition of $F(\cdot)$ in Equation (\ref{def:operator_F}), the vector $F(Q_k,Z_k)-Q_k$ has only one non-zero entry, i.e., the $(S_k,A_k)$-th one. Therefore, we have by the definition of $\mysp(\cdot)$ that
\begin{align}
    T_4=\,&\mathbb{E}[\mysp(F(Q_k,D_k,Y_k)-Q_k)^2]\nonumber\\
    =\,&\frac{1}{4}\mathbb{E}\left[\left(\frac{1}{D_k(S_k,A_k)}\right)^2\left(\mathcal{R}(S_k,A_k)+\max_{a'\in\mathcal{A}}Q_k(S_{k+1},a')-Q_k(S_k,A_k)\right)^2\right]\nonumber\\
    \leq \,&\frac{1}{4}\mathbb{E}\left[\left(\frac{1}{D_k(S_k,A_k)}\right)^2\left(|\mathcal{R}(S_k,A_k)|+\left|\max_{a'\in\mathcal{A}}Q_k(S_{k+1},a')-Q_k(S_k,A_k)\right|\right)^2\right]\nonumber\\
    \leq \,&\frac{1}{4}\mathbb{E}\left[\left(\frac{1}{D_k(S_k,A_k)}\right)^2\left(1+2\mysp(Q_k)\right)^2\right].\label{eq:T4-illustration}
\end{align}
To proceed, the immediate challenge we face is that the random variable $D_k(S_k, A_k)$, which represents the frequency of visiting the state-action pair $(S_k, A_k)$ in the first $k$ time steps, is strongly correlated with $Q_k$.

\paragraph{Step One: Time-Varying Almost-Sure Bounds.}

The first step of our approach to break the correlation between $D_k(S_k,A_k)$ and $Q_k$ is to show that the iterates $Q_k$, while not uniformly bounded by a constant (which is a key difficulty compared to the discounted counterpart \cite{gosavi2006boundedness}), admit a {time-varying} almost-sure bound. Specifically, we have the following proposition.

\begin{proposition}\label{prop:log-growth}
The following inequality holds a.s. for all $k\geq 1$: $\mysp(Q_k)\leq b_k$,
where $b_k=\mysp(Q_1)+\alpha|\mathcal{S}||\mathcal{A}|\log (\frac{\lceil (k-1) / (|\mathcal{S}||\mathcal{A}|) \rceil+h}{h})$.
\end{proposition}

\begin{remark}
    Note that the proof of Theorem \ref{thm:QL-set} for $k\in \{1,2,\cdots,K-1\}$ is complete, since Proposition \ref{prop:log-growth} implies $\mysp(Q_k - Q^*) \leq \mysp(Q_k) + \mysp(Q^*) \leq b_k + \mysp(Q^*)$ a.s. for all $k$.
\end{remark}

Proposition \ref{prop:log-growth} is a non-trivial observation, as the bound holds independent of the randomness in the sample trajectory and grows logarithmically in $k$. These two features together enable us to decouple the iterate $Q_k$ and the time-inhomogeneous Markovian noise $Z_k$. The proof of Proposition \ref{prop:log-growth} (presented in Appendix \ref{pf:prop:log-growth}) relies on a combination of induction and a combinatorial argument.

Apply the almost-sure time-varying bound from  Proposition \ref{prop:log-growth} to Equation (\ref{eq:T4-illustration}), we have
\begin{align}
    T_4
    \leq \,&\frac{1}{4}\mathbb{E}\left[\left(\frac{1}{D_k(S_k,A_k)}\right)^2\left(1+2\mysp(Q_k)\right)^2\right]\nonumber\\
    \leq \,&\left(1+b_k\right)^2\mathbb{E}\left[\frac{1}{D_k(S_k,A_k)^2}\right]\nonumber\\
    = \,&\left(1+b_k\right)^2\sum_{s,a}\mathbb{E}\left[\frac{\mathbf{1}_{\{(S_k,A_k)=(s,a)\}}}{D_k(s,a)^2}\right].\label{eq:T4-illustration1}
\end{align}

\paragraph{Step Two: A Conditioning Argument.}
In view of Equation (\ref{eq:T4-illustration1}), it remains to bound the quantity $\mathbb{E}[\mathbf{1}_{\{(S_k,A_k) = (s,a)\}} / D_k(s,a)^2]$ for any $(s,a)$. The immediate challenge lies in handling the correlation between the empirical frequency $D_k(s,a)$ and the indicator function $\mathbf{1}_{\{(S_k,A_k) = (s,a)\}}$. To address this issue, we apply a conditioning argument, which is inspired by \cite{srikant2019finite}.
Specifically, since $D_k(s,a)=(N_k(s,a)+h)/(k+h)$, we have for any $(s,a)$ and $\tilde{\tau}\leq k-1$ that
\begin{align}
    \mathbb{E}\left[\frac{\mathbf{1}_{\{(S_k,A_k)=(s,a)\}}}{D_k(s,a)^2}\right]
    =\,&\mathbb{E}\left[\frac{\mathbf{1}_{\{(s,a)=(S_k,A_k)\}}(k+h)^2}{(N_k(s,a)+h)^2}\right]\nonumber\\
    =\,&\mathbb{E}\left[\frac{\mathbf{1}_{\{(s,a)=(S_k,A_k)\}}(k+h)^2}{(N_{k-1}(s,a)+1+h)^2}\right]\tag{The pair $(S_k,A_k)$ is visited at time step $k$.}\nonumber\\
    \leq \,&\mathbb{E}\left[\frac{\mathbf{1}_{\{(s,a)=(S_k,A_k)\}}(k+h)^2}{(N_{k-\tilde{\tau}-1}(s,a)+1+h)^2}\right]\tag{$N_k(s,a)$ is an increasing function of $k$.}\nonumber\\
    =\,&\mathbb{E}\left[\mathbb{P}(S_k=s,A_k=a\mid  S_{k-\tilde{\tau}-1},A_{k-\tilde{\tau}-1})\frac{(k+h)^2}{(N_{k-\tilde{\tau}-1}(s,a)+1+h)^2}\right],\label{eq:T4-illustration2}
\end{align}
where the last equality follows from the tower property and the Markov property.

Under Assumption \ref{as:MC}, the Markov chain $\{(S_k,A_k)\}$ also enjoys geometric mixing, which implies
\begin{align}
    \mathbb{P}(S_k=s,A_k=a| S_{k-\tilde{\tau}-1},A_{k-\tilde{\tau}-1})
    \leq \,&|\mathbb{P}(S_k=s,A_k=a| S_{k-\tilde{\tau}-1},A_{k-\tilde{\tau}-1})-D(s,a)|+D(s,a)\nonumber\\
    \leq \,&2C\rho^{\tilde{\tau}}+D(s,a)\nonumber\\
    \leq \,&2D(s,a),\label{eq:T4-illustration3}
\end{align}
where the last line follows from choosing $\tilde{\tau}=\min\{t:C\rho^t\leq D_{\min}\}$. See Appendix \ref{pf:le:T4} for more details. Combining Equations \eqref{eq:T4-illustration1}, \eqref{eq:T4-illustration2}, and \eqref{eq:T4-illustration3}, we have
\begin{align}
    T_4\leq 2\left(1+b_k\right)^2\sum_{s,a}D(s,a)\mathbb{E}\left[\frac{(k+h)^2}{(N_{k-\tilde{\tau}-1}(s,a)+1+h)^2}\right].\label{eq:T4-illustration4}
\end{align}

\paragraph{Step Three: Markov Chain Concentration.}
To proceed from Equation (\ref{eq:T4-illustration4}), the last challenge we face here is that the random variable $N_{k-\tilde{\tau}-1}(s,a)$ appears in the denominator of the fraction, which breaks the linearity. We overcome this challenge by using Markov chain concentration inequalities. 

For simplicity of notation, denote $\bar{D}_{k-\tilde{\tau}-1}(s,a)=N_{k-\tilde{\tau}-1}(s,a)/(k-\tilde{\tau}-1)$. Given $\delta\in (0,1)$, for any $(s,a)$, let $E_\delta(s,a)=\{|\bar{D}_{k-\tilde{\tau}-1}(s,a)-D(s,a)|\leq \delta D(s,a)\}$
and let $E_\delta^c(s,a)$ be the complement of event $E_\delta(s,a)$. Note that on the event $E_\delta(s,a)$, we have $|\bar{D}_{k-\tilde{\tau}-1}(s,a)-D(s,a)|\leq \delta D(s,a)$, which implies $\bar{D}_{k-\tilde{\tau}-1}(s,a)\geq (1-\delta)D(s,a)$,
while on the event $E_\delta^c(s,a)$, we have the trivial bound $\bar{D}_{k-\tilde{\tau}-1}(s,a)\geq 0$. Therefore, we obtain
\begin{align}
    \mathbb{E}\left[\frac{(k+h)^2}{(N_{k-\tilde{\tau}-1}(s,a)+1+h)^2}\right]
    =\,&\mathbb{E}\left[\frac{(k+h)^2}{((k-\tilde{\tau}-1)\bar{D}_{k-\tilde{\tau}-1}(s,a)+1+h)^2}\right]\nonumber\\
    =\,&(k+h)^2\mathbb{E}\left[\frac{\mathbf{1}_{\{E_\delta(s,a)\}}+\mathbf{1}_{\{E_\delta^c(s,a)\}}
    }{((k-\tilde{\tau}-1)\bar{D}_{k-\tilde{\tau}-1}(s,a)+1+h)^2}\right]\nonumber\\
    \leq \,&\frac{(k+h)^2}{(k-\tilde{\tau}-1)^2}\frac{1}{(1-\delta)^2D(s,a)^2}+\frac{(k+h)^2}{(h+1)^2}\mathbb{P}(E_\delta^c(s,a)).\label{eq:T4-illustration5}
\end{align}
To bound $\mathbb{P}(E_\delta^c(s,a))$, we use the following Markov chain concentration inequality, which is a consequence of \cite[Corollary 2.11]{paulin2015concentration}. The proof of Lemma \ref{le:MC_concentration} is presented in Appendix \ref{pf:le:MC_concentration}.

\begin{lemma}\label{le:MC_concentration}
    Let $\{X_k\}_{k\geq 1}$ be a finite, irreducible, and aperiodic Markov chain taking values in $\mathcal{X}=\{1,2,3,\cdots,n\}$. Denote its unique stationary distribution by $\nu\in\Delta(\mathcal{X})$. Let $\tilde{C}\geq 1$ and $\tilde{\rho}\in (0,1)$ be such that $\max_{x\in\mathcal{X}}\|p(X_k=\cdot\mid X_1=x)-\nu(\cdot)\|_{\text{TV}}\leq \tilde{C}\tilde{\rho}^{k-1}$ for all $k\geq 1$. Then, there exists $c>0$ (which depends on the mixing time of the Markov chain) such that the following inequality holds for all $\epsilon\geq \frac{4\tilde{C}}{(1-\tilde{\rho})k}$:
    \begin{align*}
        \mathbb{P}\left(|\hat{\nu}_k(x)-\nu(x)|\geq \epsilon \right)\leq 2\exp(-ck\epsilon^2),\quad \forall\,x\in \mathcal{X}.
    \end{align*}
    where $\hat{\nu}_k(x)=\sum_{j=1}^{k}\mathbf{1}_{\{X_j=x\}}/k$.
\end{lemma}

Apply Lemma \ref{le:MC_concentration} and we have $\mathbb{P}(E_\delta^c(s,a))\leq 2\exp(-c_{mc} (k-\tilde{\tau}-1)\delta^2D(s,a)^2)$
for any $\delta\geq 4C/[D_{\min}(1-\rho)(k-\tilde{\tau}-1)]$, where $c_{mc}>0$ is a constant depending on the mixing time of the Markov chain $\{(S_k,A_k)\}$.
Combining the previous inequality with Equation (\ref{eq:T4-illustration5}) yields
\begin{align*}
    \mathbb{E}\left[\frac{(k+h)^2}{(N_{k-\tilde{\tau}-1}(s,a)+1+h)^2}\right]
    \leq \,&\frac{(k+h)^2}{(k-\tilde{\tau}-1)^2}\frac{1}{(1-\delta)^2D(s,a)^2}\\
    &+\frac{2(k+h)^2}{(h+1)^2}\exp(-c_{mc} (k-\tilde{\tau}-1)\delta^2D(s,a)^2)\\
    \leq \,&\frac{3}{D(s,a)^2},
    \end{align*}
    where the last line follows from (1) choosing $\delta$ properly based on $k$, and (2) when $k$ is large enough. See Appendix \ref{pf:le:T4} for more details. 
    
    Finally, using the previous inequality in Equation (\ref{eq:T4-illustration4}), we have the following lemma, which shows that $T_4=\tilde{\mathcal{O}}(1)$. A more detailed proof of Lemma \ref{le:T4} (following the road map described above) is presented in Appendix \ref{pf:le:T4}. 
    
 \begin{lemma}\label{le:T4}
The following inequality holds for all $k\geq K$:
   \begin{align*}
       T_4\leq \frac{18|\mathcal{S}||\mathcal{A}|(b_k+\mysp(Q^*)+1)^2}{D_{\min}}.
   \end{align*}
\end{lemma}   

Following similar ideas, specifically, combining the time-varying almost-sure bounds (cf. Proposition \ref{prop:log-growth}) with conditioning arguments and Markov chain concentration results (cf. Lemma \ref{le:MC_concentration}), we are able to bound the terms $T_2$ and $T_3$ on the right-hand side of Equation (\ref{eq:decomposition_illustration}). The results are presented in the following two propositions.

\begin{proposition}\label{prop:T2}
    The following inequality holds for all $k\geq K$:
    \begin{align*}
        T_2\leq \frac{28\tau_k L|\mathcal{S}||\mathcal{A}|(b_k+\mysp(Q^*)+1)^2}{D_{\min}^2}\alpha_k.
    \end{align*}
\end{proposition}

\begin{proposition}\label{prop:T3}
    The following inequality holds for all $k\geq K$:
    \begin{align*}
        T_3\leq\phi_1\mathbb{E}\left[M(Q_k-Q^*)\right]+\frac{32C|\mathcal{S}||\mathcal{A}|(b_k+\mysp(Q^*)+1)^2}{\ell_m^2\phi_1(1-\rho)\alpha D_{\min}^2}\alpha_k.
    \end{align*}
\end{proposition}

The proofs of Propositions \ref{prop:T2} and \ref{prop:T3} are presented in Appendices \ref{pf:prop:T2} and \ref{pf:prop:T3}, respectively. Using the bounds we obtained for the terms $T_1$, $T_2$, $T_3$, and $T_4$ altogether in Equation (\ref{eq:decomposition_illustration}), we obtain the following result.

\begin{proposition}\label{prop:recursion}
	The following inequality holds for all $k\geq K$:
	\begin{align}
		\mathbb{E}[M(Q_{k+1}-Q^*)]
		\leq (1-\phi_1\alpha_k)\mathbb{E}[M(Q_k-Q^*)]+35\phi_2\tau_k(b_k+\mysp(Q^*)+1)^2\alpha_k^2.\label{eq:recursion}
	\end{align}
    where
    \begin{align*}
        \phi_2 = \frac{C|\mathcal{S}||\mathcal{A}|}{(1 - \rho)\ell_m^2\phi_1D_{\min}^2\alpha} + \frac{L|\mathcal{S}||\mathcal{A}|}{D_{\min}^2}.
    \end{align*}
\end{proposition}

Note that Equation (\ref{eq:recursion}) is a one-step Lyapunov drift inequality of the desired form: it exhibits a negative drift with an additive error that is order-wise smaller than the magnitude of the drift.

\subsection{Solving the Recursion}
Repeatedly using Equation (\ref{eq:recursion}) from Proposition \ref{prop:recursion}, we obtain a finite-time bound on $\mathbb{E}[M(Q_k-Q^*)]$. The final steps are:
\begin{enumerate}[(1)]
    \item translating the bound on $\mathbb{E}[M(Q_k - Q^*)]$ into a bound on $\mathbb{E}[\mysp(Q_k - Q^*)^2]$ using Lemma \ref{le:Moreau},
    \item substituting $\alpha_k = \alpha / (k + h)$ to make the bound explicit in $k$,
    \item selecting the parameters $\theta$ and $q$ (introduced in the definition of the Lyapunov function $M(\cdot)$) to make all constants explicit.
\end{enumerate}
The details are presented in Appendix \ref{ap:recursion}. The proof of Theorem \ref{thm:QL-set} (for Algorithm \ref{alg:QL-generic}) is now complete after finishing these three steps.

\section{Discussion on Assumption \ref{as:seminorm-contraction}}\label{sec:discussion}

Although this work focuses on the setting where the Bellman operator is a span-seminorm contraction mapping, it is of interest to investigate how to conduct a finite-time analysis of $Q$-learning without this assumption. Unlike the discounted setting, the algorithm design and analysis in the average-reward case crucially depend on the structural properties of the underlying MDP (e.g., multichain, weakly communicating, communicating, unichain, recurrent, and $\mysp(\cdot)$-contractive)~\cite{puterman2014markov}. A pictorial illustration of the relationships among these models is provided in~\cite[Figure~8.3.1]{puterman2014markov}. As the first study establishing convergence rates for average-reward $Q$-learning, our Assumption~\ref{as:seminorm-contraction} is admittedly among the strongest. In this section, we illustrate the main challenges and outline potential approaches for relaxing the $\mysp(\cdot)$-contraction assumption. In particular, we will consider the unichain model, which generalizes the recurrent model, and the weakly communicating model, which generalizes the communicating model.

\paragraph{Unichain MDPs.} An MDP is called \emph{unichain} if, for any $\pi \in \Pi_d$ (where $\Pi_d$ denotes the set of all deterministic policies), the induced Markov chain consists of a single recurrent class, possibly accompanied by a set of transient states \cite{levin2017markov}. Without loss of generality, one can further assume that the induced Markov chain is aperiodic within the recurrent class by applying a simple data transformation, as detailed in~\cite[Section~8.5.4]{puterman2014markov}. 

Under the unichain and aperiodicity assumptions, it has been shown that the Bellman operator admits a multi-step contraction property with respect to the span seminorm $\mysp(\cdot)$ \cite{puterman2014markov}. Specifically, for any $Q \in \mathbb{R}^{|\mathcal{S}||\mathcal{A}|}$, let $\pi_Q$ denote the policy greedily induced by $Q$, i.e., $\pi_Q(s) = \arg\max_{a \in \mathcal{A}} Q(s,a)$ for all $s \in \mathcal{S}$. Then, there exist a positive integer $J$ and a constant $\beta \in (0,1)$ such that 
\begin{align*}
    \mysp\left([\mathcal{H}^{\pi_{Q_1}}]^J(Q_1) - [\mathcal{H}^{\pi_{Q_2}}]^J(Q_2)\right) 
    \leq \beta\, \mysp(Q_1 - Q_2), 
    \quad \forall\, Q_1, Q_2 \in \mathbb{R}^{|\mathcal{S}||\mathcal{A}|}.
\end{align*}
Note that $[\mathcal{H}^{\pi_Q}]^J(Q) \neq \mathcal{H}^J(Q)$ since the policy is fixed to be $\pi_Q$ during the $J$ applications of the operator. This multi-step contraction property has been leveraged in \cite{zhang2021finite,chen2025non} to study the convergence rate of synchronous $Q$-learning, where access to a generative model enables the construction of a conditionally unbiased estimator of the multi-step operator $[\mathcal{H}^{\pi_{Q_k}}]^J(Q_k)$. However, when only a single trajectory of Markovian samples generated by a behavior policy is available, as in this work, it is unclear how to construct such an estimator. If an appropriate estimator for the multi-step operator can be constructed, then our analysis can be readily extended to the unichain setting.

\paragraph{Weakly Communicating MDPs.} An MDP is called \emph{weakly communicating} if its state space $\mathcal{S}$ can be partitioned into two disjoint subsets $\mathcal{S}_1$ and $\mathcal{S}_2$, where all states in $\mathcal{S}_1$ are transient under any stationary policy, and within $\mathcal{S}_2$, every state is reachable from every other state under some stationary policy. In this case, the Bellman operator is not known to satisfy any contraction property with respect to any seminorm. However, since the Bellman operator is always non-expansive with respect to $\|\cdot\|_\infty$ (see our discussion after Lemma~\ref{le:H_properties}), one can model $Q$-learning as a non-expansive SA with time-inhomogeneous Markovian noise induced by the adaptive stepsizes. 

The convergence rate of general SA with non-expansive operators has been studied in the literature for both martingale-difference noise~\cite{bravo2024stochastic} and Markovian noise~\cite{blaser2024asymptotic}, where a rate of $\mathcal{O}(k^{-1/10})$ was established for the expected residual $\mathbb{E}\big[\|\mathcal{H}(Q_k)-Q_k\|_\infty\big]$. We envision that, by combining our techniques for handling adaptive stepsizes (including time-varying almost-sure bounds, conditioning arguments, and Markov chain concentration) with those developed in~\cite{bravo2024stochastic,blaser2024asymptotic} for analyzing non-expansive SA, our results can be extended to weakly communicating MDPs. However, whether an $\mathcal{O}(1/\sqrt{k})$ rate for the expected residual (or equivalently, an $\mathcal{O}(\epsilon^{-2})$ sample complexity) can be achieved remains an open question.

\section{Conclusion}\label{sec:conclusion}
In this work, we present the first study on the last-iterate convergence rates of average-reward $Q$-learning under an asynchronous implementation. We investigate two algorithms—one exhibiting set convergence and the other pointwise convergence—both of which achieve an $\tilde{\mathcal{O}}(1/k)$ convergence rate in the mean-square sense. Moreover, we show that the key to attaining this rate lies in the use of adaptive stepsizes, which can be interpreted as a form of implicit importance sampling that compensates for asynchronous updates. We believe that our analysis is broadly applicable to the study of general SA algorithms with adaptive stepsizes. 

Regarding future work, we identify two promising directions: 
\begin{enumerate}[(1)]
    \item The sample complexity dependence on the size of the state--action space and the seminorm contraction factor established in this work is generally not tight, in light of the existing lower bound in~\cite{jin2021towards} (although this lower bound is developed in the generative model setting, whereas our setting involves the more challenging Markovian sampling regime). Developing refined analyses or improved algorithmic designs—such as incorporating Polyak averaging or variance reduction—to tighten these bounds is an interesting direction for future research.
    \item As discussed in Section~\ref{sec:discussion}, Assumption~\ref{as:seminorm-contraction} is a strong structural assumption on the underlying MDP. Relaxing this assumption to encompass more general MDP models, e.g., unichain, communicating, and weakly communicating MDPs, is an immediate direction for further investigation. The associated challenges and envisioned approaches are detailed in Section~\ref{sec:discussion}. 
\end{enumerate}

\section*{Acknowledgement}
We would like to thank Mr. Shaan Ul Haque from Georgia Tech for his help in identifying the seminorm contraction property of the asynchronous Bellman operator presented in Lemma \ref{le:ABO-contraction}. We would also like to thank Dr. Siva Theja Maguluri from Georgia Tech for his valuable feedback on an early draft of this work.

\bibliographystyle{apalike}
\bibliography{references}

\begin{thebibliography}{}

\bibitem[Abounadi et~al., 2001]{abounadi2001learning}
Abounadi, J., Bertsekas, D., and Borkar, V.~S. (2001).
\newblock Learning algorithms for {Markov} decision processes with average
  cost.
\newblock {\em SIAM Journal on Control and Optimization}, 40(3):681--698.

\bibitem[Abounadi et~al., 2002]{abounadi2002stochastic}
Abounadi, J., Bertsekas, D.~P., and Borkar, V. (2002).
\newblock Stochastic approximation for nonexpansive maps: Application to
  {$Q$}-learning algorithms.
\newblock {\em SIAM Journal on Control and Optimization}, 41(1):1--22.

\bibitem[Agrawal and Agrawal, 2025]{agrawal2024optimistic}
Agrawal, P. and Agrawal, S. (2025).
\newblock Optimistic {Q}-learning for average reward and episodic reinforcement
  learning.
\newblock In {\em The Thirty Eighth Annual Conference on Learning Theory},
  pages 1--1. PMLR.

\bibitem[Azar et~al., 2012]{azar2012dynamic}
Azar, M.~G., G{\'o}mez, V., and Kappen, H.~J. (2012).
\newblock Dynamic policy programming.
\newblock {\em The Journal of Machine Learning Research}, 13(1):3207--3245.

\bibitem[Beck and Srikant, 2012]{beck2012error}
Beck, C.~L. and Srikant, R. (2012).
\newblock Error bounds for constant stepsize {$Q$}-learning.
\newblock {\em Systems \& control letters}, 61(12):1203--1208.

\bibitem[Beck and Srikant, 2013]{beck2013improved}
Beck, C.~L. and Srikant, R. (2013).
\newblock Improved upper bounds on the expected error in constant stepsize
  {$Q$}-learning.
\newblock In {\em 2013 American Control Conference}, pages 1926--1931. IEEE.

\bibitem[Benveniste et~al., 2012]{benveniste2012adaptive}
Benveniste, A., M{\'e}tivier, M., and Priouret, P. (2012).
\newblock {\em Adaptive Algorithms and Stochastic Approximations}, volume~22.
\newblock Springer Science \& Business Media.

\bibitem[Bertsekas and Tsitsiklis, 1996]{bertsekas1996neuro}
Bertsekas, D.~P. and Tsitsiklis, J.~N. (1996).
\newblock {\em Neuro-Dynamic Programming}.
\newblock Athena Scientific.

\bibitem[Bhandari et~al., 2018]{bhandari2018finite}
Bhandari, J., Russo, D., and Singal, R. (2018).
\newblock A finite-time analysis of temporal difference learning with linear
  function approximation.
\newblock In {\em Conference on learning theory}, pages 1691--1692. PMLR.

\bibitem[Blaser and Zhang, 2024]{blaser2024asymptotic}
Blaser, E. and Zhang, S. (2024).
\newblock Asymptotic and finite sample analysis of nonexpansive stochastic
  approximations with {Markovian} noise.
\newblock {\em Preprint arXiv:2409.19546}.

\bibitem[Borkar, 1998]{borkar1998asynchronous}
Borkar, V.~S. (1998).
\newblock Asynchronous stochastic approximations.
\newblock {\em SIAM Journal on Control and Optimization}, 36(3):840--851.

\bibitem[Borkar, 2008]{borkar_SA}
Borkar, V.~S. (2008).
\newblock {\em {Stochastic Approximation: A Dynamical Systems Viewpoint}}.
\newblock Cambridge University Press.

\bibitem[Borkar and Meyn, 2000]{borkar2000ode}
Borkar, V.~S. and Meyn, S.~P. (2000).
\newblock The {ODE} method for convergence of stochastic approximation and
  reinforcement learning.
\newblock {\em SIAM Journal on Control and Optimization}, 38(2):447--469.

\bibitem[Bottou et~al., 2018]{bottou2018optimization}
Bottou, L., Curtis, F.~E., and Nocedal, J. (2018).
\newblock Optimization methods for large-scale machine learning.
\newblock {\em Siam Review}, 60(2):223--311.

\bibitem[Bravo and Cominetti, 2024]{bravo2024stochastic}
Bravo, M. and Cominetti, R. (2024).
\newblock Stochastic fixed-point iterations for nonexpansive maps: Convergence
  and error bounds.
\newblock {\em SIAM Journal on Control and Optimization}, 62(1):191--219.

\bibitem[Brown et~al., 2020]{brown2020language}
Brown, T., Mann, B., Ryder, N., Subbiah, M., Kaplan, J.~D., Dhariwal, P.,
  Neelakantan, A., Shyam, P., Sastry, G., Askell, A., et~al. (2020).
\newblock Language models are few-shot learners.
\newblock {\em Advances in neural information processing systems},
  33:1877--1901.

\bibitem[Bucklew and Bucklew, 2004]{bucklew2004introduction}
Bucklew, J.~A. and Bucklew, J. (2004).
\newblock {\em Introduction to Rare Event Simulation}, volume~5.
\newblock Springer.

\bibitem[Chandak and Borkar, 2023]{chandak2023concentration}
Chandak, S. and Borkar, V.~S. (2023).
\newblock A concentration bound for {TD(0)} with function approximation.
\newblock {\em Preprint arXiv:2312.10424}.

\bibitem[Chandak et~al., 2025]{chandak2025finite}
Chandak, S., Haque, S.~U., and Bambos, N. (2025).
\newblock Finite-time bounds for two-timescale stochastic approximation with
  arbitrary norm contractions and {Markovian} noise.
\newblock {\em Preprint arXiv:2503.18391}.

\bibitem[Chen et~al., 2023]{chen2023target}
Chen, Z., Clarke, J.-P., and Maguluri, S.~T. (2023).
\newblock Target network and truncation overcome the deadly triad in
  {$Q$}-learning.
\newblock {\em SIAM Journal on Mathematics of Data Science}, 5(4):1078--1101.

\bibitem[Chen et~al., 2020]{chen2020finite}
Chen, Z., Maguluri, S.~T., Shakkottai, S., and Shanmugam, K. (2020).
\newblock Finite-sample analysis of contractive stochastic approximation using
  smooth convex envelopes.
\newblock {\em Advances in Neural Information Processing Systems}, 33.

\bibitem[Chen et~al., 2024]{chen2024lyapunov}
Chen, Z., Maguluri, S.~T., Shakkottai, S., and Shanmugam, K. (2024).
\newblock {A Lyapunov theory for finite-sample guarantees of Markovian
  stochastic approximation}.
\newblock {\em Operations Research}, 72(4):1352--1367.

\bibitem[Chen et~al., 2025a]{chen2025concentration}
Chen, Z., Maguluri, S.~T., and Zubeldia, M. (2025a).
\newblock Concentration of contractive stochastic approximation: Additive and
  multiplicative noise.
\newblock {\em The Annals of Applied Probability}, 35(2):1298--1352.

\bibitem[Chen et~al., 2022]{chen2019finitesample}
Chen, Z., Zhang, S., Doan, T.~T., Clarke, J.-P., and Maguluri, S.~T. (2022).
\newblock Finite-sample analysis of nonlinear stochastic approximation with
  applications in reinforcement learning.
\newblock {\em Automatica}, 146:110623.

\bibitem[Chen et~al., 2025b]{chen2025non}
Chen, Z., Zhang, S., Zhang, Z., Haque, S.~U., and Maguluri, S.~T. (2025b).
\newblock A non-asymptotic theory of seminorm {Lyapunov} stability: From
  deterministic to stochastic iterative algorithms.
\newblock {\em Preprint arXiv:2502.14208}.

\bibitem[Dai and Gluzman, 2022]{dai2022queueing}
Dai, J.~G. and Gluzman, M. (2022).
\newblock Queueing network controls via deep reinforcement learning.
\newblock {\em Stochastic Systems}, 12(1):30--67.

\bibitem[Devraj and Meyn, 2017]{devraj2017zap}
Devraj, A.~M. and Meyn, S. (2017).
\newblock Zap {$Q$}-learning.
\newblock In {\em Advances in Neural Information Processing Systems}, pages
  2235--2244.

\bibitem[Durmus et~al., 2021]{durmus2021tight}
Durmus, A., Moulines, E., Naumov, A., Samsonov, S., Scaman, K., and Wai, H.-T.
  (2021).
\newblock Tight high probability bounds for linear stochastic approximation
  with fixed stepsize.
\newblock {\em Advances in Neural Information Processing Systems},
  34:30063--30074.

\bibitem[Espeholt et~al., 2018]{espeholt2018impala}
Espeholt, L., Soyer, H., Munos, R., Simonyan, K., Mnih, V., Ward, T., Doron,
  Y., Firoiu, V., Harley, T., Dunning, I., et~al. (2018).
\newblock {IMPALA: Scalable Distributed Deep-RL with Importance Weighted
  Actor-Learner Architectures}.
\newblock In {\em International Conference on Machine Learning}, pages
  1407--1416.

\bibitem[Even-Dar et~al., 2003]{even2003learning}
Even-Dar, E., Mansour, Y., and Bartlett, P. (2003).
\newblock Learning rates for {$Q$}-learning.
\newblock {\em Journal of machine learning Research}, 5(1).

\bibitem[Gosavi, 2006]{gosavi2006boundedness}
Gosavi, A. (2006).
\newblock Boundedness of iterates in {$Q$}-learning.
\newblock {\em Systems \& control letters}, 55(4):347--349.

\bibitem[Haque et~al., 2023]{haque2023tight}
Haque, S.~U., Khodadadian, S., and Maguluri, S.~T. (2023).
\newblock Tight finite time bounds of two-time-scale linear stochastic
  approximation with {Markovian} noise.
\newblock {\em Preprint arXiv:2401.00364}.

\bibitem[Haque and Maguluri, 2024]{haque2024stochastic}
Haque, S.~U. and Maguluri, S.~T. (2024).
\newblock Stochastic approximation with unbounded {Markovian} noise: A
  general-purpose theorem.
\newblock {\em Preprint arXiv:2410.21704}.

\bibitem[Harchol-Balter, 2013]{harchol2013performance}
Harchol-Balter, M. (2013).
\newblock {\em Performance Modeling and Design of Computer Systems: Queueing
  Theory in Action}.
\newblock Cambridge University Press.

\bibitem[Huo et~al., 2023]{huo2023bias}
Huo, D., Chen, Y., and Xie, Q. (2023).
\newblock Bias and extrapolation in {Markovian} linear stochastic approximation
  with constant stepsizes.
\newblock In {\em Abstract Proceedings of the 2023 ACM SIGMETRICS International
  Conference on Measurement and Modeling of Computer Systems}, pages 81--82.

\bibitem[Jaakkola et~al., 1994]{jaakkola1994convergence}
Jaakkola, T., Jordan, M.~I., and Singh, S.~P. (1994).
\newblock Convergence of stochastic iterative dynamic programming algorithms.
\newblock In {\em Advances in neural information processing systems}, pages
  703--710.

\bibitem[Jaksch et~al., 2010]{jaksch2010near}
Jaksch, T., Ortner, R., and Auer, P. (2010).
\newblock Near-optimal regret bounds for reinforcement learning.
\newblock {\em Journal of Machine Learning Research}, 11(Apr):1563--1600.

\bibitem[Jin et~al., 2024]{jin2024feasible}
Jin, Y., Gummadi, R., Zhou, Z., and Blanchet, J. (2024).
\newblock Feasible {$Q$}-learning for average reward reinforcement learning.
\newblock In {\em International Conference on Artificial Intelligence and
  Statistics}, pages 1630--1638. PMLR.

\bibitem[Jin and Sidford, 2021]{jin2021towards}
Jin, Y. and Sidford, A. (2021).
\newblock Towards tight bounds on the sample complexity of average-reward
  {MDPs}.
\newblock In {\em International Conference on Machine Learning}, pages
  5055--5064. PMLR.

\bibitem[Kara and Yuksel, 2023]{kara2023q}
Kara, A.~D. and Yuksel, S. (2023).
\newblock {$Q$-learning for continuous state and action MDPs under average cost
  criteria}.
\newblock {\em Preprint arXiv:2308.07591}.

\bibitem[Kloek and Van~Dijk, 1978]{kloek1978bayesian}
Kloek, T. and Van~Dijk, H.~K. (1978).
\newblock Bayesian estimates of equation system parameters: An application of
  integration by {Monte Carlo}.
\newblock {\em Econometrica: Journal of the Econometric Society}, pages 1--19.

\bibitem[Kushner and Clark, 2012]{kushner2012stochastic}
Kushner, H.~J. and Clark, D.~S. (2012).
\newblock {\em Stochastic Approximation Methods for Constrained and
  Unconstrained Systems}, volume~26.
\newblock Springer Science \& Business Media.

\bibitem[Lan, 2020]{lan2020first}
Lan, G. (2020).
\newblock {\em First-Order and Stochastic Optimization Methods for Machine
  Learning}.
\newblock Springer.

\bibitem[Lauand and Meyn, 2024]{lauand2024revisiting}
Lauand, C.~K. and Meyn, S. (2024).
\newblock Revisiting stepsize assumptions in stochastic approximation.
\newblock {\em Preprint arXiv:2405.17834}.

\bibitem[Lee, 2024]{lee2024final}
Lee, D. (2024).
\newblock Final iteration convergence bound of {$Q$}-learning: Switching system
  approach.
\newblock {\em IEEE Transactions on Automatic Control}, 69(7):4765--4772.

\bibitem[Levin and Peres, 2017]{levin2017markov}
Levin, D.~A. and Peres, Y. (2017).
\newblock {\em Markov Chains and Mixing Times}, volume 107.
\newblock American Mathematical Soc.

\bibitem[Li et~al., 2021]{li2021tightening}
Li, G., Cai, C., Chen, Y., Gu, Y., Wei, Y., and Chi, Y. (2021).
\newblock Tightening the dependence on horizon in the sample complexity of
  {$Q$}-learning.
\newblock In {\em International Conference on Machine Learning}, pages
  6296--6306. PMLR.

\bibitem[Li et~al., 2024a]{li2024q}
Li, G., Cai, C., Chen, Y., Wei, Y., and Chi, Y. (2024a).
\newblock {Is $Q$-learning minimax optimal? a tight sample complexity
  analysis}.
\newblock {\em Operations Research}, 72(1):222--236.

\bibitem[Li et~al., 2020]{li2020sample}
Li, G., Wei, Y., Chi, Y., Gu, Y., and Chen, Y. (2020).
\newblock {Sample complexity of asynchronous $Q$-learning: Sharper analysis and
  variance reduction}.
\newblock {\em Advances in neural information processing systems},
  33:7031--7043.

\bibitem[Li et~al., 2024b]{li2024stochastic}
Li, T., Wu, F., and Lan, G. (2024b).
\newblock Stochastic first-order methods for average-reward {Markov} decision
  processes.
\newblock {\em Mathematics of Operations Research}.

\bibitem[Li et~al., 2023]{li2023statistical}
Li, X., Yang, W., Liang, J., Zhang, Z., and Jordan, M.~I. (2023).
\newblock {A statistical analysis of Polyak-Ruppert averaged $Q$-learning}.
\newblock In {\em International Conference on Artificial Intelligence and
  Statistics}, pages 2207--2261. PMLR.

\bibitem[Liu et~al., 2022]{liu2022rl}
Liu, B., Xie, Q., and Modiano, E. (2022).
\newblock {RL-QN: A reinforcement learning framework for optimal control of
  queueing systems}.
\newblock {\em ACM Transactions on Modeling and Performance Evaluation of
  Computing Systems}, 7(1):1--35.

\bibitem[Mahadevan, 1996]{mahadevan1996average}
Mahadevan, S. (1996).
\newblock Average reward reinforcement learning: Foundations, algorithms, and
  empirical results.
\newblock {\em Machine learning}, 22(1):159--195.

\bibitem[Melo et~al., 2008]{melo2008analysis}
Melo, F.~S., Meyn, S.~P., and Ribeiro, M.~I. (2008).
\newblock An analysis of reinforcement learning with function approximation.
\newblock In {\em Proceedings of the 25th international conference on Machine
  learning}, pages 664--671.

\bibitem[Meyn, 2024]{meyn2024projected}
Meyn, S. (2024).
\newblock The projected {Bellman} equation in reinforcement learning.
\newblock {\em IEEE Transactions on Automatic Control}.

\bibitem[Meyn and Tweedie, 2012]{meyn2012markov}
Meyn, S.~P. and Tweedie, R.~L. (2012).
\newblock {\em Markov Chains and Stochastic Stability}.
\newblock Springer Science \& Business Media.

\bibitem[Mnih et~al., 2015]{mnih2015human}
Mnih, V., Kavukcuoglu, K., Silver, D., Rusu, A.~A., Veness, J., Bellemare,
  M.~G., Graves, A., Riedmiller, M., Fidjeland, A.~K., Ostrovski, G., et~al.
  (2015).
\newblock Human-level control through deep reinforcement learning.
\newblock {\em nature}, 518(7540):529--533.

\bibitem[Mou et~al., 2020]{mou2020linear}
Mou, W., Li, C.~J., Wainwright, M.~J., Bartlett, P.~L., and Jordan, M.~I.
  (2020).
\newblock On linear stochastic approximation: Fine-grained {Polyak-Ruppert} and
  non-asymptotic concentration.
\newblock In {\em Conference on Learning Theory}, pages 2947--2997. PMLR.

\bibitem[Mou et~al., 2024]{mou2021optimal}
Mou, W., Pananjady, A., Wainwright, M.~J., and Bartlett, P.~L. (2024).
\newblock Optimal and instance-dependent guarantees for {Markovian} linear
  stochastic approximation.
\newblock {\em Mathematical Statistics and Learning}, 7(1):41--153.

\bibitem[Nanda and Chen, 2025]{nanda2025Qlearning}
Nanda, P. and Chen, Z. (2025).
\newblock A minimal-assumption analysis of {Q}-learning with time-varying
  policies.
\newblock {\em Preprint arXiv:1910.02140}.

\bibitem[Paulin, 2015]{paulin2015concentration}
Paulin, D. (2015).
\newblock Concentration inequalities for {Markov} chains by {Marton} couplings
  and spectral methods.
\newblock {\em Electronic Journal of Probability}, 20:1--32.

\bibitem[Puterman, 2014]{puterman2014markov}
Puterman, M.~L. (2014).
\newblock {\em {Markov Decision Processes: Discrete Stochastic Dynamic
  Programming}}.
\newblock John Wiley \& Sons.

\bibitem[Qian et~al., 2024]{qian2024almost}
Qian, X., Xie, Z., Liu, X., and Zhang, S. (2024).
\newblock Almost sure convergence rates and concentration of stochastic
  approximation and reinforcement learning with {Markovian} noise.
\newblock {\em Preprint arXiv:2411.13711}.

\bibitem[Qu and Wierman, 2020]{qu2020finite}
Qu, G. and Wierman, A. (2020).
\newblock Finite-time analysis of asynchronous stochastic approximation and
  {$Q$}-learning.
\newblock In {\em Conference on Learning Theory}, pages 3185--3205. PMLR.

\bibitem[Robbins and Monro, 1951]{robbins1951stochastic}
Robbins, H. and Monro, S. (1951).
\newblock A stochastic approximation method.
\newblock {\em The Annals of Mathematical Statistics}, pages 400--407.

\bibitem[Shalev-Shwartz et~al., 2012]{shalev2012online}
Shalev-Shwartz, S. et~al. (2012).
\newblock Online learning and online convex optimization.
\newblock {\em Foundations and Trends{\textregistered} in Machine Learning},
  4(2):107--194.

\bibitem[Silver et~al., 2018]{silver2018general}
Silver, D., Hubert, T., Schrittwieser, J., Antonoglou, I., Lai, M., Guez, A.,
  Lanctot, M., Sifre, L., Kumaran, D., Graepel, T., et~al. (2018).
\newblock {A general reinforcement learning algorithm that masters chess,
  shogi, and Go through self-play}.
\newblock {\em Science}, 362(6419):1140--1144.

\bibitem[Singh et~al., 2022]{singh2022reinforcement}
Singh, B., Kumar, R., and Singh, V.~P. (2022).
\newblock Reinforcement learning in robotic applications: a comprehensive
  survey.
\newblock {\em Artificial Intelligence Review}, 55(2):945--990.

\bibitem[Srikant and Ying, 2019]{srikant2019finite}
Srikant, R. and Ying, L. (2019).
\newblock Finite-time error bounds for linear stochastic approximation and
  {TD}-learning.
\newblock In {\em Conference on Learning Theory}, pages 2803--2830.

\bibitem[Suttle et~al., 2021]{suttle2021reinforcement}
Suttle, W., Zhang, K., Yang, Z., Liu, J., and Kraemer, D. (2021).
\newblock Reinforcement learning for cost-aware {Markov} decision processes.
\newblock In {\em International Conference on Machine Learning}, pages
  9989--9999. PMLR.

\bibitem[Sutton and Barto, 2018]{sutton2018reinforcement}
Sutton, R.~S. and Barto, A.~G. (2018).
\newblock {\em {Reinforcement Learning: An Introduction}}.
\newblock MIT press.

\bibitem[Tsitsiklis, 1994]{tsitsiklis1994asynchronous}
Tsitsiklis, J.~N. (1994).
\newblock Asynchronous stochastic approximation and {$Q$}-learning.
\newblock {\em Machine learning}, 16(3):185--202.

\bibitem[Tsitsiklis and Van~Roy, 1999]{tsitsiklis1999average}
Tsitsiklis, J.~N. and Van~Roy, B. (1999).
\newblock Average cost temporal-difference learning.
\newblock {\em Automatica}, 35(11):1799--1808.

\bibitem[Wainwright, 2019a]{wainwright2019stochastic}
Wainwright, M.~J. (2019a).
\newblock Stochastic approximation with cone-contractive operators: Sharp
  $\ell_\infty$-bounds for {$Q$}-learning.
\newblock {\em Preprint arXiv:1905.06265}.

\bibitem[Wainwright, 2019b]{wainwright2019variance}
Wainwright, M.~J. (2019b).
\newblock Variance-reduced {$Q$}-learning is minimax optimal.
\newblock {\em Preprint arXiv:1906.04697}.

\bibitem[Wan et~al., 2021a]{wan2021average}
Wan, Y., Naik, A., and Sutton, R. (2021a).
\newblock Average-reward learning and planning with options.
\newblock {\em Advances in Neural Information Processing Systems},
  34:22758--22769.

\bibitem[Wan et~al., 2021b]{wan2021learning}
Wan, Y., Naik, A., and Sutton, R.~S. (2021b).
\newblock Learning and planning in average-reward {Markov} decision processes.
\newblock In {\em International Conference on Machine Learning}, pages
  10653--10662. PMLR.

\bibitem[Wan et~al., 2024]{wan2024convergence}
Wan, Y., Yu, H., and Sutton, R.~S. (2024).
\newblock On convergence of average-reward {$Q$}-learning in weakly
  communicating {Markov} decision processes.
\newblock {\em Preprint arXiv:2408.16262}.

\bibitem[Wang et~al., 2024]{wangoptimal}
Wang, S., Blanchet, J., and Glynn, P. (2024).
\newblock Optimal sample complexity for average reward {M}arkov decision
  processes.
\newblock In {\em The Twelfth International Conference on Learning
  Representations}.

\bibitem[Watkins and Dayan, 1992]{watkins1992q}
Watkins, C.~J. and Dayan, P. (1992).
\newblock {$Q$}-learning.
\newblock {\em Machine learning}, 8(3-4):279--292.

\bibitem[Wei et~al., 2020]{wei2020model}
Wei, C.-Y., Jahromi, M.~J., Luo, H., Sharma, H., and Jain, R. (2020).
\newblock {Model-free reinforcement learning in infinite-horizon average-reward
  Markov decision processes}.
\newblock In {\em International conference on machine learning}, pages
  10170--10180. PMLR.

\bibitem[Yang et~al., 2024]{yang2024relative}
Yang, X., Hu, J., and Hu, J.-Q. (2024).
\newblock Relative {$Q$}-learning for average-reward {Markov} decision
  processes with continuous states.
\newblock {\em IEEE Transactions on Automatic Control}.

\bibitem[Yu et~al., 2024]{yu2024asynchronous}
Yu, H., Wan, Y., and Sutton, R.~S. (2024).
\newblock Asynchronous stochastic approximation and average-reward
  reinforcement learning.
\newblock {\em Preprint arXiv:2409.03915}.

\bibitem[Zhang et~al., 2021a]{zhang2021breaking}
Zhang, S., Yao, H., and Whiteson, S. (2021a).
\newblock Breaking the deadly triad with a target network.
\newblock In {\em International Conference on Machine Learning}, pages
  12621--12631. PMLR.

\bibitem[Zhang et~al., 2021b]{zhang2021finite}
Zhang, S., Zhang, Z., and Maguluri, S.~T. (2021b).
\newblock Finite sample analysis of average-reward {TD}-learning and
  {$Q$}-learning.
\newblock {\em Advances in Neural Information Processing Systems},
  34:1230--1242.

\bibitem[Zhang et~al., 2024]{zhang2024prelimit}
Zhang, Y., Huo, D., Chen, Y., and Xie, Q. (2024).
\newblock Prelimit coupling and steady-state convergence of constant-stepsize
  nonsmooth contractive {SA}.
\newblock In {\em Abstracts of the 2024 ACM SIGMETRICS/IFIP PERFORMANCE Joint
  International Conference on Measurement and Modeling of Computer Systems},
  pages 35--36.

\bibitem[Zhang and Xie, 2024]{zhang2024constant}
Zhang, Y. and Xie, Q. (2024).
\newblock Constant stepsize {$Q$}-learning: Distributional convergence, bias,
  and extrapolation.
\newblock {\em Preprint arXiv:2401.13884}.

\bibitem[Zhang and Xie, 2023]{zhang2023sharper}
Zhang, Z. and Xie, Q. (2023).
\newblock {Sharper model-free reinforcement learning for average-reward Markov
  decision processes}.
\newblock In {\em The Thirty Sixth Annual Conference on Learning Theory}, pages
  5476--5477. PMLR.

\end{thebibliography}

\addtocontents{toc}{\protect\setcounter{tocdepth}{-1}}
\newpage
\addtocontents{toc}{\protect\setcounter{tocdepth}{-1}}
\appendix
\begin{center}
    {\LARGE\bfseries Appendices}
\end{center}

\section{Proofs of All Technical Results in Sections \ref{sec:preliminaries}, \ref{sec:QL-set}, \ref{sec:QL-pointwise}, and \ref{sec:viewpoint}}
\subsection{Proof of Lemma \ref{le:span_sup_projection}}\label{pf:le:span_sup_projection}
We first verify that when $c=(\max_i x_i + \min_j x_j)/2$, we have $\|x-ce\|_\infty=(\max_i x_i - \min_j x_j)/2$. On the one hand, for any $i\in \{1,2,\cdots,d\}$, we have
\begin{align*}
    |x_i-c|\leq  \max\left(\;\left|\max_ix_i-c\right|,\left|\min_jx_j-c\right|\;\right)
    =\frac{\max_i x_i - \min_j x_j}{2},
\end{align*}
which implies
\begin{align*}
    \|x-ce\|_\infty\leq \frac{\max_i x_i - \min_j x_j}{2}.
\end{align*}
On the other hand, letting $i_1=\argmax_{i\in \{1,2,\cdots,d\}}x_i$, we have
\begin{align*}
    \|x-ce\|_\infty\geq |x_{i_1}-c|
    =\frac{\max_i x_i - \min_j x_j}{2}.
\end{align*}
Together, they imply 
\begin{align*}
    \|x-ce\|_\infty=\frac{\max_i x_i - \min_j x_j}{2}=\mysp(x).
\end{align*}

To complete the proof, it remains to show that for any $c'\neq c$, we have $\|x-c'e\|_\infty>\|x-ce\|_\infty$.
Assume without loss of generality that $c'<c=(\max_i x_i + \min_j x_j)/2$. Then, we have
\begin{align*}
    \|x-c'e\|_\infty\geq\,& |x_{i_1}-c'e|\\
    =\,&\max_ix_i-c'\\
    > \,&\max_ix_i-\frac{\max_i x_i + \min_j x_j}{2},\\
    =\,&\frac{\max_i x_i - \min_j x_j}{2}.
\end{align*}

\subsection{Proof of Lemma \ref{le:solution_same}}\label{ap:le:solution_same}

    For simplicity of notation, denote
\begin{align*}
    \mathcal{Q}_1=\{Q \mid \mathcal{H}(Q) - Q = r^* e\},\quad \text{and}\quad \mathcal{Q}_2=\{Q \mid \mysp(\mathcal{H}(Q) - Q) = 0\}.
\end{align*}
We will show that $\mathcal{Q}\subseteq \mathcal{Q}_1\subseteq \mathcal{Q}_2\subseteq \mathcal{Q}$, which would imply $\mathcal{Q}=\mathcal{Q}_1=\mathcal{Q}_2$.

For any $Q\in \mathcal{Q}$, there exists $c\in\mathbb{R}$ such that $Q=Q^*+ce$. Since 
\begin{align*}
    \mathcal{H}(Q)-Q=\mathcal{H}(Q^*+ce)-Q^*-ce=\mathcal{H}(Q^*)-Q^*=r^*e,
\end{align*}
we have $Q\in\mathcal{Q}_1$, implying $\mathcal{Q}\subseteq \mathcal{Q}_1$.

For any $Q\in\mathcal{Q}_1$, since $\mathcal{H}(Q)-Q=r^*e$, we have
$\mysp(\mathcal{H}(Q)-Q)=\mysp(r^*e)=0$. As a result, we have $Q\in \mathcal{Q}_2$, implying $\mathcal{Q}_1\subseteq \mathcal{Q}_2$.

Finally, to show that $\mathcal{Q}_2\subseteq \mathcal{Q}$, it is enough to show that for any $Q\in\mathcal{Q}_2$, we must have $Q - Q^* \in \text{ker}(\mysp)$. To see this, since $Q\in\mathcal{Q}_2$, there exists $c \in \mathbb{R}$ such that $\mathcal{H}(Q) - Q = c e$. It follows that
\begin{align*}
    \mysp(Q - Q^*) &= \mysp(\mathcal{H}(Q) - c e - \mathcal{H}(Q^*) + r^* e) \\
    &= \mysp(\mathcal{H}(Q) - \mathcal{H}(Q^*)) \\
    &\leq \beta\, \mysp(Q - Q^*),
\end{align*}
where the last line follows from Assumption \ref{as:seminorm-contraction}. Therefore, we must have $\mysp(Q - Q^*) = 0$, implying $Q - Q^* \in \text{ker}(\mysp)$.

\subsection{Proof of Corollary \ref{co:sc}}\label{pf:co:sc}
By Jensen's inequality, we have
\begin{align*}
    \mathbb{E}[\mysp(Q_k-Q^*)^2]\leq \epsilon^2\quad \Longrightarrow\quad \mathbb{E}[\mysp(Q_k-Q^*)]\leq \epsilon.
\end{align*}
To achieve $\mathbb{E}[\mysp(Q_k-Q^*)^2]\leq \epsilon^2$, using Theorem \ref{thm:QL-set} with the case $\alpha(1-\beta)=2$, we have
\begin{align*}
    k=\tilde{\mathcal{O}}\left(\frac{|\mathcal{S}|^3|\mathcal{A}|^3\mysp(Q^*)^2}{D_{\min}^2(1-\beta)^3(1-\rho)\epsilon^2}\right).
\end{align*}
To bound $\mysp(Q^*)$ in terms of $\beta$, using the Bellman equation, we have
\begin{align*}
    \mysp(Q^*)=\,&\mysp(\mathcal{H}(Q^*)-r^*e)\\
    =\,&\mysp(\mathcal{H}(Q^*)-\mathcal{H}(0))+\mysp(\mathcal{H}(0))\\
    \leq \,&\beta \mysp(Q^*)+1,
\end{align*}
where the last inequality follows from Lemma \ref{le:H_properties}. Rearranging terms, we obtain $\mysp(Q^*)\leq 1/(1-\beta)$.
Therefore, the overall sample complexity is
\begin{align*}
    \tilde{\mathcal{O}}\left(\frac{|\mathcal{S}|^3|\mathcal{A}|^3}{D_{\min}^2(1-\beta)^5(1-\rho)\epsilon^2}\right).
\end{align*}

\subsection{Proof of Lemma \ref{le:span_sup_equivalence}}\label{pf:le:span_sup_equivalence}
The first inequality always holds because
\begin{align*}
    \mysp(x-y)=\min_{c\in\mathbb{R}}\|x-y-ce\|_\infty\leq \|x-y\|_\infty.
\end{align*}
To prove the second inequality, 
let $c=[\max_i(x_i-y_i)+\min_i(x_i-y_i)]/2$. Then, we have
\begin{align}
    \|x-y\|_\infty=\,&\|x-y-ce+ce\|_\infty\nonumber\\
    \leq\,& \|x-y-ce\|_\infty+|c|\nonumber\\
    =\,&\mysp(x-y)+|c|\label{eq1:span_to_sup}.
\end{align}
To bound $|c|$, observe that for any $i\in \{1,2,\cdots,d\}$, we have
\begin{align*}
   x_i-y_i=\,&x_i-y_i-c+c\leq \|x-y-ce\|_\infty+c=\mysp(x-y)+c,\\
   x_i-y_i=\,&x_i-y_i-c+c\geq  -\|x-y-ce\|_\infty+c=-\mysp(x-y)+c
\end{align*}
It follows that
\begin{align*}
    y_i-\mysp(x-y)+c\leq x_i\leq y_i+\mysp(x-y)+c,\quad \forall\, i\in \{1,2,\cdots,d\}.
\end{align*}
The previous inequality implies:
\begin{align*}
    \max_iy_i-\mysp(x-y)+c\leq\,& \max_ix_i\leq \max_iy_i+\mysp(x-y)+c,\\
    \min_jy_j-\mysp(x-y)+c\leq \,&\min_jx_j\leq \min_jy_j+\mysp(x-y)+c.
\end{align*}
Summing up the previous two inequalities and then rearranging terms, we obtain
\begin{align*}
    c\leq\,& \mysp(x-y)+\frac{\max_ix_i+\min_jx_j}{2}-\frac{\max_iy_i+\min_jy_j}{2},\\
    c\geq\,& -\mysp(x-y)+\frac{\max_ix_i+\min_jx_j}{2}-\frac{\max_iy_i+\min_jy_j}{2}.
\end{align*}
To proceed, since 
\begin{align*}
    \|x\|_\infty=\max_i|x_i|=\max\left(\max_ix_i,-\min_jx_j\right)=\mysp(x)=\frac{\max_ix_i-\min_jx_j}{2},
\end{align*}
we must have $\max_ix_i+\min_jx_j=0$.
Similarly, we also have $\max_iy_i+\min_jy_j=0$.
It follows that 
\begin{align*}
    -\mysp(x-y)\leq  c\leq \mysp(x-y)\quad \Longleftrightarrow\quad|c|\leq \mysp(x-y).
\end{align*}
Combining the previous inequality with Equation (\ref{eq1:span_to_sup}) finishes the proof.

\subsection{Proof of Lemma \ref{le:ABO-contraction}}\label{pf:le:ABO-contraction}

Given any $Q_1,Q_2\in\mathbb{R}^{|\mathcal{S}||\mathcal{A}|}$, for simplicity of notation, denote
\begin{align*}
	M=\,&\max_{s,a}([\mathcal{H}(Q_1)](s,a)-[\mathcal{H}(Q_2)](s,a)),\quad &
	m=\,&\min_{s,a}([\mathcal{H}(Q_1)](s,a)-[\mathcal{H}(Q_2)](s,a)),\\
	L=\,&\max_{s,a}(Q_1(s,a)-Q_1(s,a)),\quad &
	\ell=\,&\min_{s,a}(Q_1(s,a)-Q_1(s,a)).
\end{align*}
To show the seminorm contraction property of the asynchronous Bellman operator $\bar{\mathcal{H}}(\cdot)$, we first show that $M\leq L$ and $m\geq \ell$.
By definition of the Bellman operator $\mathcal{H}(\cdot)$, we have
\begin{align*}
	M
	=\,& \max_{s,a} \sum_{s'\in S} p(s'| s,a) \left(\max_{a'} Q_1(s', a') - \max_{a''} Q_2(s', a'') \right) \\
	\leq\,& \max_{s,a} \sum_{s'\in S} p(s'| s,a) \max_{a'} (Q_1(s',a') - Q_2(s',a')) \\
	\leq\,& \max_{s,a}(Q_1(s,a) - Q_2(s,a))\\
	=\,&L.
\end{align*}
Similarly, we also have
\begin{align*}
	m 
	=\,& \min_{s,a} \sum_{s'\in S} p(s'| s,a) \left(\max_{a'} Q_1(s', a') - \max_{a''} Q_2(s', a'') \right) \\
	\geq\,& \min_{s,a} \sum_{s'\in S} p(s'| s,a) \min_{a'} (Q_1(s', a') - Q_2(s', a')) \\
	\geq \,&\min_{s,a} (Q_1(s,a) - Q_2(s,a))\\
	=\,&\ell.
\end{align*}
Now that we have established $M\leq L$ and $m\geq \ell$, to proceed with the proof, using the definition of $\bar{\mathcal{H}}(\cdot)$, we have
\begin{align*}
	&\mysp(\bar{\mathcal{H}}(Q_1)-\bar{\mathcal{H}}(Q_2))\\
	=\,&\mysp((I-D)(Q_1-Q_2)+D(\mathcal{H}(Q_1)-\mathcal{H}(Q_2)))\\
	=\,&\frac{1}{2}\max_{s,a}\left[(1-D(s,a))(Q_1(s,a)\!-\!Q_2(s,a))\!+\!D(s,a)([\mathcal{H}(Q_1)](s,a)\!-\![\mathcal{H}(Q_2)](s,a))\right]\\
	&-\frac{1}{2}\min_{s,a}\left[(1\!-\!D(s,a))(Q_1(s,a)\!-\!Q_2(s,a))\!+\!D(s,a)([\mathcal{H}(Q_1)](s,a)\!-\![\mathcal{H}(Q_2)](s,a))\right].
\end{align*}
On the one hand, we have
\begin{align*}
	&\max_{s,a}\left[(1-D(s,a))(Q_1(s,a)-Q_2(s,a))+D(s,a)([\mathcal{H}(Q_1)](s,a)-[\mathcal{H}(Q_2)](s,a))\right]\\
	\leq \,&\max_{s,a}\left[(1-D(s,a))L+D(s,a)M\right]\\
	= \,&\max_{s,a}\left[L-D(s,a)(L-M)\right]\\
	= \,&L-D_{\min}(L-M),
\end{align*}
where the last inequality follows from $M\leq L$. 

On the other hand, we have
\begin{align*}
	&\min_{s,a}\left[(1-D(s,a))(Q_1(s,a)-Q_2(s,a))+D(s,a)([\mathcal{H}(Q_1)](s,a)-[\mathcal{H}(Q_2)](s,a))\right]\\
	\geq \,&\min_{s,a}\left[(1-D(s,a))\ell+D(s,a)m\right]\\
	= \,&\min_{s,a}\left[\ell-D(s,a)(\ell-m)\right]\\
	= \,&\ell-D_{\min}(\ell-m),
\end{align*}
where the last inequality follows from $m\geq \ell$. 

Together, we obtain
\begin{align*}
	\mysp(\bar{\mathcal{H}}(Q_1)-\bar{\mathcal{H}}(Q_2))
	\leq \,&\frac{1}{2}(L-D_{\min}(L-M))-\frac{1}{2}(\ell-D_{\min}(\ell-m))\\
    =\,&\frac{1}{2}(1-D_{\min})(L-\ell)+\frac{1}{2}D_{\min}(M-m).
\end{align*}
Since $\mathcal{H}(\cdot)$ is a contraction mapping with respect to $\mysp(\cdot)$ (cf. Assumption \ref{as:seminorm-contraction}), we have
\begin{align*}
	\mysp(\mathcal{H}(Q_1)-\mathcal{H}(Q_2))=\frac{M-m}{2}\leq \beta \mysp(Q_1-Q_2)=\frac{\beta(L-\ell)}{2}.
\end{align*}
It follows that
\begin{align*}
	\mysp(\bar{\mathcal{H}}(Q_1)-\bar{\mathcal{H}}(Q_2))
	\leq \,&\frac{1}{2}(1-D_{\min})(L-\ell)+\frac{1}{2}D_{\min}(M-m)\\
	\leq \,&\frac{1}{2}(1-D_{\min})(L-\ell)+\frac{1}{2}D_{\min}\beta (L-\ell)\\
	=\,&\frac{1}{2}(1-D_{\min}(1-\beta))(L-\ell)\\
	=\,&(1-D_{\min}(1-\beta))\mysp(Q_1-Q_2).
\end{align*}
Finally, since $D_{\min}\in (0,1)$ under Assumption \ref{as:MC}, which implies $\bar{\beta}:=(1-D_{\min}(1-\beta))\in (0,1)$, the operator $\bar{\mathcal{H}}(\cdot)$ is a contraction mapping with respect to $\mysp(\cdot)$, with contraction factor $\bar{\beta}$.

\subsection{Proof of Proposition \ref{prop:universal-convergence}}\label{pf:prop:universal-convergence}
Using the definition of $\bar{\mathcal{H}}(\cdot)$, if $D=I/(|\mathcal{S}||\mathcal{A}|)$, we have 
\begin{align*}
    \mysp(\bar{\mathcal{H}}(Q)-Q)=0\quad \Longleftrightarrow\quad&\mysp(D(\mathcal{H}(Q)-Q))=0,\\
    \Longleftrightarrow\quad&\frac{1}{|\mathcal{S}||\mathcal{A}|}\mysp(\mathcal{H}(Q)-Q)=0,\\
     \Longleftrightarrow\quad& \mysp(\mathcal{H}(Q)-Q)=0,
\end{align*}
which implies
\begin{align*}
    \{Q| \mysp(\Bar{\mathcal{H}}(Q)-Q)=0\}= \{Q| \mysp(\mathcal{H}(Q)-Q)=0\}.
\end{align*}
If $D\neq I/(|\mathcal{S}||\mathcal{A}|)$, let $\tilde{Q}\in \mathcal{Q}=\{Q^*+ce| c\in\mathbb{R}\}=\{Q| \mysp(\mathcal{H}(Q)-Q)=0\}$ be arbitrary. Then, we have
\begin{align*}
    \mysp(\bar{\mathcal{H}}(\tilde{Q})-\tilde{Q})=\,&\mysp(D(\mathcal{H}(\tilde{Q})-\tilde{Q}))\\
    =\,&\mysp(Dr^*e)\\
    =\,&|r^*|\left(\max_{s,a}D(s,a)-\min_{s,a}D(s,a)\right)\\
    >\,&0.
\end{align*}
Since $\mysp(\bar{\mathcal{H}}(\tilde{Q})-\tilde{Q})\neq  0$ for any $\tilde{Q}$ in the set of solutions $\mathcal{Q}$ of the original Bellman equation (\ref{eq:seminorm-FPE}), we must have
\begin{align*}
    \{Q| \mysp(\Bar{\mathcal{H}}(Q)-Q)=0\}\cap \{Q| \mysp(\mathcal{H}(Q)-Q)=0\}=\emptyset.
\end{align*}

To establish the convergence rate of $\mathbb{E}[\mysp(Q_k-\bar{Q}^*)^2]$, we will apply \cite[Theorem 4.1]{chen2025non}. For the completeness of this work, we first restate \cite[Theorem 4.1]{chen2025non} for the special case of span seminorm contractive SA in the following.

\begin{theorem}[Theorem 4.1 from \cite{chen2025non}]\label{thm:seminorm-contractive-sa}
Consider a Markovian SA of the form
\begin{align*}
    x_{k+1}=x_k+\alpha_k(\mathcal{T}(x_k,Y_k)-x_k),\quad \forall\,k\geq 1,
\end{align*}
where $\{Y_k\}$ is a stochastic process taking values in a finite set $\mathcal{Y}$ and $\mathcal{T}:\mathbb{R}^d\times \mathcal{Y}\to \mathbb{R}^d$ is an operator.
Suppose that the following assumptions are satisfied:
\begin{enumerate}[(1)]
\item The stochastic process $\{Y_k\}$ is a uniformly ergodic Markov chain with stationary distribution $\nu$.
\item The operator $\bar{\mathcal{T}}:\mathbb{R}^d\to\mathbb{R}^d$ defined as $\bar{\mathcal{T}}(x)=\mathbb{E}_{Y\sim \nu}[\mathcal{T}(x,Y)]$ is a contraction mapping with respect to $\mysp(\cdot)$, with contraction factor $\beta'\in (0,1)$.
    \item The operator $\mathcal{T}(\cdot)$ satisfies $\mysp(\mathcal{T}(x_1,y)-\mathcal{T}(x_2,y))\leq A_1\mysp(x_1-x_2)$ and $\mysp(\mathcal{T}(0,y)\leq B_1$ for any $x_1,x_2\in\mathbb{R}^d$ and $y\in\mathcal{Y}$, where $A_1,B_1>0$ are constants.
\end{enumerate}
    Then, when using $\alpha_k=\alpha/(k+h)$ with appropriately chosen $\alpha$ and $h$, we have $\mathbb{E}[\mysp(x_k-x^*)^2]\leq \tilde{\mathcal{O}}(1/k)$, where $x^*$ is a particular solution to $\mysp(\bar{\mathcal{T}}(x)-x)=0$, which is guaranteed to exist due to the seminorm fixed-point theorem \cite[Theorem 2.1]{chen2025non}.
\end{theorem}

To apply Theorem \ref{thm:seminorm-contractive-sa} to $Q$-learning with universal stepsizes as described in Equation (\ref{algo:illustration}), we only need to verify Assumption (3) because Assumption (1) holds under Assumption \ref{as:MC} and Assumption (2) has been verified in Lemma \ref{le:ABO-contraction}.

Using the definition of $G(\cdot)$, for any $Q_1,Q_2\in\mathbb{R}^{|\mathcal{S}||\mathcal{A}|}$ and $y=(s_0,a_0,s_1)\in\mathcal{Y}$, observe that the vector $G(Q_1,y)-G(Q_2,y)$ and the vector $Q_1-Q_2$ differ by exactly one entry, namely the $(s_0,a_0)$-th one. Therefore, let $Q_{\text{diff}}\in\mathbb{R}^{|\mathcal{S}||\mathcal{A}|}$ be defined as
\begin{align*}
    Q_{\text{diff}}(s_0,a_0)=\,&[G(Q_1,y)](s_0,a_0)-[G(Q_2,y)](s_0,a_0)-(Q_1(s_0,a_0)-Q_2(s_0,a_0))\\
    =\,&\max_{a'}Q_1(s_1,a')-\max_{a'}Q_2(s_1,a')-(Q_1(s_0,a_0)-Q_2(s_0,a_0)),
\end{align*}
and $Q_{\text{diff}}(s,a)=0$ for any $(s,a)\neq (s_0,a_0)$. Then, we have
\begin{align*}
    G(Q_1,y)-G(Q_2,y)=Q_1-Q_2+Q_{\text{diff}}.
\end{align*}
By the triangle inequality, we have
\begin{align}\label{eq1:pf:prop:universal-convergence}
    \mysp(G(Q_1,y)-G(Q_2,y))\leq \mysp(Q_1-Q_2)+\mysp(Q_{\text{diff}}).
\end{align}
To bound $\mysp(Q_{\text{diff}})$, since $Q_{\text{diff}}$ has only one non-zero entry, we have by the definition of $\mysp(\cdot)$ that
\begin{align*}
    \mysp(Q_{\text{diff}})=\, &\frac{1}{2}\left|\max_{a'}Q_1(s_1,a')-\max_{a'}Q_2(s_1,a')-(Q_1(s_0,a_0)-Q_2(s_0,a_0))\right|.
\end{align*}
To further bound the absolute value on the right-hand side of the previous inequality, observe that on the one hand, we have
\begin{align*}
    &\max_{a'}Q_1(s_1,a')-\max_{a'}Q_2(s_1,a')-(Q_1(s_0,a_0)-Q_2(s_0,a_0))\\
    \leq \,&\max_{s',a'}(Q_1(s',a')-Q_2(s',a'))-(Q_1(s_0,a_0)-Q_2(s_0,a_0))\\
    \leq \,&\max_{s',a'}(Q_1(s',a')-Q_2(s',a'))-\min_{s',a'}(Q_1(s',a')-Q_2(s',a'))\\
    =\,&2\mysp(Q_1-Q_2).
\end{align*}
On the other hand, since
\begin{align*}
    \max_{a'}Q_1(s_1,a')-\max_{a'}Q_2(s_1,a')
    = \,&\max_{a'}Q_1(s_1,a')-Q_2(s_1,\underline{a})\tag{Denote $\underline{a}\in\argmax_{a'}Q_2(s_1,a')$}\\
    \geq \,&Q_1(s_1,\underline{a})-Q_2(s_1,\underline{a})\\
    \geq \,&\min_{s',a'}(Q_1(s',a')-Q_2(s',a')),
\end{align*}
we have
\begin{align*}
    &\max_{a'}Q_1(s_1,a')-\max_{a'}Q_2(s_1,a')-(Q_1(s_0,a_0)-Q_2(s_0,a_0))\\
    \geq \,&\min_{s',a'}(Q_1(s',a')-Q_2(s',a'))-(Q_1(s_0,a_0)-Q_2(s_0,a_0))\\
    \geq \,&\min_{s',a'}(Q_1(s',a')-Q_2(s',a'))-\max_{s',a'}(Q_1(s',a')-Q_2(s',a'))\\
    =\,&-2\mysp(Q_1-Q_2)
\end{align*}
It follows that
\begin{align*}
    \mysp(Q_{\text{diff}})=\, &\frac{1}{2}\left|\max_{a'}Q_1(s_1,a')-\max_{a'}Q_2(s_1,a')-(Q_1(s_0,a_0)-Q_2(s_0,a_0))\right|\\
    \leq \, &\mysp(Q_1-Q_2).
\end{align*}
Combining the previous inequality with Equation (\ref{eq1:pf:prop:universal-convergence}), we obtain
\begin{align*}
    \mysp(G(Q_1,y)-G(Q_2,y))\leq 2\mysp(Q_1-Q_2).
\end{align*}
To show that $\mysp(G(0,y))$ is uniformly bounded for all $y\in\mathcal{Y}$, observe that $G(0,y)$ has only one non-zero entry, i.e., the $(s_0,a_0)$-th one, which is equal to $\mathcal{R}(s_0,a_0)$. Therefore, using the definition of $G(\cdot)$, we have
\begin{align*}
    \max_{y=(s_0,a_0,s_1)\in\mathcal{Y}}\mysp(G(0,y))=\frac{1}{2}\max_{s_0,a_0}|\mathcal{R}(s_0,a_0)|\leq \frac{1}{2}.
\end{align*}

\section{Proofs of All Technical Results in Section \ref{sec:proof}}\label{pf:thm:main}
This section presents the detailed proofs of all lemmas and propositions in support of the proof of Theorem \ref{thm:QL-set}. We begin with a summary of notation.

For any $k \geq 1$, let $D_k$, $\bar{D}_k$, and $D$ be $|\mathcal{S}||\mathcal{A}| \times |\mathcal{S}||\mathcal{A}|$ diagonal matrices with diagonal entries $\{(N_k(s,a)+h)/(k+h)\}_{(s,a) \in \mathcal{S} \times \mathcal{A}}$, $\{N_k(s,a)/k\}_{(s,a) \in \mathcal{S} \times \mathcal{A}}$, and $\{\mu(s)\pi(a|s)\}_{(s,a)\in\mathcal{S}\times \mathcal{A}}$, respectively, where we recall that $N_k(s,a) = \sum_{i=1}^k \mathbf{1}_{\{(S_i,A_i) = (s,a)\}}$ counts the number of times the state-action pair $(s,a)$ has been visited up to iteration $k$. For simplicity, we will write $D_k(s,a)$ to denote the $(s,a)$-th diagonal entry of $D_k$; similarly for $\bar{D}_k(s,a)$ and $D(s,a)$. Let $D_{\min}=\min_{s,a}\mu(s)\pi(a|s)$, which is strictly positive under Assumption \ref{as:MC}.

Let $Z_k = (D_k, S_k, A_k, S_{k+1})$ for all $k \geq 1$. Note that $\{Z_k\}$ is a time-inhomogeneous Markov chain with state space denoted by $\mathcal{Z}$. Define $Y_k = (S_k, A_k, S_{k+1})$ for all $k \geq 1$. It is clear that $\{Y_k\}$ is also a Markov chain, with state space denoted by $\mathcal{Y}$. Moreover, under Assumption \ref{as:MC}, the Markov chain $\{Y_k\}$ admits a unique stationary distribution $\nu \in \Delta(\mathcal{Y})$, which satisfies $\nu(s,a,s') = \mu(s)\pi(a|s)p(s'|s,a)$ for all $y=(s,a,s') \in \mathcal{Y}$.

\subsection{Proof of Lemma \ref{le:F_properties}}\label{pf:le:F_properties}
 \begin{enumerate}[(1)]
    \item Let $x\in\mathbb{R}^{|\mathcal{S}||\mathcal{A}|}$ be defined as 
    \begin{align*}
        x=F(Q_1,\Tilde{D},y)-F(Q_2,\Tilde{D},y)-(Q_1-Q_2).
    \end{align*}
    It is clear by the definition of $F(\cdot)$ that
    \begin{align*}
        x(s,a)=\,&
            \frac{1}{\tilde{D}(s_0,a_0)}\left(\max_{a'\in\mathcal{A}}Q_1(s_1,a')-\max_{a'\in\mathcal{A}}Q_2(s_1,a')-Q_1(s_0,a_0)+Q_2(s_0,a_0)\right)
    \end{align*}
    if $(s,a)=(s_0,a_0)$ and $x(s,a)=0$ otherwise. 
    By the triangle inequality, we have
    \begin{align}\label{eq1:pf:le:F}
        \mysp(F(Q_1,\Tilde{D},y)-F(Q_2,\Tilde{D},y))\leq \mysp(Q_1-Q_2)+\mysp(x).
    \end{align}
    To bound $\mysp(x)$, since $x$ has only one non-zero entry, we have
    \begin{align}\label{eq1:le:F_properties}
        \mysp(x)=\frac{1}{2\tilde{D}(s_0,a_0)}\left|\max_{a'\in\mathcal{A}}Q_1(s_1,a')-\max_{a'\in\mathcal{A}}Q_2(s_1,a')-(Q_1(s_0,a_0)-Q_2(s_0,a_0))\right|.
    \end{align}
    To further bound the absolute value, observe that on the one hand, we have
    \begin{align*}
        &\max_{a'\in\mathcal{A}}Q_1(s_1,a')-\max_{a'\in\mathcal{A}}Q_2(s_1,a')-(Q_1(s_0,a_0)-Q_2(s_0,a_0))\\
        \leq \,&
        \max_{s',a'}\left(Q_1(s',a')-Q_2(s',a')\right)-\min_{s',a'}(Q_1(s',a')-Q_2(s',a'))\\
        =\,&2\mysp(Q_1-Q_2).
    \end{align*}
    On the other hand, since
    \begin{align*}
        \max_{a'}Q_1(s_1,a')-\max_{a'}Q_2(s_1,a')
        = \,&\max_{a'}Q_1(s_1,a')-Q_2(s_1,\underline{a})\tag{Denote $\underline{a}\in\argmax_{a'}Q_2(s_1,a')$}\\
        \geq \,&Q_1(s_1,\underline{a})-Q_2(s_1,\underline{a})\\
        \geq \,&\min_{s',a'}(Q_1(s',a')-Q_2(s',a')),
    \end{align*}
we have
\begin{align*}
    &\max_{a'\in\mathcal{A}}Q_1(s_1,a')-\max_{a'\in\mathcal{A}}Q_2(s_1,a')-(Q_1(s_0,a_0)-Q_2(s_0,a_0))\\
    \geq \,&\min_{s',a'}(Q_1(s',a')-Q_2(s',a'))-(Q_1(s_0,a_0)-Q_2(s_0,a_0)),\\
    \geq \,&\min_{s',a'}(Q_1(s',a')-Q_2(s',a'))-\max_{s',a'}\left(Q_1(s',a')-Q_2(s',a')\right)\\
    =\,&-2\mysp(Q_1-Q_2).
\end{align*}
It follows that
\begin{align*}
    \left|\max_{a'\in\mathcal{A}}Q_1(s_1,a')-\max_{a'\in\mathcal{A}}Q_2(s_1,a')-(Q_1(s_0,a_0)-Q_2(s_0,a_0))\right|\leq 2\mysp(Q_1-Q_2).
\end{align*}
Using the previous inequality in Equation (\ref{eq1:le:F_properties}), we obtain
\begin{align*}
    \mysp(x)\leq  \frac{1}{\tilde{D}(s_0,a_0)}\mysp(Q_1-Q_2).
\end{align*}
Combining the previous inequality with Equation (\ref{eq1:pf:le:F}), we have
\begin{align*}
    \mysp(F(Q_1,\Tilde{D},y)-F(Q_2,\Tilde{D},y))\leq\,& \mysp(Q_1-Q_2)+\mysp(x)\\
    \leq\,&\left(1+\frac{1}{\tilde{D}(s_0,a_0)}\right)\mysp(Q_1-Q_2)\\
    \leq \,&\frac{2}{\tilde{D}(s_0,a_0)}\mysp(Q_1-Q_2),
\end{align*} 
where the last inequality follows from $\tilde{D}(s,a)\in (0,1)$ for all $(s,a)$.
\item By Part (1) of this lemma, we have
\begin{align*}
    \mysp(F(Q,\tilde{D},y))\leq \,& \mysp(F(Q,\tilde{D},y)-F(0,\tilde{D},y))+\mysp(F(0,\tilde{D},y))\\
    \leq \,&\frac{2}{\Tilde{D}(s_0,a_0)}\mysp(Q)+\mysp(F(0,\tilde{D},y)).
\end{align*}
Since the vector $F(0,\tilde{D},y)$ has only one non-zero entry, i.e., the $(s_0,a_0)$-th entry, which is equal to $\mathcal{R}(s_0,a_0)/\tilde{D}(s_0,a_0)$, we have
\begin{align*}
    \mysp(F(0,\tilde{D},y))=\frac{1}{2\tilde{D}(s_0,a_0)}|\mathcal{R}(s_0,a_0)|\leq \frac{1}{2\tilde{D}(s_0,a_0)}.
\end{align*}
Combining the previous two inequalities, we have
\begin{align*}
    \mysp(F(Q,\tilde{D},y))\leq \frac{2}{\Tilde{D}(s_0,a_0)}\mysp(Q)+\frac{1}{2\tilde{D}(s_0,a_0)}
    \leq\frac{2}{\Tilde{D}(s_0,a_0)}(\mysp(Q)+1).
\end{align*}
\item For any $(s,a)$, we have
\begin{align*}
    &\mathbb{E}_{Y\sim \nu}[[F(Q,D,Y)](s,a)]\\
    =\,&\sum_{s_0,a_0,s_1}\frac{\mu(s_0)\pi(a_0|s_0)p(s_1|s_0,a_0)\mathbf{1}_{\{(s_0,a_0)=(s,a)\}}}{D(s,a)}\\
    &\times \left(\mathcal{R}(s_0,a_0)+\max_{a'\in\mathcal{A}}Q(s_1,a')-Q(s_0,a_0)\right)+Q(s,a)\\
    =\,&\sum_{s_1}\frac{D(s,a)p(s_1|s,a)}{D(s,a)}\left(\mathcal{R}(s,a)+\max_{a'\in\mathcal{A}}Q(s_1,a')-Q(s,a)\right)+Q(s,a)\\
    =\,&\sum_{s_1}p(s_1|s,a)\left(\mathcal{R}(s,a)+\max_{a'\in\mathcal{A}}Q(s_1,a')\right)\\
    =\,&[\mathcal{H}(Q)](s,a).
\end{align*}
    \end{enumerate}

\subsection{Proof of Lemma \ref{le:H_properties}}\label{pf:le:H_properties}

\begin{enumerate}[(1)]
    \item For any $Q_1,Q_2\in\mathbb{R}^{|\mathcal{S}||\mathcal{A}|}$ and $c\in\mathbb{R}$, we have
    \begin{align*}
        &|[\mathcal{H}(Q_1)](s,a)-[\mathcal{H}(Q_2)](s,a)-c|\\
        =\,&\left|\mathbb{E}\left[\max_{a'\in\mathcal{A}}(Q_1(S_2,a')-c)-\max_{a'\in\mathcal{A}}Q_2(S_2,a')\,|dle|\,S_1=s,A_1=a\right]\right|\\
        \leq \,&\mathbb{E}\left[\left|\max_{a'\in\mathcal{A}}(Q_1(S_2,a')-c)-\max_{a'\in\mathcal{A}}Q_2(S_2,a')\right|\,|dle|\,S_1=s,A_1=a\right]\\
        \leq \,&\mathbb{E}\left[\max_{a'\in\mathcal{A}}\left|Q_1(S_2,a')-Q_2(S_2,a')-c\right|\,|dle|\,S_1=s,A_1=a\right]\\
        \leq \,&\|Q_1-Q_2-ce\|_\infty.
    \end{align*}
    The above inequality implies
    \begin{align*}
        \|\mathcal{H}(Q_1)-\mathcal{H}(Q_2)-ce\|_\infty\leq \|Q_1-Q_2-ce\|_\infty,\quad \forall\,Q_1,Q_2\in\mathbb{R}^{|\mathcal{S}||\mathcal{A}|},\;c\in\mathbb{R}.
    \end{align*}
    As a side note, by setting $c=0$ in the above inequality, we have
    \begin{align*}
        \|\mathcal{H}(Q_1)-\mathcal{H}(Q_2)\|_\infty\leq \|Q_1-Q_2\|_\infty,\quad \forall\,Q_1,Q_2\in\mathbb{R}^{|\mathcal{S}||\mathcal{A}|}.
    \end{align*}   
    To show the non-expansiveness property with respect to $\mysp(\cdot)$, let $\bar{c}=\argmin_{c\in\mathbb{R}}\|Q_1-Q_2-ce\|_\infty$. Then, we have
    \begin{align*}
        \mysp(\mathcal{H}(Q_1)-\mathcal{H}(Q_2))=\,&\min_{c\in\mathbb{R}}\|\mathcal{H}(Q_1)-\mathcal{H}(Q_2)-ce\|_\infty\\
        \leq \,&\|\mathcal{H}(Q_1)-\mathcal{H}(Q_2)-\bar{c}e\|_\infty\\
        \leq\,& \|Q_1-Q_2-\bar{c}e\|_\infty\\
        =\,&\mysp(Q_1-Q_2).
    \end{align*}
    \item Using Part (1) of this lemma, we have
    \begin{align*}
        \mysp(\mathcal{H}(Q))\leq \mysp(\mathcal{H}(Q)-\mathcal{H}(0))+\mysp(\mathcal{H}(0))\leq \mysp(Q)+\mysp(\mathcal{H}(0)).
    \end{align*}
    Since 
    \begin{align*}
        \mysp(\mathcal{H}(0))\leq \|\mathcal{H}(0)\|_\infty=\max_{s,a}|\mathcal{R}(s,a)|\leq 1,
    \end{align*}
    we have
    \begin{align*}
        \mysp(\mathcal{H}(Q))\leq \mysp(Q)+1.
    \end{align*}
\end{enumerate}

\subsection{Proof of Lemma \ref{le:T_1}}\label{pf:le:T_1}
Since $\mathcal{H}(Q^*)-Q^*=r^*e\in\text{ker}(\mysp)$, we have by Lemma \ref{le:Moreau} (4) that
\begin{align}
    \langle\nabla M(Q_k-Q^*),\mathcal{H}(Q_k)-Q_k\rangle=\,&\langle\nabla M(Q_k-Q^*),\mathcal{H}(Q_k)-\mathcal{H}(Q^*)+Q^*-Q_k\rangle\nonumber\\
    =\,&\underbrace{\langle\nabla M(Q_k-Q^*),\mathcal{H}(Q_k)-\mathcal{H}(Q^*)\rangle}_{T_{1,1}}\nonumber\\
    &+\underbrace{\langle\nabla M(Q_k-Q^*),Q^*-Q_k\rangle}_{T_{1,2}}.\label{eq:pf:le:T1}
\end{align}
Next, we bound the terms $T_{1,1}$ and $T_{1,2}$ on the right-hand side of the previous inequality.

Since $M(Q)=p_m(Q)^2/2$ differentiable, we have
\begin{align*}
    \nabla M(Q)=p_m(Q)\nabla p_m(Q),\quad \forall\,Q\in\mathbb{R}^{|\mathcal{S}||\mathcal{A}|}.
\end{align*}
It follows that for any $c\in\mathbb{R}$, we have
\begin{align*}
    T_{1,1}
    =\,&\langle\nabla M(Q_k-Q^*),\mathcal{H}(Q_k)-\mathcal{H}(Q^*)+ce\rangle\tag{Lemma \ref{le:Moreau} (4)}\\
    =\,&p_m(Q_k-Q^*)\langle\nabla p_m(Q_k-Q^*),\mathcal{H}(Q_k)-\mathcal{H}(Q^*)-ce\rangle\\
    \leq \,&p_m(Q_k-Q^*)\|\nabla p_m(Q_k-Q^*)\|_{m,*}\|\mathcal{H}(Q_k)-\mathcal{H}(Q^*)-ce\|_m,
\end{align*}
where $\|\cdot\|_{m,*}$ is the dual norm of $\|\cdot\|_m$. Since
\begin{align*}
    |p_m(Q_1)-p_m(Q_2)|\leq p_m(Q_1-Q_2)=\min_{c'\in\mathbb{R}}\|Q_1-Q_2-c'e\|_m\leq \|Q_1-Q_2\|_m,
\end{align*}
the function $p_m(\cdot)$ is $1$-Lipschitz continous with respect to $\|\cdot\|_m$. It then follows from \cite[Lemma 2.6]{shalev2012online} that $\|\nabla p_m(Q_k-Q^*)\|_{m,*}\leq 1$.
Therefore, we have
\begin{align}
    T_{1,1}\leq \,&p_m(Q_k-Q^*)\|\nabla p_m(Q_k-Q^*)\|_{m,*}\|\mathcal{H}(Q_k)-\mathcal{H}(Q^*)-ce\|_m\nonumber\\
    \leq \,&p_m(Q_k-Q^*)\|\mathcal{H}(Q_k)-\mathcal{H}(Q^*)-ce\|_m\nonumber\\
    = \,&p_m(Q_k-Q^*)p_m(\mathcal{H}(Q_k)-\mathcal{H}(Q^*))\tag{Choosing $c=\argmin_{c'\in\mathbb{R}}\|\mathcal{H}(Q_k)-\mathcal{H}(Q^*)-c'e\|_m$}\nonumber\\
    \leq  \,&\frac{1}{\ell_m}p_m(Q_k-Q^*)\mysp(\mathcal{H}(Q_k)-\mathcal{H}(Q^*))\tag{Lemma \ref{le:Moreau} (3)}\nonumber\\
    \leq \,&\frac{\beta}{\ell_m}p_m(Q_k-Q^*)\mysp(Q_k-Q^*)\tag{Assumption \ref{as:seminorm-contraction}}\nonumber\\
    \leq \,&\frac{\beta u_m}{\ell_m}p_m(Q_k-Q^*)^2\tag{Lemma \ref{le:Moreau} (3)}\nonumber\\
    =\,&\frac{2\beta u_m}{\ell_m}M(Q_k-Q^*),\label{eq:pf:le:T11}
\end{align}
where the last inequality follows from Lemma \ref{le:Moreau} (2).

Next, we consider the term $T_{1,2}$ on the right-hand side of Equation (\ref{eq:pf:le:T1}). Since $p_m(\cdot)$ is a convex function, which follows from 
\begin{align*}
    p_m(\alpha Q_1+(1-\alpha)Q_2)\leq p_m(\alpha Q_1)+p_m((1-\alpha)Q_2)=\alpha p_m( Q_1)+(1-\alpha) p_m(Q_2)
\end{align*}
for any $\alpha\in [0,1]$ and $Q_1,Q_2\in\mathbb{R}^{|\mathcal{S}||\mathcal{A}|}$, we have
\begin{align*}
    p_m(0)-p_m(Q_k-Q^*)\geq \langle \nabla p_m(Q_k-Q^*),Q^*-Q_k)\rangle.
\end{align*}
It follows that
\begin{align}
    T_{1,2}=\,&\langle\nabla M(Q_k-Q^*),Q^*-Q_k\rangle\nonumber\\
    =\,&p_m(Q_k-Q^*)\langle\nabla p_m(Q_k-Q^*),Q^*-Q_k\rangle\nonumber\\
    \leq \,&-p_m(Q_k-Q^*)^2\nonumber\\
    =\,&-2M(Q_k-Q^*).\label{eq:pf:le:T12}
\end{align}
Finally, using Equations (\ref{eq:pf:le:T11}) and (\ref{eq:pf:le:T12}) together in Equation (\ref{eq:pf:le:T1}), we have 
\begin{align*}
    \langle\nabla M(Q_k-Q^*),\mathcal{H}(Q_k)-Q_k\rangle\leq\,& T_{1,1}+T_{1,2}\\
    \leq\,&  -2\left(1-\frac{\beta u_m}{\ell_m}\right)M(Q_k-Q^*)\\
    =\,&  -2\phi_1 M(Q_k-Q^*).
\end{align*}
Taking expectations on both sides of the previous inequality finishes the proof.

\subsection{Proof of Proposition \ref{prop:log-growth}}\label{pf:prop:log-growth}
The following lemma is needed for the proof.
\begin{lemma}\label{le:time-varying-bound}
The following inequality holds for all $k \geq 1$:
\begin{align*}
    \mysp(Q_{k+1}) \leq \mysp(Q_k) + \alpha_k(S_k,A_k).
\end{align*}
\end{lemma}
The proof of Lemma \ref{le:time-varying-bound} is presented in Appendix \ref{pf:le:time-varying-bound}.

Recursively applying Lemma \ref{le:time-varying-bound} and then using the definition of $\alpha_k(S_k,A_k)$, we obtain
\begin{align*}
    \mysp(Q_k) \leq \mysp(Q_1) + \sum_{i=1}^{k-1} \frac{\alpha}{N_i(S_i,A_i)+h}, \quad \forall\, k \geq 1.
\end{align*}
It remains to bound the quantity $\sum_{i=1}^{k-1}\alpha/(N_i(S_i,A_i)+h)$.
For any $k \geq 1$, define the set $\mathcal{M}_{k-1}$ as
\begin{align*}
    \mathcal{M}_{k-1} = (\mathcal{S} \times \mathcal{A})^{k-1} 
    = \{(s_1,a_1,\dots,s_{k-1},a_{k-1}) | s_i \in \mathcal{S},\, a_i \in \mathcal{A},\, \forall\, i = 1,2,\dots,k-1\}.
\end{align*}
Then, the following inequality holds with probability one:
\begin{align*}
    \sum_{i=1}^{k-1} \frac{\alpha}{N_i(S_i,A_i)+h} 
    \leq \max_{(s_1,a_1,\dots,s_{k-1},a_{k-1}) \in \mathcal{M}_{k-1}} \sum_{i=1}^{k-1} \frac{\alpha}{N_i(s_i,a_i)+h}.
\end{align*}

Next, we identify a scheduling $(s_1,a_1,\dots,s_{k-1},a_{k-1})$ that achieves the maximum on the right-hand side of the previous inequality. Let
\begin{align*}
    N_{\max} &= \max_{(s,a)} \sum_{i=1}^{k-1} \mathbf{1}_{\{(s_i,a_i) = (s,a)\}}, & 
    (\bar{s},\bar{a}) &\in \argmax_{(s,a)} \sum_{i=1}^{k-1} \mathbf{1}_{\{(s_i,a_i) = (s,a)\}}, \\[6pt]
    N_{\min} &= \min_{(s,a)} \sum_{i=1}^{k-1} \mathbf{1}_{\{(s_i,a_i) = (s,a)\}}, & 
    (\underline{s},\underline{a}) &\in \argmin_{(s,a)} \sum_{i=1}^{k-1} \mathbf{1}_{\{(s_i,a_i) = (s,a)\}}.
\end{align*}
We now show that if a scheduling is optimal, it must hold that $N_{\max} - N_{\min} \leq 1$. Suppose, for contradiction, that $N_{\max} - N_{\min} \geq 2$. Consider a modified scheduling identical to the original one except that, instead of choosing $(\bar{s},\bar{a})$ at its last occurrence, we select $(\underline{s},\underline{a})$ instead. Under this modification, the objective increases by exactly 
\begin{align*}
    \frac{\alpha}{N_{\min}+1} - \frac{\alpha}{N_{\max}},
\end{align*}
which is strictly positive since $N_{\max} - N_{\min} \geq 2$, contradicting the optimality of the original scheduling. Therefore, an optimal scheduling must satisfy $N_{\max} - N_{\min} \leq 1$. 

Next, we show that all schedulings satisfying $N_{\max} - N_{\min} \leq 1$ must yield the same value. Let $k-1 = m|\mathcal{S}||\mathcal{A}| + n$, where $m$ is the quotient and $n < |\mathcal{S}||\mathcal{A}|$ is the remainder. The only way to satisfy $N_{\max} - N_{\min} \leq 1$ is for exactly $n$ state-action pairs to be visited $m+1$ times and the remaining $|\mathcal{S}||\mathcal{A}| - n$ pairs to be visited $m$ times. For any such scheduling $(s_1,a_1,\cdots,s_{k-1},a_{k-1})$, the corresponding value is
\begin{align*}
    \sum_{i=1}^{k-1} \frac{\alpha}{N_i(s_i,a_i)+h} 
    =  \left(|\mathcal{S}||\mathcal{A}|\sum_{i=1}^m \frac{\alpha}{i+h} + \frac{n\alpha}{m+1+h}\right).
\end{align*}
Therefore, we obtain
\begin{align*}
    \sum_{i=1}^{k-1} \frac{\alpha}{N_i(S_i,A_i)+h} 
    \leq\, & \max_{(s_1,a_1,\dots,s_{k-1},a_{k-1}) \in \mathcal{M}_k} \sum_{i=1}^{k-1} \frac{\alpha}{N_i(s_i,a_i)+h} \\
    =\, & \alpha\left(|\mathcal{S}||\mathcal{A}| \sum_{i=1}^m \frac{1}{i+h} + \frac{n}{m+1+h}\right) \\
    \leq\, & \alpha|\mathcal{S}||\mathcal{A}| \sum_{i=1}^{m+1} \frac{1}{i+h}\\
    =\, & \alpha|\mathcal{S}||\mathcal{A}| \sum_{i=1}^{t} \frac{1}{i+h},
\end{align*}
where $t=\lceil(k-1)/(|\mathcal{S}||\mathcal{A}|)\rceil$.
Since
\begin{align*}
    \sum_{i=1}^t \frac{1}{i+h} \leq \int_0^t \frac{1}{x+h}\, dx = \log\left(\frac{t+h}{h}\right),
\end{align*}
we conclude that
\begin{align*}
    \mysp(Q_k)\leq \mysp(Q_1)+\alpha|\mathcal{S}||\mathcal{A}|\log \left(\frac{\lceil (k-1) / (|\mathcal{S}||\mathcal{A}|) \rceil+h}{h}\right)=b_k.
\end{align*}

\subsection{Proof of Lemma \ref{le:MC_concentration}}\label{pf:le:MC_concentration}
Fixing $x\in \mathcal{X}$, let $W_x:\mathcal{X}^k\to \mathbb{R}$ be a function defined as
\begin{align*}    W_x(x_1,x_2,\cdots,x_k)=\frac{\sum_{i=1}^k\mathbf{1}_{\{x_i=x\}}}{k},\quad \forall\,x_{1,k}\in\mathcal{X}^k,
\end{align*}
where we use $x_{1,k}$ to denote $(x_1,x_2,\cdots,x_k)$ for simplicity of notation.

It is clear that $W(\cdot)$ satisfies
\begin{align*}
    W_x(x_1,x_2,\cdots,x_k)-W_x(y_1,y_2,\cdots,y_k)\leq \sum_{i=1}^k\frac{1}{k}\mathbf{1}_{\{x_i\neq y_i\}},\quad \forall\,x_{1,k},y_{1,k}\in\mathcal{X}^k.
\end{align*}
With the above notation, our goal is to bound $\mathbb{P}\left(|W_x(X_{1,k})-\nu(x)|\geq \epsilon\right)$. Observe that
\begin{align*}
    |W_x(X_{1,k})-\nu(x)|\leq |W_x(X_{1,k})-\mathbb{E}[W_x(X_{1,k})]|+|\mathbb{E}[W_x(X_{1,k})]-\nu(x)|.
\end{align*}
Since 
\begin{align*}
    |\mathbb{E}[W_x(X_{1,k})]-\nu(x)|=\,&\frac{1}{k}\left|\sum_{i=1}^k\left[p(X_i=x\mid X_1)-\mu(x)\right]\right|\\
    \leq\,& \frac{1}{k}\sum_{i=1}^k\left|p(X_i=x\mid X_1)-\mu(x)\right|\\
    \leq \,&\frac{1}{k}\sum_{i=1}^k2\tilde{C}\tilde{\rho}^{i-1}\\
    \leq \,&\frac{2\tilde{C}}{(1-\tilde{\rho})k},
\end{align*}
we have
\begin{align*}
    |W_x(X_{1,k})-\nu(x)|\leq |W_x(X_{1,k})-\mathbb{E}[W_x(X_{1,k})]|+\frac{2\tilde{C}}{(1-\tilde{\rho})k}.
\end{align*}
Therefore, for any $\epsilon\geq \frac{4\tilde{C}}{(1-\tilde{\rho})k}$, applying Corollary 2.11 from \cite{paulin2015concentration}, we have
\begin{align*}
    \mathbb{P}\left(|W_x(X_{1,k})-\nu(x)|\geq \epsilon\right)\leq \,&\mathbb{P}\left(|W_x(X_{1,k})-\mathbb{E}[W_x(X_{1,k})]|\geq \epsilon-\frac{2\tilde{C}}{(1-\tilde{\rho})k}\right)\\
    \leq \,&2\exp\left(-c'k\left(\epsilon-\frac{2\tilde{C}}{(1-\tilde{\rho})k}\right)^2\right)\\
    \leq \,&2\exp\left(-\frac{c'k\epsilon^2}{4}\right),
\end{align*}
where $c'$ is a constant depending on the mixing time of the Markov chain $\{X_k\}$. The result follows by redefining $c=c'/4$.

\subsection{Proof of Proposition \ref{prop:T2}}\label{pf:prop:T2}
We begin with the following decomposition:
\begin{align}
    T_2
    =\,&\mathbb{E}[\langle\nabla M(Q_k-Q^*),F(Q_k,D,Y_k)-\mathcal{H}(Q_k)\rangle]\nonumber\\
    =\,&\underbrace{\mathbb{E}[\langle\nabla M(Q_{k-\tau_k}-Q^*),F(Q_{k-\tau_k}, D,Y_k)-\mathcal{H}(Q_{k-\tau_k})\rangle]}_{:=T_{2,1}}\nonumber\\
	&+\underbrace{\mathbb{E}[\langle\nabla M(Q_{k-\tau_k}-Q^*),F(Q_k,D,Y_k)-F(Q_{k-\tau_k},D,Y_k)-\mathcal{H}(Q_k)+\mathcal{H}(Q_{k-\tau_k})\rangle]}_{:=T_{2,2}}\nonumber\\
	&+\underbrace{\mathbb{E}[\langle\nabla M(Q_k-Q^*)-\nabla M(Q_{k-\tau_k}-Q^*),F(Q_k,D,Y_k)-\mathcal{H}(Q_k)\rangle]}_{:=T_{2,3}},\label{eq:T2_decomposition}
\end{align}
where we recall that $\tau_k=\min\{t:C\rho^{t-1}\leq \alpha/(k+h)\}$.

\subsubsection{Bounding the Term $T_{2,1}$}\label{ap:T_21}
We begin with some notation. For any $k\geq 1$, let $\mathcal{F}_k$ be the $\sigma$-algebra generated by the set of random variables $\{Q_1,S_1,A_1,S_2,A_2,\cdots,S_{k-1},A_{k-1},S_k\}$. Note that $Q_k$ is measurable with respect to $\mathcal{F}_k$. By the tower property of conditional expectations, we have for any $k\geq \tau_k+1$ that
\begin{align*}
    T_{2,1}=\,&\mathbb{E}[\langle\nabla M(Q_{k-\tau_k}-Q^*),F(Q_{k-\tau_k},D,Y_k)-\mathcal{H}(Q_{k-\tau_k})\rangle]\\
    =\,&\mathbb{E}[\langle\nabla M(Q_{k-\tau_k}-Q^*),\mathbb{E}[F(Q_{k-\tau_k},D,Y_k)| \mathcal{F}_{k-\tau_k}]-\mathcal{H}(Q_{k-\tau_k})\rangle]\\
    \leq \,&L\mathbb{E}[\mysp(Q_{k-\tau_k}-Q^*)\mysp(\mathbb{E}[F(Q_{k-\tau_k},D,Y_k)| \mathcal{F}_{k-\tau_k}]-\mathcal{H}(Q_{k-\tau_k}))],
\end{align*}
where the last inequality follows from Lemma \ref{le:Moreau} (5).

According to Proposition \ref{prop:log-growth}, we have
\begin{align*}
    \mysp(Q_{k-\tau_k}-Q^*)\leq \mysp(Q_{k-\tau_k})+\mysp(Q^*)\leq b_{k-\tau_k}+\mysp(Q^*)\leq b_k+\mysp(Q^*).
\end{align*}
which further implies
\begin{align*}
    T_{2,1}
    \leq \,&L(b_k+\mysp(Q^*))\mathbb{E}[\mysp(\mathbb{E}[F(Q_{k-\tau_k},D,Y_k)| \mathcal{F}_{k-\tau_k}]-\mathcal{H}(Q_{k-\tau_k}))]\\
    \leq \,&L(b_k+\mysp(Q^*))\mathbb{E}\left[\sum_{y\in\mathcal{Y}}|p(Y_k=y| Y_{k-\tau_k-1})-\nu(y)|\mysp(F(Q_{k-\tau_k},D,y))\right],
\end{align*}
where the last line follows from Lemma \ref{le:F_properties} (3) and the triangle inequality.

On the one hand, we have by Lemma \ref{le:F_properties} (2) and Proposition \ref{prop:log-growth} that
\begin{align*}
    \mysp(F(Q_{k-\tau_k},D,y))\leq \frac{2(\mysp(Q_{k-\tau_k})+1)}{D_{\min}}\leq \frac{2(b_{k-\tau_k}+1)}{D_{\min}}
    \leq \frac{2(b_k+1)}{D_{\min}}.
\end{align*}
On the other hand, we have by Assumption \ref{as:MC} that
\begin{align*}
    \sum_{y\in\mathcal{Y}}|p(Y_k=y| Y_{k-\tau_k-1})-\nu(y)|=\,&\sum_{s_0,a_0,s_1}|p(Y_k=(s_0,a_0,s_1)| Y_{k-\tau_k-1})-\nu(s_0,a_0,s_1)|\\
    =\,&\sum_{s_0,a_0,s_1}|p_\pi(S_k=s_0| S_{k-\tau_k})-\mu(s_0)|\pi(a_0|s_0)p(s_1|s_0,a_0)\\
    =\,&\sum_{s_0}|p_\pi(S_k=s_0| S_{k-\tau_k})-\mu(s_0)|\\
    =\,&2\|p_\pi(S_k=\cdot| S_{k-\tau_k})-\mu(\cdot)\|_{\text{TV}}\\
    \leq \,&2 C\rho^{\tau_k}.
\end{align*}
Together, they imply
\begin{align}
    T_{2,1}\leq \,&L(b_k+\mysp(Q^*))\mathbb{E}\left[\sum_{y\in\mathcal{Y}}|p(Y_k=y| Y_{k-\tau_k-1})-\nu(y)|\mysp(F(Q_{k-\tau_k},D,y))\right]\nonumber\\
    \leq \,&\frac{4L(b_k+\mysp(Q^*))(b_k+1)}{D_{\min}}C\rho^{\tau_k}\nonumber\\
    \leq\,& \frac{4L(b_k+\mysp(Q^*))(b_k+1)}{D_{\min}}\alpha_k\tag{This follows from the definition of $\tau_k$.}\\
    \leq\,& \frac{4L(b_k+\mysp(Q^*)+1)^2}{D_{\min}}\alpha_k.\label{eq:bound:T_21}
\end{align}
for all $k\geq \tau_k+1$.

\subsubsection{Bounding the Term $T_{2,2}$}\label{ap:T_22}
The following lemma is needed to bound the terms $T_{2,2}$ and $T_{2,3}$ on the right-hand side of Equation (\ref{eq:T2_decomposition}). See Appendix \ref{pf:le:difference} for its proof.
\begin{lemma}\label{le:difference}
	Let $k_1$ be a positive integer. Then, we have for all $k\geq k_1$ that
	\begin{align*}
		\mysp(Q_k-Q_{k_1})\leq  f\left(\frac{\alpha(k-k_1)}{N_{k_1-1,\min}+1+h}\right)(\mysp(Q_{k_1})+1),
	\end{align*}
    where $f(x):=xe^x$ for all $x>0$ and $N_{k,\min}:=\min_{s,a}N_k(s,a)$ for any $k\geq 1$.
\end{lemma}

Next, we proceed to bound the term $T_{2,2}$ on the right-hand side of Equation (\ref{eq:T2_decomposition}).

Using Lemma \ref{le:Moreau} (5), we have
\begin{align*}
    T_{2,2}=\,&\mathbb{E}[\langle\nabla M(Q_{k-\tau_k}-Q^*),F(Q_k,D,Y_k)-F(Q_{k-\tau_k},D,Y_k)\rangle]\\
    &+\mathbb{E}[\langle\nabla M(Q_{k-\tau_k}-Q^*),\mathcal{H}(Q_{k-\tau_k})-\mathcal{H}(Q_k)\rangle]\\
    \leq \,&L\mathbb{E}[\mysp(Q_{k-\tau_k}-Q^*)\mysp(F(Q_k,D,Y_k)-F(Q_{k-\tau_k},D,Y_k))]\\
    &+L\mathbb{E}[\mysp(Q_{k-\tau_k}-Q^*)\mysp(\mathcal{H}(Q_k)-\mathcal{H}(Q_{k-\tau_k}))]\\
    \leq \,&\left(\frac{2L}{D_{\min}}+L\right)\mathbb{E}[\mysp(Q_{k-\tau_k}-Q^*)\mysp(Q_k-Q_{k-\tau_k})]\tag{Lemma \ref{le:F_properties} (1) and Lemma \ref{le:H_properties} (1)}\\
    \leq \,&\frac{3L}{D_{\min}}\mathbb{E}[\mysp(Q_{k-\tau_k}-Q^*)\mysp(Q_k-Q_{k-\tau_k})],
\end{align*}
where the last inequality follows from $D_{\min}\in (0,1)$.

On the one hand, we have by Lemma \ref{le:difference} and Proposition \ref{prop:log-growth} that
\begin{align*}
    \mysp(Q_k-Q_{k-\tau_k})\leq\,& f\left(\frac{\alpha\tau_k}{N_{k-\tau_k-1,\min}+1+h}\right)(\mysp(Q_{k-\tau_k})+1)\\
    \leq\,& f\left(\frac{\alpha\tau_k}{N_{k-\tau_k-1,\min}+1+h}\right)(b_{k-\tau_k}+1)\\
    \leq\,& f\left(\frac{\alpha\tau_k}{N_{k-\tau_k-1,\min}+1+h}\right)(b_k+1).
\end{align*}
On the other hand, we have
\begin{align*}
    \mysp(Q_{k-\tau_k}-Q^*)\leq \mysp(Q_{k-\tau_k})+\mysp(Q^*)\leq b_{k-\tau_k}+\mysp(Q^*)\leq b_k+\mysp(Q^*).
\end{align*}
Together, they imply 
\begin{align}\label{eq1:pf:T22}
    T_{2,2}\leq \,&\frac{3L}{D_{\min}}\mathbb{E}[\mysp(Q_{k-\tau_k}-Q^*)\mysp(Q_k-Q_{k-\tau_k})]\nonumber\\
    \leq \,&\frac{3L(b_k+1)(b_k+\mysp(Q^*))}{D_{\min}}\mathbb{E}\left[f\left(\frac{\alpha\tau_k}{N_{k-\tau_k-1,\min}+1+h}\right)\right]\nonumber\\
    \leq \,&\frac{3L(b_k+\mysp(Q^*)+1)^2}{D_{\min}}\mathbb{E}\left[f\left(\frac{\alpha\tau_k}{N_{k-\tau_k-1,\min}+1+h}\right)\right].
\end{align}
It remains to bound the term $\mathbb{E}[f(\alpha\tau_k/(N_{k-\tau_k-1,\min}+1+h)]$. 

Using the definition of $f(\cdot)$ (cf. Lemma \ref{le:difference}), we have
\begin{align*}
    &\mathbb{E}\left[f\left(\frac{\alpha\tau_k}{N_{k-\tau_k-1,\min}+1+h}\right)\right]\\
    =\,&\mathbb{E}\left[\frac{\alpha\tau_k }{N_{k-\tau_k-1,\min}+1+h}\exp\left(\frac{\alpha\tau_k}{N_{k-\tau_k-1,\min}+1+h}\right)\right]\\
    =\,&\mathbb{E}\left[\max_{s,a}\left\{\frac{\alpha\tau_k }{N_{k-\tau_k-1}(s,a)+1+h}\exp\left(\frac{\alpha\tau_k}{N_{k-\tau_k-1}(s,a)+1+h}\right)\right\}\right]\tag{$f(x)=xe^x$ is an increasing function of $x$ on $[0,\infty)$.}\\
    \leq \,&\sum_{s,a}\mathbb{E}\left[\frac{\alpha\tau_k }{N_{k-\tau_k-1}(s,a)+1+h}\exp\left(\frac{\alpha\tau_k}{N_{k-\tau_k-1}(s,a)+1+h}\right)\right]\\
    =\,&\sum_{s,a}\mathbb{E}\left[\frac{\alpha\tau_k}{(k-\tau_k-1)\bar{D}_{k-\tau_k-1}(s,a)+1+h}\exp\left(\frac{\alpha\tau_k}{(k-\tau_k-1)\bar{D}_{k-\tau_k-1}(s,a)+1+h}\right)\right],
\end{align*}
where the last line follows from our notation $\bar{D}_k(s,a)=N_k(s,a)/k$. 

To proceed, for any $\delta>0$ and $(s,a)$, let 
\begin{align*}
    E_\delta(s,a)=\{|\bar{D}_{k-\tau_k-1}(s,a)-D(s,a)|\leq \delta D(s,a)\}
\end{align*}
and let $E_\delta^c(s,a)$ be the complement of the event $E_\delta(s,a)$. Note that on the event $E_\delta(s,a)$, we have
\begin{align*}
    \bar{D}_{k-\tau_k-1}(s,a)\geq (1-\delta)D(s,a),
\end{align*}
while on the event $E_\delta^c(s,a)$, we have $\bar{D}_{k-\tau_k-1}(s,a)\geq 0$. Therefore, we obtain
\begin{align*}
    &\mathbb{E}\left[f\left(\frac{\alpha\tau_k}{N_{k-\tau_k-1,\min}+1+h}\right)\right]\\
    \leq \,&\sum_{s,a}\mathbb{E}\left[\frac{\alpha\tau_k}{(k-\tau_k-1)\bar{D}_{k-\tau_k-1}(s,a)+1+h}\exp\left(\frac{\alpha\tau_k}{(k-\tau_k-1)\bar{D}_{k-\tau_k-1}(s,a)+1+h}\right)\right]\\
    =\,&\sum_{s,a}\mathbb{E}\left[\frac{\alpha\tau_k (\mathbf{1}_{\{E_\delta(s,a)\}}+\mathbf{1}_{\{E_\delta^c(s,a)\}})}{(k-\tau_k-1)\bar{D}_{k-\tau_k-1}(s,a)+1+h}\exp\left(\frac{\alpha\tau_k}{(k-\tau_k-1)\bar{D}_{k-\tau_k-1}(s,a)+1+h}\right)\right]\\
    \leq \,&\sum_{s,a}\frac{\alpha\tau_k }{(k-\tau_k-1)(1-\delta)D(s,a)+1+h}\exp\left(\frac{\alpha\tau_k}{(k-\tau_k-1) (1-\delta)D(s,a)+1+h)}\right)\\
    &+\frac{\alpha\tau_k}{h+1}\exp\left(\frac{\alpha\tau_k}{h+1}\right)\sum_{s,a}\mathbb{P}(E_\delta^c(s,a)).
\end{align*}
To proceed, let $K'$ be such that $k\geq 2\tau_k$ for all $k\geq K'$, which is always possible since $\tau_k=\min\{t:C\rho^{t-1}\leq \alpha_k\}=\mathcal{O}(\log(k))$. Then, we have 
\begin{align*}
    \frac{\alpha\tau_k }{(k-\tau_k-1)(1-\delta)D(s,a)+1+h}
    \leq \,&\frac{\alpha\tau_k }{(k-\tau_k-1)(1-\delta)D(s,a)+(1+h)(1-\delta)D(s,a)}\\
    =\,&\frac{\alpha}{(1-\delta)D(s,a)}\frac{\tau_k }{k+h-\tau_k}\\
    \leq \,&\frac{\alpha}{(1-\delta)D(s,a)}\frac{2\tau_k }{k+h}\\
    \leq \,&\frac{\alpha}{(1-\delta)D_{\min}}\frac{2\tau_k }{k+h},
\end{align*}
which further implies
\begin{align*}
    \mathbb{E}\left[f\left(\frac{\alpha\tau_k}{N_{k-\tau_k-1,\min}+1+h}\right)\right]\leq  \,&\frac{2\alpha\tau_k |\mathcal{S}||\mathcal{A}| }{(k+h)(1-\delta)D_{\min}}\exp\left(\frac{2\alpha\tau_k}{(k+h) (1-\delta)D_{\min}}\right)\\
    &+\frac{\alpha\tau_k}{h+1}\exp\left(\frac{\alpha\tau_k}{h+1}\right)\sum_{s,a}\mathbb{P}(E_\delta^c(s,a))
\end{align*}
because 
$f(x)=xe^x$ is an increasing function of $x$ on $[0,\infty)$.

To bound $\mathbb{P}(E_\delta^c(s,a))$, we use the Markov chain concentration inequality stated in Lemma \ref{le:MC_concentration}. As long as 
\begin{align}\label{eq:delta}
    \delta\geq \frac{4C}{D_{\min}(1-\rho)(k-\tau_k-1)},
\end{align}
we have
\begin{align*}
    \mathbb{P}(E_\delta^c(s,a))= \,&\mathbb{P}\left(|\bar{D}_{k-\tau_k-1}(s,a)-D(s,a)|> \delta D(s,a)\right)\\
    \leq \,&2\exp(-c_{mc} (k-\tau_k-1)\delta^2D(s,a)^2),
\end{align*}
where $c_{mc}$ is a constant depending on the mixing time of the Markov chain $\{S_k\}$ induced by $\pi$. It follows that 
\begin{align*}
    &\mathbb{E}\left[f\left(\frac{\alpha\tau_k}{N_{k-\tau_k-1,\min}+1+h}\right)\right]\\
    \leq \,&\frac{2\alpha\tau_k |\mathcal{S}||\mathcal{A}| }{(k+h)(1-\delta)D_{\min}}\exp\left(\frac{2\alpha\tau_k}{(k+h) (1-\delta)D_{\min}}\right)\\
    &+\frac{2\alpha\tau_k}{h+1}\exp\left(\frac{\alpha\tau_k}{h+1}\right)\sum_{s,a}\exp(-c_{mc} (k-\tau_k-1)\delta^2D(s,a)^2)\\
    \leq \,&\frac{2\alpha\tau_k |\mathcal{S}||\mathcal{A}| }{(k+h)(1-\delta)D_{\min}}\exp\left(\frac{2\alpha\tau_k}{(k+h) (1-\delta)D_{\min}}\right)\\
    &+\frac{2\alpha\tau_k |\mathcal{S}||\mathcal{A}|}{h+1}\exp\left(\frac{\alpha\tau_k}{h+1}\right)\exp(-c_{mc} (k-\tau_k-1)\delta^2D_{\min}^2).
\end{align*}
The next step is to choose $\delta$ appropriately. Specifically, we want to choose $\delta$ such that
\begin{align*}
    \frac{2\alpha\tau_k |\mathcal{S}||\mathcal{A}|}{h+1}\exp\left(\frac{\alpha\tau_k}{h+1}\right)\exp(-c_{mc} (k-\tau_k-1)\delta^2D_{\min}^2)=\frac{\alpha \tau_k |\mathcal{S}||\mathcal{A}|}{(k+h)D_{\min}},
\end{align*}
which is equivalent to
\begin{align}\label{eq:choose-delta}
    \delta 
    =
    \sqrt{
        \frac{1}{c_{mc}(k-\tau_k-1)\,D_{\min}^2}
        \left[\log\left(\frac{2(k+h)D_{\min}}{h+1}\right)
        +
        \frac{\alpha\tau_k}{h+1}\right]
    }.
\end{align}
In view of Equation (\ref{eq:delta}) on the lower bound of $\delta$, let $K''$ be such that the following inequality holds for all $k\geq K''$:
\begin{align*}
    \sqrt{
        \frac{1}{c_{mc}(k-\tau_k-1)\,D_{\min}^2}
        \left[\log\left(\frac{2(k+h)D_{\min}}{h+1}\right)
        +
        \frac{\alpha\tau_k}{h+1}\right]
    }\geq \frac{4C}{D_{\min}(1-\rho)(k-\tau_k-1)},
\end{align*}
Then, for all $k\geq K''$, choosing $\delta$ according to Equation (\ref{eq:choose-delta}),
we have
\begin{align*}
    \mathbb{E}\left[f\left(\frac{\alpha\tau_k}{N_{k-\tau_k-1,\min}+1+h}\right)\right]
    \leq \,&\frac{2\alpha\tau_k |\mathcal{S}||\mathcal{A}| }{(k+h)(1-\delta)D_{\min}}\exp\left(\frac{2\alpha\tau_k}{(k+h) (1-\delta)D_{\min}}\right)\\
    &+\frac{\alpha \tau_k |\mathcal{S}||\mathcal{A}|}{(k+h)D_{\min}}.
\end{align*}
Since $\tau_k=\min\{t:C\rho^{t-1}\leq \alpha_k\}=\mathcal{O}(\log(k))$, there exists $K'''>0$ such that the following inequality holds for all $k\geq K'''$:
\begin{align*}
    \frac{2\alpha\tau_k |\mathcal{S}||\mathcal{A}| }{(k+h)(1-\delta)D_{\min}}\exp\left(\frac{2\alpha\tau_k}{(k+h) (1-\delta)D_{\min}}\right)\leq \frac{3\alpha\tau_k |\mathcal{S}||\mathcal{A}| }{(k+h)D_{\min}}.
\end{align*}
Therefore, when $k\geq K'''$, we have 
\begin{align}\label{eq:Bound_on_f}
    \mathbb{E}\left[f\left(\frac{\alpha\tau_k}{N_{k-\tau_k-1,\min}+1+h}\right)\right]\leq \frac{4\alpha\tau_k |\mathcal{S}||\mathcal{A}|}{(k+h)D_{\min}}=\frac{4\tau_k |\mathcal{S}||\mathcal{A}|}{D_{\min}}\alpha_k.
\end{align}
Finally, using the previous inequality in Equation (\ref{eq1:pf:T22}), we have for any $k\geq K_2:=\max(K',K'',K''')$ that
\begin{align}
    T_{2,2}\leq\,& \frac{3L(b_k+\mysp(Q^*)+1)^2}{D_{\min}}\mathbb{E}\left[f\left(\frac{\alpha\tau_k}{N_{k-\tau_k-1,\min}+1+h}\right)\right]\nonumber\\
    \leq \,&\frac{3L(b_k+\mysp(Q^*)+1)^2}{D_{\min}}\frac{4\tau_k |\mathcal{S}||\mathcal{A}|}{D_{\min}}\alpha_k\nonumber\\
    =\,&\frac{12\tau_k L|\mathcal{S}||\mathcal{A}| (b_k+\mysp(Q^*)+1)^2}{ D_{\min}^2}\alpha_k.\label{eq:bound:T_22}
\end{align}

\subsubsection{Bounding the Term $T_{2,3}$}\label{ap:T_23}
Using Lemma \ref{le:Moreau} (5), we have
\begin{align*}
    T_{2,3}=\,&\mathbb{E}[\langle\nabla M(Q_k-Q^*)-\nabla M(Q_{k-\tau_k}-Q^*),F(Q_k,D,Y_k)-\mathcal{H}(Q_k)\rangle]\\
    \leq \,&L\mathbb{E}[\mysp(Q_k-Q_{k-\tau_k})\mysp(F(Q_k,D,Y_k)-\mathcal{H}(Q_k))].
\end{align*}
On the one hand, we have by Lemma \ref{le:difference} and Proposition \ref{prop:log-growth} that
\begin{align*}
    \mysp(Q_k-Q_{k-\tau_k})\leq\,& f\left(\frac{\alpha\tau_k}{N_{k-\tau_k-1,\min}+1+h}\right)(\mysp(Q_{k-\tau_k})+1)\\
    \leq\,& f\left(\frac{\alpha\tau_k}{N_{k-\tau_k-1,\min}+1+h}\right)(b_{k-\tau_k}+1)\\
    \leq\,& f\left(\frac{\alpha\tau_k}{N_{k-\tau_k-1,\min}+1+h}\right)(b_k+1).
\end{align*}
On the other hand, we have by Lemma \ref{le:F_properties} (2), Lemma \ref{le:H_properties} (2), and Proposition \ref{prop:log-growth} that
\begin{align*}
    \mysp(F(Q_k,D,Y_k)-\mathcal{H}(Q_k))\leq \,&\mysp(F(Q_k,D,Y_k))+\mysp(\mathcal{H}(Q_k))\\
    \leq \,&\frac{2(\mysp(Q_k)+1)}{D_{\min}}+\mysp(Q_k)+1\\
    \leq \,&\frac{3(b_k+1)}{D_{\min}}.
\end{align*}
Together, they imply
\begin{align*}
    T_{2,3}
    \leq \,&L\mathbb{E}[\mysp(Q_k-Q_{k-\tau_k})\mysp(F(Q_k,D,Y_k)-\mathcal{H}(Q_k))]\\
    \leq \,&\frac{3L(b_k+1)^2}{ D_{\min}}\mathbb{E}\left[f\left(\frac{\alpha\tau_k}{N_{k-\tau_k-1,\min}+1+h}\right)\right]\\
    \leq \,&\frac{3L(b_k+\mysp(Q^*)+1)^2}{ D_{\min}}\mathbb{E}\left[f\left(\frac{\alpha\tau_k}{N_{k-\tau_k-1,\min}+1+h}\right)\right].
\end{align*}
Recall that we have already shown in the previous section (cf. Equation (\ref{eq:Bound_on_f})) that 
\begin{align*}
    \mathbb{E}\left[f\left(\frac{\alpha\tau_k}{N_{k-\tau_k-1,\min}+1+h}\right)\right]\leq \frac{4\tau_k |\mathcal{S}||\mathcal{A}|}{D_{\min}}\alpha_k,\quad \forall\,k\geq K_2.
\end{align*}
Therefore, we have for any $k\geq K_2$ that
\begin{align}\label{eq:bound:T_23}
    T_{2,3}\leq\frac{12\tau_k L|\mathcal{S}||\mathcal{A}|(b_k+\mysp(Q^*)+1)^2}{D_{\min}^2}\alpha_k.
\end{align}

\subsubsection{Combining Everything Together}
Finally, using the bounds we obtained for the terms $T_{2,1}$ (cf. Equation (\ref{eq:bound:T_21}) from Appendix \ref{ap:T_21}), $T_{2,2}$ (cf. Equation (\ref{eq:bound:T_22}) from Appendix \ref{ap:T_22}), and $T_{2,3}$ (cf. Equation (\ref{eq:bound:T_23}) from Appendix \ref{ap:T_23}) altogether in Equation (\ref{eq:T2_decomposition}), we have
for all $k\geq K_2$ that
\begin{align*}
    T_2\leq\,&\frac{4L(b_k+\mysp(Q^*)+1)^2}{D_{\min}}\alpha_k+\frac{12\tau_k L|\mathcal{S}||\mathcal{A}| (b_k+\mysp(Q^*)+1)^2}{ D_{\min}^2}\alpha_k\\
    &+\frac{12\tau_k L|\mathcal{S}||\mathcal{A}|(b_k+\mysp(Q^*)+1)^2}{D_{\min}^2}\alpha_k\\
    \leq \,&\frac{28\tau_k L|\mathcal{S}||\mathcal{A}|(b_k+\mysp(Q^*)+1)^2}{D_{\min}^2}\alpha_k.
\end{align*}
The proof of Proposition \ref{prop:T2} is complete.

\subsection{Proof of Proposition \ref{prop:T3}}\label{pf:prop:T3}
Using Lemma \ref{le:Moreau} (3) and (4), we have for any $c\in\mathbb{R}$ that
\begin{align*}
    &\langle\nabla M(Q_k-Q^*),F(Q_k,D_k,Y_k)-F(Q_k,D,Y_k)\rangle\\
    =\,&\langle\nabla M(Q_k-Q^*),F(Q_k,D_k,Y_k)-F(Q_k,D,Y_k)-ce\rangle\\
    = \,&p_m(Q_k-Q^*)\langle\nabla p_m(Q_k-Q^*),F(Q_k,D_k,Y_k)-F(Q_k,D,Y_k)-ce\rangle\\
    \leq  \,&p_m(Q_k-Q^*)\|\nabla p_m(Q_k-Q^*)\|_{m,*}\|F(Q_k,D_k,Y_k)-F(Q_k,D,Y_k)-ce\|_m.
\end{align*}
By choosing
\begin{align*}
    c=\argmin_{c'\in\mathbb{R}}\|F(Q_k,D_k,Y_k)-F(Q_k,D,Y_k)-c'e\|_m,
\end{align*}
we have by Lemma \ref{le:Moreau} (2) that
\begin{align*}
    &\langle\nabla M(Q_k-Q^*),F(Q_k,D_k,Y_k)-F(Q_k,D,Y_k)\rangle\\
    \leq  \,&p_m(Q_k-Q^*)\|\nabla p_m(Q_k-Q^*)\|_{m,*}p_m(F(Q_k,D_k,Y_k)-F(Q_k,D,Y_k)).
\end{align*}
Recall that we have shown in the proof of Lemma \ref{le:T_1} that $\|\nabla p_m(Q_k-Q^*)\|_{m,*}\leq 1$. Therefore, we have
\begin{align*}
    &\langle\nabla M(Q_k-Q^*),F(Q_k,D_k,Y_k)-F(Q_k,D,Y_k)\rangle\\
    \leq  \,&p_m(Q_k-Q^*)p_m(F(Q_k,D_k,Y_k)-F(Q_k,D,Y_k))\\
    \leq  \,&\frac{1}{\ell_m}p_m(Q_k-Q^*)\mysp(F(Q_k,D_k,Y_k)-F(Q_k,D,Y_k))\tag{Lemma \ref{le:Moreau} (3)}\\
    \leq \,&\frac{1}{2c'\ell_m}p_m^2(Q_k-Q^*)+\frac{c'}{2\ell_m}\mysp(F(Q_k,D_k,Y_k)-F(Q_k,D,Y_k))^2\tag{This follows from Cauchy - Schwarz inequality, where $c'>0$ can be arbitrary.}\\
    = \,&\frac{1}{c'\ell_m}M(Q_k-Q^*)+\frac{c'}{2\ell_m}\mysp(F(Q_k,D_k,Y_k)-F(Q_k,D,Y_k))^2,
\end{align*}
where the last line follows from $M(Q)=p_m(Q)^2/2$ (cf. Lemma \ref{le:Moreau} (2)). By choosing 
\begin{align*}
   c'= \frac{1}{\ell_m(1 - \beta u_{m}/\ell_m)},
\end{align*}
we have from  the previous inequality that
\begin{align*}
    &\langle\nabla M(Q_k-Q^*),F(Q_k,D_k,Y_k)-F(Q_k,D,Y_k)\rangle\\
    \leq \,&\frac{1}{c'\ell_m}M(Q_k-Q^*)+\frac{c'}{2\ell_m}\mysp(F(Q_k,D_k,Y_k)-F(Q_k,D,Y_k))^2\\
    =\,&\left(1-\frac{\beta u_m}{\ell_m}\right)M(Q_k-Q^*)+\frac{\mysp(F(Q_k,D_k,Y_k)-F(Q_k,D,Y_k))^2}{2\ell_m^2(1-\beta u_m/\ell_m)}.
\end{align*}
To proceed, observe that the vector $F(Q_k,D_k,Y_k)-F(Q_k,D,Y_k)$ has only one non-zero entry, i.e., the $(S_k,A_k)$-th one. Therefore, we have by the definition of $\mysp(\cdot)$ that
\begin{align*}
    &\mysp(F(Q_k,D_k,Y_k)-F(Q_k,D,Y_k))^2\\
    =\,&\frac{1}{4}\left(\frac{1}{D_k(S_k,A_k)}-\frac{1}{D(S_k,A_k)}\right)^2\left(\mathcal{R}(S_k,A_k)+\max_{a'\in\mathcal{A}}Q_k(S_{k+1},a')-Q_k(S_k,A_k)\right)^2\\
    \leq \,&\frac{1}{4}\left(\frac{1}{D_k(S_k,A_k)}-\frac{1}{D(S_k,A_k)}\right)^2\left(1+2\mysp(Q_k)\right)^2\tag{Definition of $\mysp(\cdot)$ and $\max_{s,a}|\mathcal{R}(s,a)|\leq 1$}\\
    \leq \,&(1+b_k)^2\left(\frac{1}{D_k(S_k,A_k)}-\frac{1}{D(S_k,A_k)}\right)^2,
\end{align*}
where the last line follows from Proposition \ref{prop:log-growth}. 

Combining the previous two inequalities together, we have
\begin{align}\label{eq2:pf:T3}
    T_3=\,&\mathbb{E}\left[\langle\nabla M(Q_k-Q^*),F(Q_k,D_k,Y_k)-F(Q_k,D,Y_k)\rangle\right]\nonumber\\
    \leq\,& \left(1-\frac{\beta u_m}{\ell_m}\right)\mathbb{E}\left[M(Q_k-Q^*)\right]+\frac{\mathbb{E}\left[\mysp(F(Q_k,D_k,Y_k)-F(Q_k,D,Y_k))^2\right]}{2\ell_m^2(1-\beta u_m/\ell_m)}\nonumber\\
    \leq\,& \left(1-\frac{\beta u_m}{\ell_m}\right)\mathbb{E}\left[M(Q_k-Q^*)\right]+\frac{(1+b_k)^2\mathbb{E}\left[\left(\frac{1}{D_k(S_k,A_k)}-\frac{1}{D(S_k,A_k)}\right)^2\right]}{2\ell_m^2(1-\beta u_m/\ell_m)}.
\end{align}

It remains to bound $\mathbb{E}[(1/D_k(S_k,A_k) - 1/D(S_k,A_k))^2]$, which is highly nontrivial due to the following reasons: (1) for any fixed $(s,a)$, $D_k(s,a)$ depends on the entire history of the Markov chain induced by the behavior policy; (2) $D_k(s,a)$ is correlated with $(S_k,A_k)$; and (3) the appearance of random variables in the denominators of fractions destroys linearity.

For any $\tilde{\tau}\leq k-1$, we have
\begin{align*}
    &\mathbb{E}\left[\left(\frac{1}{D_k(S_k,A_k)}-\frac{1}{D(S_k,A_k)}\right)^2\right]\\
    =\,&\mathbb{E}\left[\frac{(D_k(S_k,A_k)-D(S_k,A_k))^2}{D_k(S_k,A_k)^2D(S_k,A_k)^2}\right]\\
    =\,&\mathbb{E}\left[\frac{(N_k(S_k,A_k)+h-(k+h)D(S_k,A_k))^2}{(N_k(S_k,A_k)+h)^2D(S_k,A_k)^2}\right]\tag{$D_k(S_k,A_k)=(N_k(S_k,A_k)+h)/(k+h)$}\\
    =\,&\mathbb{E}\left[\frac{(N_{k-1}(S_k,A_k)+1+h-(k+h)D(S_k,A_k))^2}{(N_{k-1}(S_k,A_k)+1+h)^2D(S_k,A_k)^2}\right]\tag{$(S_k,A_k)$ is visited at time step $k$}\\
    =\,&\sum_{s,a}\mathbb{E}\left[\mathbf{1}_{\{(S_k,A_k)=(s,a)\}}\frac{(N_{k-1}(s,a)+1+h-(k+h)D(s,a))^2}{(N_{k-1}(s,a)+1+h)^2D(s,a)^2}\right]\\
    \leq  \,&\sum_{s,a}\mathbb{E}\left[\mathbf{1}_{\{(S_k,A_k)=(s,a)\}}\frac{(N_{k-1}(s,a)+1+h-(k+h)D(s,a))^2}{(N_{k-\tilde{\tau}-1}(s,a)+1+h)^2D(s,a)^2}\right]\tag{$N_{k_1}(s,a)\leq N_{k_2}(s,a)$ for any $k_1\leq k_2$.}\\
    =\,&\sum_{s,a}\mathbb{E}\left[\mathbf{1}_{\{(S_k,A_k)=(s,a)\}}\frac{(A_{k,\tilde{\tau}}+B_{k,\tilde{\tau}}+C_{k,\tilde{\tau}})^2}{(N_{k-\tilde{\tau}-1}(s,a)+1+h)^2D(s,a)^2}\right],
\end{align*}
where
\begin{align*}
    A_{k,\tilde{\tau}}:=\,&N_{k-\tilde{\tau}-1}(s,a)-(k-\tilde{\tau}-1)D(s,a)=(k-\tilde{\tau}-1)(\bar{D}_{k-\tilde{\tau}-1}(s,a)-D(s,a)),\\
    B_{k,\tilde{\tau}}:=\,&N_{k-1}(s,a)-N_{k-\tilde{\tau}-1}(s,a)-\tilde{\tau} D(s,a),\text{ which satisfies }|B_{k,\tilde{\tau}}|\leq \tilde{\tau},\\
    C_{k,\tilde{\tau}}:=\,&(1+h)(1-D(s,a)), \text{ which satisfies }|C_{k,\tilde{\tau}}|\leq (1+h).
\end{align*}
It follows that
\begin{align*}
    &\mathbb{E}\left[\left(\frac{1}{D_k(S_k,A_k)}-\frac{1}{D(S_k,A_k)}\right)^2\right]\\
    \leq \,&\sum_{s,a}\mathbb{E}\left[\mathbf{1}_{\{(S_k,A_k)=(s,a)\}}\frac{(A_{k,\tilde{\tau}}+B_{k,\tilde{\tau}}+C_{k,\tilde{\tau}})^2}{(N_{k-\tilde{\tau}-1}(s,a)+1+h)^2D(s,a)^2}\right]\\
    \leq \,&\sum_{s,a}\mathbb{E}\left[\mathbf{1}_{\{(S_k,A_k)=(s,a)\}}\frac{3(A_{k,\tilde{\tau}}^2+\tilde{\tau}^2+(h+1)^2)}{(N_{k-\tilde{\tau}-1}(s,a)+1+h)^2D(s,a)^2}\right]\tag{$(a+b+c)^2\leq 3(a^2+b^2+c^2)$ for any $a,b,c\in\mathbb{R}$.}\\
    =\,&\sum_{s,a}\mathbb{E}\left[\mathbb{P}(S_k=s,A_k=a| S_{k-\tilde{\tau}-1},A_{k-\tilde{\tau}-1})\frac{3(A_{k,\tilde{\tau}}^2+\tilde{\tau}^2+(h+1)^2)}{(N_{k-\tilde{\tau}-1}(s,a)+1+h)^2D(s,a)^2}\right],
\end{align*}
where the last line follows from the tower property of conditional expectations and the Markov property. 
	
Observe that
\begin{align*}
    &\mathbb{P}(S_k=s,A_k=a| S_{k-\tilde{\tau}-1},A_{k-\tilde{\tau}-1})\\
    \leq \,&|\mathbb{P}(S_k=s,A_k=a| S_{k-\tilde{\tau}-1},A_{k-\tilde{\tau}-1})-D(s,a)|+D(s,a)\\
    =\,&\sum_{s'}|\mathbb{P}(S_{k-\tilde{\tau}}=s'| S_{k-\tilde{\tau}-1},A_{k-\tilde{\tau}-1})p_\pi(S_k=s| S_{k-\tilde{\tau}}=s')\pi(a|s)-\mu(s)\pi(a|s)|+D(s,a)\\
    \leq \,&\max_{s'}|p_\pi(S_k=s| S_{k-\tilde{\tau}}=s')-\mu(s)|+D(s,a)\\
    \leq \,&\max_{s'}\sum_{s}|p_\pi(S_k=s| S_{k-\tilde{\tau}}=s')-\mu(s)|+D(s,a)\\
    \leq \,&2C\rho^{\tilde{\tau}}+D(s,a),
\end{align*}
where the last inequality follows from Assumption \ref{as:MC}. By choosing
\begin{align*}
    \tilde{\tau}=\min\{t:2C\rho^t\leq D_{\min}\},
\end{align*}
we have
\begin{align}
    \mathbb{P}(S_k=s,A_k=a| S_{k-\tilde{\tau}-1},A_{k-\tilde{\tau}-1})\leq 2D(s,a).\label{eq:indicator_probability}
\end{align}
It follows that
\begin{align*}
    &\mathbb{E}\left[\left(\frac{1}{D_k(S_k,A_k)}-\frac{1}{D(S_k,A_k)}\right)^2\right]\\
    \leq \,&2\sum_{s,a}D(s,a)\mathbb{E}\left[\frac{3(A_{k,\tilde{\tau}}^2+\tilde{\tau}^2+(h+1)^2)}{(N_{k-\tilde{\tau}-1}(s,a)+1+h)^2D(s,a)^2}\right]\\
    =\,&2\sum_{s,a}D(s,a)\mathbb{E}\left[\frac{3(k-\tilde{\tau}-1)^2(\bar{D}_{k-\tilde{\tau}-1}(s,a)-D(s,a))^2}{((k-\tilde{\tau}-1)\bar{D}_{k-\tilde{\tau}-1}(s,a)+1+h)^2D(s,a)^2}\right]\\
    &+2\sum_{s,a}D(s,a)\mathbb{E}\left[\frac{3(\tilde{\tau}^2+(h+1)^2)}{((k-\tilde{\tau}-1)\bar{D}_{k-\tilde{\tau}-1}(s,a)+1+h)^2D(s,a)^2}\right].
\end{align*}
To proceed, for any $\delta\in (0,1)$ and $(s,a)$, let $E_\delta(s,a)=\{|\bar{D}_{k-\tilde{\tau}-1}(s,a)-D(s,a)|\leq \delta D(s,a)\}$ and let $E_\delta^c(s,a)$ be the complement of event $E_\delta(s,a)$.  Note that on the event $E_\delta(s,a)$, we have
\begin{align*}
    (1-\delta)D(s,a)\leq \bar{D}_{k-\tilde{\tau}-1}(s,a)\leq (1+\delta)D(s,a),
\end{align*}
while on the event $E_\delta^c(s,a)$, we trivially have $\bar{D}_{k-1}(s,a)\geq 0$. It follows that
\begin{align*}
    &\mathbb{E}\left[\left(\frac{1}{D_k(S_k,A_k)}-\frac{1}{D(S_k,A_k)}\right)^2\right]\\
    \leq \,&2\sum_{s,a}D(s,a)\mathbb{E}\left[\frac{3(k-\tilde{\tau}-1)^2(\bar{D}_{k-\tilde{\tau}-1}(s,a)-D(s,a))^2(\mathbf{1}_{E_\delta(s,a)}+\mathbf{1}_{E_\delta(s,a)^c})}{((k-\tilde{\tau}-1)\bar{D}_{k-\tilde{\tau}-1}(s,a)+1+h)^2D(s,a)^2}\right]\\
    &+2\sum_{s,a}D(s,a)\mathbb{E}\left[\frac{3(\tilde{\tau}^2+(h+1)^2)(\mathbf{1}_{E_\delta(s,a)}+\mathbf{1}_{E_\delta(s,a)^c})}{((k-\tilde{\tau}-1)\bar{D}_{k-\tilde{\tau}-1}(s,a)+1+h)^2D(s,a)^2}\right]\\
    \leq \,&2\sum_{s,a}\frac{3(k-\tilde{\tau}-1)^2}{((k-\tilde{\tau}-1)(1-\delta)D(s,a)+1+h)^2D(s,a)}\mathbb{E}\left[(\bar{D}_{k-\tilde{\tau}-1}(s,a)-D(s,a))^2\right]\\
    &+2\sum_{s,a}\frac{3(k-\tilde{\tau}-1)^2}{(1+h)^2D(s,a)}\mathbb{P}(E_\delta^c(s,a))\\
    &+2\sum_{s,a}\frac{3(\tilde{\tau}^2+(h+1)^2)}{((k-\tilde{\tau}-1)(1-\delta)D(s,a)+1+h)^2D(s,a)}\\
    &+2\sum_{s,a}\frac{3(\tilde{\tau}^2+(h+1)^2)}{(1+h)^2D(s,a)}\mathbb{P}(E_\delta^c(s,a))\\
    = \,&2\sum_{s,a}\frac{3(k-\tilde{\tau}-1)^2}{((k-\tilde{\tau}-1)(1-\delta)D(s,a)+1+h)^2D(s,a)}\mathbb{E}\left[(\bar{D}_{k-\tilde{\tau}-1}(s,a)-D(s,a))^2\right]\\
    &+2\sum_{s,a}\frac{3(\tilde{\tau}^2+(h+1)^2)}{((k-\tilde{\tau}-1)(1-\delta)D(s,a)+1+h)^2D(s,a)}\\
    &+2\sum_{s,a}\frac{3((k-\tilde{\tau}-1)^2+\tilde{\tau}^2+(1+h)^2)}{(1+h)^2D(s,a)}\mathbb{P}(E_\delta^c(s,a)).
\end{align*}
It remains to bound the terms $\mathbb{E}[(\bar{D}_{k-\tilde{\tau}-1}(s,a) - D(s,a))^2]$ and $\mathbb{P}(E_\delta^c(s,a))$. To bound $\mathbb{E}[(\bar{D}_{k-\tilde{\tau}-1}(s,a) - D(s,a))^2]$, we use the following lemma, which is a mean-square bound for functions of finite, irreducible, and aperiodic Markov chains. The proof of Lemma \ref{le:MC-MSE} is provided in Appendix \ref{pf:le:MC-MSE}.

\begin{lemma}\label{le:MC-MSE}
    Under Assumption \ref{as:MC}, the following inequality holds for all $k\geq 14C/[(1-\rho)D_{\min}]$:
    \begin{align*}
        \mathbb{E}\left[(\bar{D}_{k}(s,a)-D(s,a))^2\right]\leq \frac{10CD(s,a)}{(1-\rho)k}.
    \end{align*}
\end{lemma}
For simplicity of notation, denote $C_{mc}=C/(1-\rho)$.
Apply Lemma \ref{le:MC-MSE}, we have when $k\geq 14C_{mc}/D_{\min}$ that
\begin{align*}
    \mathbb{E}\left[\left(\frac{1}{D_k(S_k,A_k)}-\frac{1}{D(S_k,A_k)}\right)^2\right]
    \leq \,&\sum_{s,a}\frac{60 C_{mc}(k-\tilde{\tau}-1)}{((k-\tilde{\tau}-1)(1-\delta)D(s,a)+1+h)^2}\\
    &+\sum_{s,a}\frac{6(\tilde{\tau}^2+(h+1)^2)}{((k-\tilde{\tau}-1)(1-\delta)D(s,a)+1+h)^2D(s,a)}\\
    &+\sum_{s,a}\frac{6((k-\tilde{\tau}-1)^2+\tilde{\tau}^2+(1+h)^2)}{(1+h)^2D(s,a)}\mathbb{P}(E_\delta^c(s,a)).
    \end{align*}
To proceed, observe that
\begin{align*}
    \frac{60 C_{mc}(k-\tilde{\tau}-1)}{((k-\tilde{\tau}-1)(1-\delta)D(s,a)+1+h)^2}\leq \,&\frac{60 C_{mc}(k-\tilde{\tau}-1)}{((k-\tilde{\tau}-1)(1-\delta)D(s,a)+(1+h)(1-\delta)D(s,a))^2}\\
    \leq \,&\frac{60 C_{mc}(k-\tilde{\tau}-1)}{(k+h-\tilde{\tau})^2(1-\delta)^2D(s,a)^2}
\end{align*}
and
\begin{align*}
    &\frac{6(\tilde{\tau}^2+(h+1)^2)}{((k-\tilde{\tau}-1)(1-\delta)D(s,a)+1+h)^2D(s,a)}\\
    \leq \,&\frac{6(\tilde{\tau}^2+(h+1)^2)}{((k-\tilde{\tau}-1)(1-\delta)D(s,a)+(1+h)(1-\delta)D(s,a))^2D(s,a)}\\
    =\,&\frac{6(\tilde{\tau}^2+(h+1)^2)}{(k+h-\tilde{\tau})^2(1-\delta)^2D(s,a)^3}.
\end{align*}
Therefore, we have
\begin{align*}
    \mathbb{E}\left[\left(\frac{1}{D_k(S_k,A_k)}-\frac{1}{D(S_k,A_k)}\right)^2\right]\leq \,&\frac{60C_{mc}|\mathcal{S}||\mathcal{A}|(k-\tilde{\tau}-1)}{(k+h-\tilde{\tau})^2(1-\delta)^2D_{\min}^2}+\frac{6|\mathcal{S}||\mathcal{A}|(\tilde{\tau}^2+(h+1)^2)}{(k+h-\tilde{\tau})^2(1-\delta)^2D_{\min}^3}\\
		&+\frac{6|\mathcal{S}||\mathcal{A}|((k-\tilde{\tau}-1)^2+\tilde{\tau}^2+(1+h)^2)}{(1+h)^2D_{\min}}\mathbb{P}(E_\delta^c(s,a)).
\end{align*}
To bound $\mathbb{P}(E_\delta^c(s,a))$, we use the Markov chain concentration inequality stated in Lemma \ref{le:MC_concentration}, which implies
\begin{align*}
    \mathbb{P}(E_\delta^c(s,a))=\,&\mathbb{P}\left(|\bar{D}_{k-\tilde{\tau}-1}(s,a)-D(s,a)|> \delta D(s,a)\right)\\
    \leq \,&2\exp(-c_{mc} (k-\tilde{\tau}-1)\delta^2D(s,a)^2)
\end{align*}
for all $\delta\geq 4C/[D_{\min}(1-\rho)(k-\tilde{\tau}-1)]$.
It follows that
\begin{align*}
    &\mathbb{E}\left[\left(\frac{1}{D_k(S_k,A_k)}-\frac{1}{D(S_k,A_k)}\right)^2\right]\\
    \leq \,&\frac{60C_{mc}|\mathcal{S}||\mathcal{A}|(k-\tilde{\tau}-1)}{(k+h-\tilde{\tau})^2(1-\delta)^2D_{\min}^2}+\frac{6|\mathcal{S}||\mathcal{A}|(\tilde{\tau}^2+(h+1)^2)}{(k+h-\tilde{\tau})^2(1-\delta)^2D_{\min}^3}\\
    &+\frac{12|\mathcal{S}||\mathcal{A}|((k-\tilde{\tau}-1)^2+\tilde{\tau}^2+(1+h)^2)}{(1+h)^2D_{\min}}\exp(-c_{mc} (k-\tilde{\tau}-1)\delta^2D_{\min}^2).
\end{align*}
Let $\tilde{K}'$ be such that the following inequality holds for all $k\geq \tilde{K}'$:
\begin{align*}
    \frac{4C}{D_{\min}(1-\rho)(k-\tilde{\tau}-1)}\leq \sqrt{\frac{\log\left(\frac{12(k+h)((k-\tilde{\tau}-1)^2+\tilde{\tau}^2+(1+h)^2)}{(1+h)^2}\right)}{c_{mc} (k-\tilde{\tau}-1)D_{\min}^2}}.
\end{align*}
Then, for any $k\geq \tilde{K}'$, choose
\begin{align*}
    \delta=\sqrt{\frac{1}{c_{mc} (k-\tilde{\tau}-1)D_{\min}^2}\log\left(\frac{12(k+h)((k-\tilde{\tau}-1)^2+\tilde{\tau}^2+(1+h)^2)}{(1+h)^2}\right)},
\end{align*}
we have
\begin{align*}
    \frac{12|\mathcal{S}||\mathcal{A}|((k-\tilde{\tau}-1)^2+\tilde{\tau}^2+(1+h)^2)}{(1+h)^2D_{\min}}\exp(-c_{mc} (k-\tilde{\tau}-1)\delta^2D_{\min}^2)=\frac{|\mathcal{S}||\mathcal{A}|}{(k+h)D_{\min}}.
\end{align*}
It follows that
\begin{align*}
    \mathbb{E}\left[\left(\frac{1}{D_k(S_k,A_k)}-\frac{1}{D(S_k,A_k)}\right)^2\right]
    \leq \,&\frac{60C_{mc}|\mathcal{S}||\mathcal{A}|(k-\tilde{\tau}-1)}{(k+h-\tilde{\tau})^2(1-\delta)^2D_{\min}^2}+\frac{6|\mathcal{S}||\mathcal{A}|(\tilde{\tau}^2+(h+1)^2)}{(k+h-\tilde{\tau})^2(1-\delta)^2D_{\min}^3}\\
    &+\frac{|\mathcal{S}||\mathcal{A}|}{(k+h)D_{\min}}.
\end{align*}
Let $\tilde{K}''$ be such that the following inequalities hold for all $k\geq \tilde{K}''$:
\begin{align*}
    \frac{(k-\tilde{\tau}-1)}{(k+h-\tilde{\tau})^2(1-\delta)^2}\leq\frac{61}{60(k+h)},\quad 
    \frac{(\tilde{\tau}^2+(h+1)^2)}{(k+h-\tilde{\tau})^2(1-\delta)^2D_{\min}}\leq \frac{1}{6(k+h)}.
\end{align*}
Then, we have for all $k\geq \tilde{K}''$ that
\begin{align*}
    \mathbb{E}\left[\left(\frac{1}{D_k(S_k,A_k)}-\frac{1}{D(S_k,A_k)}\right)^2\right]
    \leq \,&\frac{61C_{mc}|\mathcal{S}||\mathcal{A}|}{(k+h)D_{\min}^2}+\frac{|\mathcal{S}||\mathcal{A}|}{(k+h)D_{\min}^2}+\frac{|\mathcal{S}||\mathcal{A}|}{(k+h)D_{\min}}\\
    \leq \,&\frac{63C_{mc}|\mathcal{S}||\mathcal{A}|}{(k+h)D_{\min}^2},
\end{align*}
where the last line follows from $C_{mc}=C/(1-\rho)\geq 1$. Finally, combining the previous inequality with Equation (\ref{eq2:pf:T3}), we have for all $k\geq K_3:=\max(14C_{mc}/D_{\min},\tilde{K}',\tilde{K}'')$ that
\begin{align}
    T_3\leq\,& \left(1-\frac{\beta u_m}{\ell_m}\right)\mathbb{E}\left[M(Q_k-Q^*)\right]+\frac{(1+b_k)^2\mathbb{E}\left[\left(\frac{1}{D_k(S_k,A_k)}-\frac{1}{D(S_k,A_k)}\right)^2\right]}{2\ell_m^2(1-\beta u_m/\ell_m)}\nonumber\\
    \leq \,&\left(1-\frac{\beta u_m}{\ell_m}\right)\mathbb{E}\left[M(Q_k-Q^*)\right]+\frac{32C_{mc}|\mathcal{S}||\mathcal{A}|(1+b_k)^2}{\ell_m^2(1-\beta u_m/\ell_m)\alpha D_{\min}^2}\alpha_k\nonumber\\
    \leq\,& \left(1-\frac{\beta u_m}{\ell_m}\right)\mathbb{E}\left[M(Q_k-Q^*)\right]+\frac{32C_{mc}|\mathcal{S}||\mathcal{A}|(b_k+\mysp(Q^*)+1)^2}{\ell_m^2(1-\beta u_m/\ell_m)\alpha D_{\min}^2}\alpha_k\nonumber\\
    =\,& \phi_1\mathbb{E}\left[M(Q_k-Q^*)\right]+\frac{32C|\mathcal{S}||\mathcal{A}|(b_k+\mysp(Q^*)+1)^2}{\ell_m^2\phi_1(1-\rho)\alpha D_{\min}^2}\alpha_k,\label{eq:bound:T_24}
\end{align}
where the last line follows from our definition $\phi_1=1-\beta u_m/\ell_m$.

\subsection{Proof of Lemma \ref{le:T4}}\label{pf:le:T4}
Since the vector $F(Q_k,D_k,Y_k)-Q_k$ has only one non-zero entry, i.e., the $(S_k,A_k)$-th one, we have by the definition of $\mysp(\cdot)$ that
\begin{align*}
    &\mysp(F(Q_k,D_k,Y_k)-Q_k)^2\\
    =\,&\frac{1}{4}\left(\frac{1}{D_k(S_k,A_k)}\right)^2\left(\mathcal{R}(S_k,A_k)+\max_{a'\in\mathcal{A}}Q_k(S_{k+1},a')-Q_k(S_k,A_k)\right)^2\\
    \leq \,&\frac{1}{4}\left(\frac{1}{D_k(S_k,A_k)}\right)^2\left(1+2\mysp(Q_k)\right)^2\\
    \leq \,&\frac{1}{4}\left(\frac{1}{D_k(S_k,A_k)}\right)^2\left(1+2b_k\right)^2\tag{Proposition \ref{prop:log-growth}}\\
    \leq \,&\left(1+b_k\right)^2\left(\frac{1}{D_k(S_k,A_k)}\right)^2.
\end{align*}
Taking expectations on both sides, we obtain
\begin{align}\label{eq3:pf:T4}
    T_4\leq \left(1+b_k\right)^2\mathbb{E}\left[\frac{1}{D_k(S_k,A_k)^2}\right].
\end{align}
It remains to bound $\mathbb{E}[1/D_k(S_k,A_k)^2]$. Observe that for any $\tilde{\tau}\leq k-1$, we have
\begin{align*}
    \mathbb{E}\left[\frac{1}{D_k(S_k,A_k)^2}\right]=\,&\mathbb{E}\left[\frac{(k+h)^2}{(N_k(S_k,A_k)+h)^2}\right]\tag{$D_k(s,a)=(N_k(s,a)+h)/(k+h)$}\\
    =\,&\mathbb{E}\left[\frac{(k+h)^2}{(N_{k-1}(S_k,A_k)+1+h)^2}\right]\tag{The pair $(S_k,A_k)$ is visited at time step $k$.}\\
    =\,&\sum_{s,a}\mathbb{E}\left[\frac{\mathbf{1}_{\{(s,a)=(S_k,A_k)\}}(k+h)^2}{(N_{k-1}(s,a)+1+h)^2}\right]\\
    \leq \,&\sum_{s,a}\mathbb{E}\left[\frac{\mathbf{1}_{\{(s,a)=(S_k,A_k)\}}(k+h)^2}{(N_{k-\tilde{\tau}-1}(s,a)+1+h)^2}\right]\tag{$N_{k_1}(s,a)\leq N_{k_2}(s,a)$ for any $k_1\leq k_2$}\\
    =\,&\sum_{s,a}\mathbb{E}\left[\mathbb{P}(S_k=s,A_k=a| S_{k-\tilde{\tau}-1},A_{k-\tilde{\tau}-1})\frac{(k+h)^2}{(N_{k-\tilde{\tau}-1}(s,a)+1+h)^2}\right],
\end{align*}
where the last line follows from the tower property of conditional expectations and the Markov property. 

Recall that we have shown in Equation (\ref{eq:indicator_probability}) that, when choosing $\tilde{\tau}=\min\{t:C\rho^t\leq D_{\min}\}$,  we have
\begin{align*}
    \mathbb{P}\left(S_k=s,A_k=a| S_{k-\tilde{\tau}-1},A_{k-\tilde{\tau}-1}\right)
    \leq  2D(s,a).
\end{align*}
It follows that
\begin{align}\label{eq2:pf:T4}
    \mathbb{E}\left[\frac{1}{D_k(S_k,A_k)^2}\right]\leq \sum_{s,a}2D(s,a)\mathbb{E}\left[\frac{(k+h)^2}{(N_{k-\tilde{\tau}-1}(s,a)+1+h)^2}\right].
\end{align}
To proceed and bound $\mathbb{E}\left[(k+h)^2/(N_{k-\tilde{\tau}-1}(s,a)+1+h)^2\right]$, given $\delta\in (0,1)$, for any $(s,a)$, let 
\begin{align*}
    E_\delta(s,a)=\{|\bar{D}_{k-\tilde{\tau}-1}(s,a)-D(s,a)|\leq \delta D(s,a)\},
\end{align*}
and let $E_\delta^c(s,a)$ be the complement of event $E_\delta(s,a)$. Note that on the event $E_\delta(s,a)$, we have $|\bar{D}_{k-\tilde{\tau}-1}(s,a)-D(s,a)|\leq \delta D(s,a)$, which implies
\begin{align*}
    \bar{D}_{k-\tilde{\tau}-1}(s,a)\geq (1-\delta)D(s,a),
\end{align*}
while on the event $E_\delta^c(s,a)$, we have the trivial bound $\bar{D}_{k-\tilde{\tau}-1}(s,a)\geq 0$. Therefore, we obtain
\begin{align*}
    \mathbb{E}\left[\frac{(k+h)^2}{(N_{k-\tilde{\tau}-1}(s,a)+1+h)^2}\right]=\,&\frac{(k+h)^2}{(k-\tilde{\tau}-1)^2}\mathbb{E}\left[\frac{(k-\tilde{\tau}-1)^2}{(N_{k-\tilde{\tau}-1}(s,a)+1+h)^2}\right]\\
    =\,&\frac{(k+h)^2}{(k-\tilde{\tau}-1)^2}\mathbb{E}\left[\frac{1}{(\bar{D}_{k-\tilde{\tau}-1}(s,a)+(1+h)/(k-\tilde{\tau}-1))^2}\right]\\
    =\,&\frac{(k+h)^2}{(k-\tilde{\tau}-1)^2}\mathbb{E}\left[\frac{\mathbf{1}_{\{E_\delta(s,a)\}}+\mathbf{1}_{\{E_\delta^c(s,a)\}}
    }{(\bar{D}_{k-\tilde{\tau}-1}(s,a)+(1+h)/(k-\tilde{\tau}-1))^2}\right]\\
    \leq \,&\frac{(k+h)^2}{(k-\tilde{\tau}-1)^2}\frac{1}{((1-\delta)D(s,a)+(1+h)/(k-\tilde{\tau}-1))^2}\\
    &+\frac{(k+h)^2}{(h+1)^2}\mathbb{P}(E_\delta^c(s,a))\\
    \leq \,&\frac{(k+h)^2}{(k-\tilde{\tau}-1)^2}\frac{1}{(1-\delta)^2D(s,a)^2}+\frac{(k+h)^2}{(h+1)^2}\mathbb{P}(E_\delta^c(s,a)).
\end{align*}
To bound $\mathbb{P}(E_\delta^c(s,a))$, we use the Markov chain concentration inequality stated in Lemma \ref{le:MC_concentration}, which implies
\begin{align*}
    \mathbb{P}(E_\delta^c(s,a))=\,&\mathbb{P}\left(|\bar{D}_{k-\tilde{\tau}-1}(s,a)-D(s,a)|> \delta D(s,a)\right)\\
    \leq \,&2\exp(-c_{mc} (k-\tilde{\tau}-1)\delta^2D(s,a)^2)
\end{align*}
for any $\delta\geq 4C/[D_{\min}(1-\rho)(k-\tilde{\tau}-1)]$.
Combining the previous two inequalities, we have
\begin{align*}
    \mathbb{E}\left[\frac{(k+h)^2}{(N_{k-\tilde{\tau}-1}(s,a)+1+h)^2}\right]
    \leq \,&\frac{(k+h)^2}{(k-\tilde{\tau}-1)^2}\frac{1}{(1-\delta)^2D(s,a)^2}\\
    &+\frac{2(k+h)^2}{(h+1)^2}\exp(-c_{mc} (k-\tilde{\tau}-1)\delta^2D(s,a)^2).
    \end{align*}
    Let $\bar{K}'$ be such that the following inequality holds for all $k\geq \bar{K}'$:
    \begin{align*}
        \frac{4C}{D_{\min}(1-\rho)(k-\tilde{\tau}-1)}\leq \max_{s,a}\sqrt{ \frac{\log [2(k+h)^2 D(s,a)^2/(h+1)^2]}{c_{mc} (k-\tilde{\tau}-1) D(s,a)^2}  }.
    \end{align*}
    Then, for any $k\geq \bar{K}'$, choosing 
    \begin{align*}
    &\delta = \max_{s,a}\sqrt{ \frac{\log [2(k+h)^2 D(s,a)^2/(h+1)^2]}{c_{mc} (k-\tilde{\tau}-1) D(s,a)^2}  },
\end{align*}
we have
\begin{align*}
    \frac{2(k+h)^2}{(h+1)^2}\exp(-c_{mc} (k-\tilde{\tau}-1)\delta^2D(s,a)^2)\leq \frac{1}{D(s,a)^2}.
\end{align*}
It follows that
\begin{align*}
    \mathbb{E}\left[\frac{(k+h)^2}{(N_{k-\tilde{\tau}-1}(s,a)+1+h)^2}\right]\leq \,&\frac{(k+h)^2}{(k-\tilde{\tau}-1)^2}\frac{1}{(1-\delta)^2D(s,a)^2}\\
    &+\frac{2(k+h)^2}{(h+1)^2}\exp(-c_{mc} (k-\tilde{\tau}-1)\delta^2D(s,a)^2)\\
    \leq \,&\frac{(k+h)^2}{(k-\tilde{\tau}-1)^2}\frac{1}{(1-\delta)^2D(s,a)^2}+\frac{1}{D(s,a)^2}.
\end{align*}
Let $\bar{K}''$ be such that the following inequality holds for all $k\geq \bar{K}''$:
\begin{align*}
    \frac{(k+h)^2}{(k-\tilde{\tau}-1)^2}\frac{1}{(1-\delta)^2}\leq 2.
\end{align*}
Then, for any $k\geq \bar{K}''$, we have
\begin{align*}
    \mathbb{E}\left[\frac{(k+h)^2}{(N_{k-\tilde{\tau}-1}(s,a)+1+h)^2}\right]\leq \frac{2}{D(s,a)^2}+\frac{1}{D(s,a)^2}
    \leq \frac{3}{D(s,a)^2}.
\end{align*}
Using the previous inequality in Equation (\ref{eq2:pf:T4}), we have
\begin{align*}
    \mathbb{E}\left[\frac{1}{D_k(S_k,A_k)^2}\right]
    \leq \sum_{s,a}\frac{6}{D(s,a)}
    \leq \frac{6|\mathcal{S}||\mathcal{A}|}{D_{\min}}.
\end{align*}
Finally, using the previous inequality in Equation (\ref{eq3:pf:T4}), we obtain for any $k\geq K_4:=\max(\bar{K}',\bar{K}'')$ that
\begin{align*}
    T_4\leq \frac{6|\mathcal{S}||\mathcal{A}|(1+b_k)^2}{D_{\min}}\leq \frac{6|\mathcal{S}||\mathcal{A}|(b_k+\mysp(Q^*)+1)^2}{D_{\min}}.
\end{align*}

\subsection{Proof of Proposition \ref{prop:recursion}}
Using the upper bounds we obtained for the terms $T_1$ (cf. Lemma \ref{le:T_1}), $T_2$ (cf. Proposition \ref{prop:T2}), $T_3$ (cf. Proposition \ref{prop:T3}), and $T_4$ (cf. Lemma \ref{le:T4}), in Equation (\ref{eq:decomposition_illustration}), we obtain for any $k\geq K=\max(K_2,K_3,K_4)$ that
\begin{align*}
    \mathbb{E}[M(Q_{k+1}-Q^*)]
    \leq  \,&\mathbb{E}[M(Q_k-Q^*)]+\alpha_k  (T_1+T_2+T_3)+\frac{L\alpha_k^2}{2}T_3\\
    \leq \,&\left(1-\left(1-\frac{\beta u_m}{\ell_m}\right)\alpha_k\right)\mathbb{E}\left[M(Q_k-Q^*)\right]\\
    &+\frac{35 |\mathcal{S}||\mathcal{A}|}{ D_{\min}^2}\left(\frac{C}{\ell_m^2(1-\beta u_m/\ell_m)(1-\rho)\alpha}+L\right)\tau_k(b_k+\mysp(Q^*)+1)^2\alpha_k\\
    = \,&\left(1-\phi_1\alpha_k\right)\mathbb{E}\left[M(Q_k-Q^*)\right]+35\phi_2\tau_k(b_k+\mysp(Q^*)+1)^2\alpha_k^2,
\end{align*}
where 
\begin{align*}
    \phi_2=\frac{ |\mathcal{S}||\mathcal{A}|}{ D_{\min}^2}\left(\frac{C}{\ell_m^2(1-\rho)\phi_1\alpha}+L\right).
\end{align*}

\subsection{Solving the Recursion}\label{ap:recursion}
This part is standard in the non-asymptotic analysis of SA algorithms. Repeatedly applying Proposition \ref{prop:recursion}, we have for any $k\geq K$ that
\begin{align*}
    \mathbb{E}[M(Q_k-Q^*)]
    \leq\,& \prod_{j=K}^{k-1}\left(1-\phi_1\alpha_j\right)\mathbb{E}[M(Q_K-Q^*)]\\
    &+35\phi_2\sum_{i=K}^{k-1}\tau_i (B_i+\mysp(Q^*)+1)^2\alpha_i^2\prod_{j=i+1}^{k-1}\left(1-\phi_1\alpha_j\right).
\end{align*}
To translate the result to a bound on $\mathbb{E}[\mysp(Q_k-Q^*)^2]$, using Lemma \ref{le:Moreau} (3), we have
\begin{align*}
    \frac{1}{2 u_m^2}\mysp(Q)^2\leq M(Q)=\frac{1}{2}p_m(Q)^2\leq \frac{1}{2 \ell_m^2}\mysp(Q)^2.
\end{align*}
It follows that
\begin{align*}
\mathbb{E}[\mysp(Q_k-Q^*)^2]\leq \,&2u_m^2\mathbb{E}[M(Q_k-Q^*)]\\
\leq \,&2u_m^2\prod_{j=K}^{k-1}\left(1-\phi_1\alpha_j\right)\mathbb{E}[M(Q_K-Q^*)]\\
    &+70u_m^2\phi_2\sum_{i=K}^{k-1}\tau_i (B_i+\mysp(Q^*)+1)^2\alpha_i^2\prod_{j=i+1}^{k-1}\left(1-\phi_1\alpha_j\right)\\
    \leq \,&\frac{u_m^2}{\ell_m^2}\prod_{j=K}^{k-1}\left(1-\phi_1\alpha_j\right)\mathbb{E}[\mysp(Q_K-Q^*)^2]\nonumber\\
    &+70u_m^2\phi_2\sum_{i=K}^{k-1}\tau_i (B_i+\mysp(Q^*)+1)^2\alpha_i^2\prod_{j=i+1}^{k-1}\left(1-\phi_1\alpha_j\right)\\
    \leq \,&\frac{u_m^2}{\ell_m^2}(b_K+\mysp(Q^*))^2\prod_{j=K}^{k-1}\left(1-\phi_1\alpha_j\right)\\
    &+70u_m^2\phi_2\tau_k (b_k+\mysp(Q^*)+1)^2\sum_{i=K}^{k-1}\alpha_i^2\prod_{j=i+1}^{k-1}\left(1-\phi_1\alpha_j\right),
\end{align*}
where the last inequality follows from 
\begin{align*}
    \mysp(Q_K-Q^*)\leq \mysp(Q_K)+\mysp(Q^*)\leq b_K+\mysp(Q^*).
\end{align*}
Before we proceed, we finalize the choices of the tunable parameters $q$ and $\theta$ in the definition of $M(\cdot)$ to make all constants on the right-hand side of the previous inequality explicit. Specifically, by choosing
\begin{align*}
    \theta=\left(\frac{1+\beta}{2\beta}\right)^2-1,\;q=2\log(|\mathcal{S}||\mathcal{A}|),
\end{align*}
we have $u_q=1$ and $\ell_q=d^{-1/q}=1/\sqrt{e}$, which further implies 
\begin{align*}
u_m^2=\,&1+\theta u_q^2= 1+\theta\leq \frac{1}{\beta^2}\leq 4.\tag{Assume without loss of generality that $\beta>1/2$.}\\
    \frac{u_m^2}{\ell_m^2}=\,&\frac{1+\theta u_q^2}{1+\theta \ell_q^2}= \frac{1+\theta}{1+\theta/\sqrt{e}}\leq \sqrt{e}\leq 3,\\
    \phi_1=\,& 1-\frac{\beta u_m}{\ell_m}\geq 1-\beta\frac{1+\beta}{2\beta}= \frac{1-\beta}{2},\\
    L=\,&\frac{q-1}{\theta \ell_q^2}\leq \frac{6\log(|\mathcal{S}||\mathcal{A}|)}{1-\beta},\\
    70u_m^2\phi_2=\,&\frac{ 70u_m^2|\mathcal{S}||\mathcal{A}|}{ D_{\min}^2}\left(\frac{C_{mc}}{\ell_m^2(1-\beta u_m/\ell_m)\alpha}+L\right)\\
    \leq \,&\frac{ 280|\mathcal{S}||\mathcal{A}|}{ D_{\min}^2}\left(\frac{2C}{(1-\rho)(1-\beta)\alpha}+\frac{6\log(|\mathcal{S}||\mathcal{A}|)}{1-\beta}\right)\\
    \leq \,&\frac{2240C|\mathcal{S}||\mathcal{A}|\log(|\mathcal{S}||\mathcal{A}|)}{(1-\beta)(1-\rho)D_{\min}^2\min(1,\alpha)}.
\end{align*}
Therefore, we have for all $k\geq K$ that
\begin{align}
   & \mathbb{E}[\mysp(Q_k-Q^*)^2]\leq  3(b_K+\mysp(Q^*))^2\underbrace{\prod_{j=K}^{k-1}\left(1-\frac{(1-\beta)\alpha_j}{2}\right)}_{:=E_1}\nonumber\\
    &\;+\frac{2240C|\mathcal{S}||\mathcal{A}|\log(|\mathcal{S}||\mathcal{A}|)}{(1-\beta)(1-\rho) D_{\min}^2\min(1,\alpha)}\tau_k (b_k+\mysp(Q^*)+1)^2\underbrace{\sum_{i=K}^{k-1}\alpha_i^2\prod_{j=i+1}^{k-1}\left(1-\frac{(1-\beta)\alpha_j}{2}\right)}.\label{eq:solving_recursion}
\end{align}

It remains to bound the terms $E_1$ and $E_2$. For simplicity of notation, denote $c=(1-\beta)/2$. Then, since $\alpha_k=\alpha/(k+h)$, we have 
\begin{align*}
	E_1=\,&\prod_{j=K}^{k-1}\left(1-c\alpha_j\right)\\
	\leq \,&\exp\left(-\sum_{j=K}^{k-1}c\alpha_j\right)\\
	=\,&\exp\left(-\sum_{j=K}^{k-1}\frac{c \alpha}{j+h}\right)\\
	\leq \,&\exp\left(-\int_{K}^{k}\frac{c \alpha}{x+h}dx\right)\\
	=\,&\exp\left(-c \alpha\log\left(\frac{k+h}{K+h}\right)\right)\\
	=\,&\left(\frac{K+h}{k+h}\right)^{c\alpha}.
\end{align*}
Similarly, we also have
\begin{align*}
	E_2=\,&\sum_{i=K}^{k-1}\alpha_i^2\prod_{j=i+1}^{k-1}\left(1-c\alpha_j\right)\\
	\leq \,&\sum_{i=K}^{k-1}\alpha_i^2\prod_{j=i+1}^{k-1}\left(1-c\alpha_j\right)\\
	\leq \,&\sum_{i=K}^{k-1}\frac{\alpha^2}{(i+h)^2}\left(\frac{i+1+h}{k+h}\right)^{c\alpha}\\
	\leq  \,&\frac{4\alpha^2}{(k+h)^{c\alpha}}\sum_{i=K}^{k-1}\frac{1}{(i+1+h)^{2-c\alpha}}\\
	\leq  \,&\frac{4\alpha^2}{(k+h)^{c\alpha}}\begin{dcases}
		\frac{1}{1-c\alpha} &c\alpha\in (0,1),\\
		\log(k+h)&c\alpha=1,\\
		\frac{(k+1+h)^{c\alpha-1}}{c\alpha-1}&c\alpha\in (1,\infty).
	\end{dcases}\\
	\leq \,&\begin{dcases}
		\frac{1}{(k+h)^{c\alpha}}\frac{4\alpha^2}{1-c\alpha} &c\alpha\in (0,1),\\
		\frac{4\log(k+h)}{c^2(k+h)}&c\alpha=1,\\
		\frac{1}{(k+h)}\frac{12\alpha^2}{c\alpha-1}&c\alpha\in (1,\infty).
	\end{dcases}
\end{align*}
Using the previous two inequalities together in Equation (\ref{eq:solving_recursion}) and recalling that $c=(1-\beta)/2$, we finally obtain the desired finite-time bound:
\begin{enumerate}[(1)]
    \item When $\alpha(1-\beta)<2$, we have
    \begin{align*}
        &\mathbb{E}[\mysp(Q_k-Q^*)^2]\leq 3(b_K+\mysp(Q^*))^2\left(\frac{K+h}{k+h}\right)^{\frac{\alpha(1-\beta)}{2}}\\
        &\quad +\frac{17920\alpha^2C|\mathcal{S}||\mathcal{A}|\log(|\mathcal{S}||\mathcal{A}|)}{(1-\beta) (1-\rho)D_{\min}^2\min(1,\alpha)(2-(1-\beta)\alpha)}\frac{\tau_k (b_k+\mysp(Q^*)+1)^2}{(k+h)^{\alpha(1-\beta)/2}}.
    \end{align*}
    \item When $\alpha(1-\beta)=2$, we have
    \begin{align*}
        &\mathbb{E}[\mysp(Q_k-Q^*)^2]\leq 3(b_K+\mysp(Q^*))^2\left(\frac{K+h}{k+h}\right)\\
        &\quad +\frac{35840C|\mathcal{S}||\mathcal{A}|\log(|\mathcal{S}||\mathcal{A}|)}{(1-\rho)(1-\beta)^3 D_{\min}^2}\frac{\tau_k (b_k+\mysp(Q^*)+1)^2\log(k+h)}{(k+h)}.
    \end{align*}
    \item When $\alpha(1-\beta)>2$, we have
    \begin{align*}
        &\mathbb{E}[\mysp(Q_k-Q^*)^2]\leq 3(b_K+\mysp(Q^*))^2\left(\frac{K+h}{k+h}\right)^{\frac{\alpha(1-\beta)}{2}}\\
        &\quad +\frac{53760\alpha^2C|\mathcal{S}||\mathcal{A}|\log(|\mathcal{S}||\mathcal{A}|)}{(1-\rho)(1-\beta) D_{\min}^2((1-\beta)\alpha-2)}\frac{\tau_k (b_k+\mysp(Q^*)+1)^2}{(k+h)}.
    \end{align*}
\end{enumerate}
The proof of Theorem \ref{thm:QL-set} is complete.

\subsection{Proofs of Auxiliary Lemmas}
\subsubsection{Proof of Lemma \ref{le:time-varying-bound}}\label{pf:le:time-varying-bound}
Since $Q_{k+1}-\tilde{Q}_{k+1}\in\text{ker}(\mysp)$, there exists $c_k$ such that
\begin{align*}
    Q_{k+1}-\tilde{Q}_{k+1}=c_ke.
\end{align*}
To show the desired inequality, it suffices to prove that
\begin{align}
    \max_{s,a}Q_{k+1}(s,a) \leq\, & \max_{s,a}Q_k(s,a) + \alpha_k(S_k,A_k)+c_k,\label{eq1:pf:time-varying-bound}\\
    \min_{s,a}Q_{k+1}(s,a) \geq\, & \min_{s,a}Q_k(s,a) - \alpha_k(S_k,A_k)+c_k.\label{eq2:pf:time-varying-bound}
\end{align}
The result then follows from the definition of the span seminorm:
\begin{align*}
    \mysp(Q_{k+1})=\,&\frac{1}{2}\left(\max_{s,a}Q_{k+1}(s,a)-\min_{s,a}Q_{k+1}(s,a)\right)\\
    \leq \,&\frac{1}{2}\left(\max_{s,a}Q_k(s,a)-\min_{s,a}Q_k(s,a)\right)+\alpha_k(S_k,A_k)\\
    =\,&\mysp(Q_k)+\alpha_k(S_k,A_k).
\end{align*}
In the following, we will prove Equation (\ref{eq1:pf:time-varying-bound}); the proof of Equation (\ref{eq2:pf:time-varying-bound}) follows from an identical argument.

In view of Algorithm \ref{alg:QL-generic}, when $(s,a) \neq (S_k,A_k)$, we have
\begin{align*}
    Q_{k+1}(s,a) = Q_k(s,a)+c_k \leq \max_{s,a}Q_k(s,a) + \alpha_k(S_k,A_k)+c_k.
\end{align*}
When $(s,a) = (S_k,A_k)$, since $\max_{s,a}|\mathcal{R}(s,a)|\leq 1$, we have
\begin{align*}
    Q_{k+1}(s,a) =\, & Q_k(s,a) + \alpha_k(S_k,A_k)\left(\mathcal{R}(S_k,A_k) + \max_{a'\in\mathcal{A}}Q_k(S_{k+1},a') - Q_k(S_k,A_k)\right)+c_k\\
    =\, & \left(1 - \alpha_k(S_k,A_k)\right)Q_k(S_k,A_k) + \alpha_k(S_k,A_k)\left(\mathcal{R}(s,a) + \max_{a'\in\mathcal{A}}Q_{k+1}(S_{k+1},a')\right)+c_k\\
    \leq\, & \left(1 - \alpha_k(S_k,A_k)\right)\max_{s',a'}Q_k(s',a') + \alpha_k(S_k,A_k)\left(1 + \max_{s',a'}Q_k(s',a')\right)+c_k\\
    =\, & \max_{s',a'}Q_k(s',a') + \alpha_k(S_k,A_k)+c_k.
\end{align*}
Combining both cases, we obtain
\begin{align*}
    \max_{s,a}Q_{k+1}(s,a) \leq \max_{s,a}Q_k(s,a) + \alpha_k(S_k,A_k)+c_k,
\end{align*}
which establishes Equation (\ref{eq1:pf:time-varying-bound}).

\subsubsection{Proof of Lemma \ref{le:difference}}\label{pf:le:difference}
	Using the definition of $\mysp(\cdot)$, we have by Algorithm \ref{alg:QL-generic} that
	\begin{align}
		\mysp(Q_{k+1})-\mysp(Q_k)\leq\,& \mysp(Q_{k+1}-Q_k)\nonumber\\
        =\,&\frac{\alpha_k(S_k,A_k)}{2}\left|\mathcal{R}(S_k,A_k)+\max_{a'\in\mathcal{A}}Q_k(S_{k+1},a')-Q_k(S_k,A_k)\right|\nonumber\\
        \leq \,&\frac{\alpha_k(S_k,A_k)}{2}\left(\left|\mathcal{R}(S_k,A_k)\right|+\left|\max_{a'\in\mathcal{A}}Q_k(S_{k+1},a')-Q_k(S_k,A_k)\right|\right)\nonumber\\
        \leq \,&\frac{\alpha_k(S_k,A_k)}{2}\left(\left|\mathcal{R}(S_k,A_k)\right|+\max_{s',a'}Q_k(s',a')-\min_{s',a'}Q_k(s',a')\right)\nonumber\\
        \leq \,&\frac{\alpha_k(S_k,A_k)}{2}(1+2\mysp(Q_k))\nonumber\\
        \leq  \,&\alpha_k(S_k,A_k)\left(\mysp(Q_k)+1\right).\label{eq1:le:difference}
	\end{align}
	Rearranging terms, we obtain
	\begin{align*}
		\mysp(Q_{k+1})+1\leq \left(1+\alpha_k(S_k,A_k)\right)\left(\mysp(Q_k)+1\right).
	\end{align*}
	Therefore, we have for all $k\geq k_1$ that
	\begin{align*}
    \mysp(Q_k) 
    \leq\, & \prod_{j=k_1}^{k-1} \left(1 + \alpha_j(S_j,A_j)\right) \left(\mysp(Q_{k_1}) + 1\right) - 1 \\
    =\, & \prod_{j=k_1}^{k-1} \left(1 + \frac{\alpha}{N_j(S_j,A_j)+h}\right) \left(\mysp(Q_{k_1}) + 1\right) - 1 \\
    =\, & \prod_{j=k_1}^{k-1} \left(1 + \frac{\alpha}{N_{j-1}(S_j, A_j) + 1+h}\right) \left(\mysp(Q_{k_1}) + 1\right) - 1 \tag{$(S_j,A_j)$ is visited in the $j$-th iteration}\\
    \leq\, & \prod_{j=k_1}^{k-1} \left(1 + \frac{\alpha}{N_{k_1 - 1, \min} + 1+h}\right) \left(\mysp(Q_{k_1}) + 1\right) - 1 \\
    \leq\, & \exp\left(\frac{\alpha(k - k_1)}{N_{k_1 - 1, \min} + 1+h}\right) \left(\mysp(Q_{k_1}) + 1\right) - 1,
\end{align*}
where the last line follows from $(1+x)\leq e^x$ for all $x\in\mathbb{R}$.
Rearranging terms, we have
\begin{align*}
    \mysp(Q_k)+1\leq \,&\exp\left(\frac{\alpha(k - k_1)}{N_{k_1 - 1, \min} + 1+h}\right) \left(\mysp(Q_{k_1}) + 1\right),\quad \forall\,k\geq k_1.
\end{align*}
Using the previous inequality in Equation (\ref{eq1:le:difference}), we have
\begin{align*}
    \mysp(Q_{k+1}-Q_k)\leq\,& \alpha_k(S_k,A_k)(\mysp(Q_k)+1)\\
    \leq\,& \frac{\alpha}{(N_{k_1-1,\min}+1+h)}\exp\left(\frac{\alpha(k - k_1)}{N_{k_1 - 1, \min} + 1+h}\right)\left(\mysp(Q_{k_1})+1\right).
\end{align*}
By telescoping, we have for any $k\geq k_1$ that
\begin{align*}
    \mysp(Q_k-Q_{k_1})\leq\,& \sum_{i=k_1}^{k-1}\mysp(Q_{i+1}-Q_i)\\
    \leq\,& \sum_{i=k_1}^{k-1}\frac{\alpha}{N_{k_1-1,\min}+1+h}\exp\left(\frac{\alpha(i - k_1)}{N_{k_1 - 1, \min} + 1+h}\right)(\mysp(Q_{k_1})+1)\\
    \leq\,& \frac{\alpha(k-k_1)}{N_{k_1-1,\min}+1+h}\exp\left(\frac{\alpha(k - k_1)}{N_{k_1 - 1, \min} + 1+h}\right)(\mysp(Q_{k_1})+1)\\
    =\,&f\left(\frac{\alpha(k - k_1)}{N_{k_1 - 1, \min} + 1+h}\right)(\mysp(Q_{k_1})+1),
\end{align*}
where the last line follows from the definition of $f(\cdot)$.

\subsubsection{Proof of Lemma \ref{le:MC-MSE}}\label{pf:le:MC-MSE}
    For simplicity of notation, let $X_k(s,a)=\mathbf{1}_{\{(S_k,A_k)=(s,a)\}}$ for all $k\geq 1$. By definition of $\bar{D}_k(s,a)$, we have
\begin{align}
    \mathbb{E}\left[(\bar{D}_{k}(s,a)-D(s,a))^2\right]=\,&\mathbb{E}\left[(\bar{D}_{k}(s,a)-\mathbb{E}[\bar{D}_{k}(s,a)]+\mathbb{E}[\bar{D}_{k}(s,a)]-D(s,a))^2\right]\nonumber\\
    =\,&\underbrace{\mathbb{E}\left[(\bar{D}_{k}(s,a)-\mathbb{E}[\bar{D}_{k}(s,a)])^2\right]}_{\text{Variance}}+\underbrace{(\mathbb{E}[\bar{D}_{k}(s,a)]-D(s,a))^2}_{\text{bias}^2}.\label{eq:le:MC-MSE:bias-variance}
\end{align}
We first bound the bias term. Observe that
\begin{align*}   \text{bias}^2=\left(\frac{\sum_{i=1}^k(\mathbb{E}[X_i(s,a)]-D(s,a))}{k}\right)^2
    \leq \frac{1}{k^2}\left(\sum_{i=1}^k|\mathbb{E}[X_i(s,a)]-D(s,a)|\right)^2.
\end{align*}
Moreover, under Assumption \ref{as:MC}, we have
\begin{align}
    |\mathbb{E}[X_i(s,a)]-D(s,a)|= \,&|\mathbb{P}(S_i=s,A_i=a| S_1)-\mu(s)\pi(a|s)|\nonumber\\
    =\,&|p_\pi(S_i=s| S_1)-\mu(s)|\pi(a|s)\nonumber\\
    \leq \,&\max_{s'}\sum_{s}|p_\pi(S_i=s| S_1=s')-\mu(s)|\nonumber\\
    \leq \,&2C\rho^{i-1}.\label{eq1:le:MC-MSE}
\end{align}
It follows that 
\begin{align}\label{eq:bias:le:MC-MSE}
    \text{bias}^2\leq \frac{1}{k^2}\left(\sum_{i=1}^k2C\rho^{i-1}\right)^2\leq \frac{4C^2}{k^2(1-\rho)^2}.
\end{align}
Next, we turn to the variance term on the right-hand side of Eq (\ref{eq:le:MC-MSE:bias-variance}). Observe that
\begin{align}
    \text{Variance}=\,&\frac{1}{k^2}\text{Var}\left(\sum_{i=1}^kX_i(s,a)\right)\nonumber\\
    =\,&\frac{1}{k^2}\sum_{i=1}^k\text{Var}(X_i(s,a))+\frac{2}{k^2}\sum_{1\leq i<j\leq k}\text{Cov}(X_i(s,a),X_j(s,a))\label{eq:le:MC-MSE:variance}
\end{align}
On the one hand, we have
\begin{align}
    \text{Var}(X_i(s,a))=\,&\mathbb{E}[X_i(s,a)^2]-(\mathbb{E}[X_i(s,a)])^2\nonumber\\
    = \,&\mathbb{E}[X_i(s,a)]-(\mathbb{E}[X_i(s,a)])^2\nonumber\\
    \leq \,&\mathbb{E}[X_i(s,a)]\nonumber\\
    \leq \,&|\mathbb{E}[X_i(s,a)]-D(s,a)|+D(s,a)\nonumber\\
    \leq \,&2C\rho^{i-1}+D(s,a),\label{eq2:le:MC-MSE}
\end{align}
where the last line follows from Equation (\ref{eq1:le:MC-MSE}).
On the other hand, we have
\begin{align*}
    \text{Cov}(X_i(s,a),X_j(s,a))=\,&\mathbb{E}[X_i(s,a)X_j(s,a)]-\mathbb{E}[X_i(s,a)]\mathbb{E}[X_j(s,a)]\\
    =\,&\mathbb{P}(X_i(s,a)=1,X_j(s,a)=1)-\mathbb{P}(X_i(s,a)=1)\mathbb{P}(X_j(s,a)=1)\\
    =\,&\mathbb{P}(X_i(s,a)=1)(\mathbb{P}(X_j(s,a)=1| X_i(s,a)=1)-\mathbb{P}(X_j(s,a)=1))\\
    \leq \,&\mathbb{P}(X_i(s,a)=1)|\mathbb{P}(X_j(s,a)=1| X_i(s,a)=1)-\mathbb{P}(X_j(s,a)=1)|.
\end{align*}
Using the same analysis as in Equation (\ref{eq1:le:MC-MSE}), we obtain
\begin{align*}
    \mathbb{P}(X_i(s,a)=1)\leq |\mathbb{P}(X_i(s,a)=1)-D(s,a)|+D(s,a)\leq 2C\rho^{i-1}+D(s,a)
\end{align*}
and
\begin{align*}
    &|\mathbb{P}(X_j(s,a)=1| X_i(s,a)=1)-\mathbb{P}(X_j(s,a)=1)|\\
    \leq \,&|\mathbb{P}(X_j(s,a)=1| X_i(s,a)=1)-D(s,a)|+|D(s,a)-\mathbb{P}(X_j(s,a)=1)|\\
    \leq \,&2C\rho^{j-i-1}+2C\rho^{j-1}\\
    \leq \,&4C\rho^{j-i-1}.
\end{align*}
Combining the previous three inequalities together, we have
\begin{align*}
    \text{Cov}(X_i(s,a),X_j(s,a))\leq \,&(C\rho^{i-1}+D(s,a))4C\rho^{j-i-1}\\
    =\,&4C^2\rho^{j-2}+D(s,a)4C\rho^{j-i-1}.
\end{align*}
Using the previous inequality and Equation (\ref{eq2:le:MC-MSE}) together in Equation (\ref{eq:le:MC-MSE:variance}), we have
\begin{align*}
    \text{Variance}
    =\,&\frac{1}{k^2}\sum_{i=1}^k\text{Var}(X_i(s,a))+\frac{2}{k^2}\sum_{1\leq i<j\leq k}\text{Cov}(X_i(s,a),X_j(s,a))\\
    \leq \,&\frac{1}{k^2}\sum_{i=1}^k2C\rho^{i-1}+\frac{D(s,a)}{k}+\frac{2}{k^2}\sum_{i=1}^k\sum_{j=i+1}^k(4C^2\rho^{j-2}+D(s,a)4C\rho^{j-i-1})\\
    \leq \,&\frac{2C}{k^2(1-\rho)}+\frac{D(s,a)}{k}+\frac{2}{k^2}\frac{4C^2}{(1-\rho)^2}+\frac{8C}{k(1-\rho)}D(s,a)\\
    \leq \,&\frac{10C^2}{k^2(1-\rho)^2}+\frac{9CD(s,a)}{k(1-\rho)},
\end{align*}
where the last line follows from $C\geq 1$ and $\rho\in (0,1)$. 

Finally, using the previous inequality and Equation (\ref{eq:bias:le:MC-MSE}) together in Equation (\ref{eq:le:MC-MSE:bias-variance}), we obtain
\begin{align*}
    \mathbb{E}\left[(\bar{D}_{k}(s,a)-D(s,a))^2\right]=\,&\text{variance}+\text{bias}^2\\
    \leq \,&\frac{14C^2}{k^2(1-\rho)^2}+\frac{9CD(s,a)}{k(1-\rho)}\\
    \leq \,&\frac{10CD(s,a)}{(1-\rho)k},
\end{align*}
where the last inequality follows from $k\geq 14C/[(1-\rho)D_{\min}]$.

\section{Numerical Simulations}\label{sec:simulations}
This section presents our numerical simulations.

\subsection{Verifying Proposition \ref{prop:universal-convergence}}

To demonstrate the negative result presented in Proposition \ref{prop:universal-convergence}, we construct an average-reward MDP that has two states $s_1,s_2$ and two actions $a_1,a_2$. The transition probabilities and the reward function are defined in the following:
\begin{align*}
    p(s_1 \mid s_1, a_1) =\, &\frac{1}{5},& \quad p(s_2 \mid s_1, a_1) =\, &\frac{4}{5}, &\quad p(s_1 \mid s_1, a_2) =\, &\frac{4}{5}, &\quad p(s_2 \mid s_1, a_2) =\, &\frac{1}{5},\\
    p(s_1 \mid s_2, a_1) =\, &\frac{1}{2}, &\quad p(s_2 \mid s_2, a_1) =\, &\frac{1}{2}, &\quad p(s_1 \mid s_2, a_2) =\, &\frac{1}{2}, &\quad p(s_2 \mid s_2, a_2) =\, &\frac{1}{2},\\
    \mathcal{R}(s_1, a_1) =\, &1, &\quad \mathcal{R}(s_1, a_2) =\, &1, &\quad \mathcal{R}(s_2, a_1) =\, &2, &\quad \mathcal{R}(s_2, a_2) =\, &3.
\end{align*}
One can easily verify that, in this example, we have $\max_{(s,a),(s',a')}\|p(\cdot| s,a)-p(\cdot| s',a')\|_{\text{TV}}=0.6$. Therefore, Assumption \ref{as:seminorm-contraction} is satisfied with $\beta=0.6$.

\paragraph{The Optimal Policy and the Optimal Value.}
To identify an optimal policy $\pi^*$ and the optimal value $r^*$ in this example, observe that we must have $\pi^*(a_2\mid s_2)=1$ since the transition from state $s_2$ is independent of the actions, and $\mathcal{R}(s_2,a_2)=3>\mathcal{R}(s_2,a_1)=2$. As for $\pi^*(\cdot|s_1)$, similarly, we must have $\pi^*(a_1\mid s_1)=1$ because the rewards are the same for both actions at state $s_1$, but taking action $a_1$ results in a higher probability of going to state $s_2$, where the agent can achieve higher rewards. In summary, for the MDP example described above, there is a unique optimal policy $\pi^*$, which always takes action $a_1$ at state $s_1$ and action $a_2$ at state $s_2$. To compute $r^*$, with straightforward calculation, we see that the Markov chain induced by the optimal policy $\pi^*$ has a unique stationary distribution, which is given by $\mu(s_1)=5/13$ and $\mu(s_2)=8/13$. Therefore, the optimal value $r^*$ is given by $r^*=\sum_{s,a}\mu(s)\pi(a|s)\mathcal{R}(s,a)=29/13$.

\begin{figure}[ht]
    \centering
    \begin{minipage}{0.45\textwidth}
        \centering
        \includegraphics[width=\linewidth]{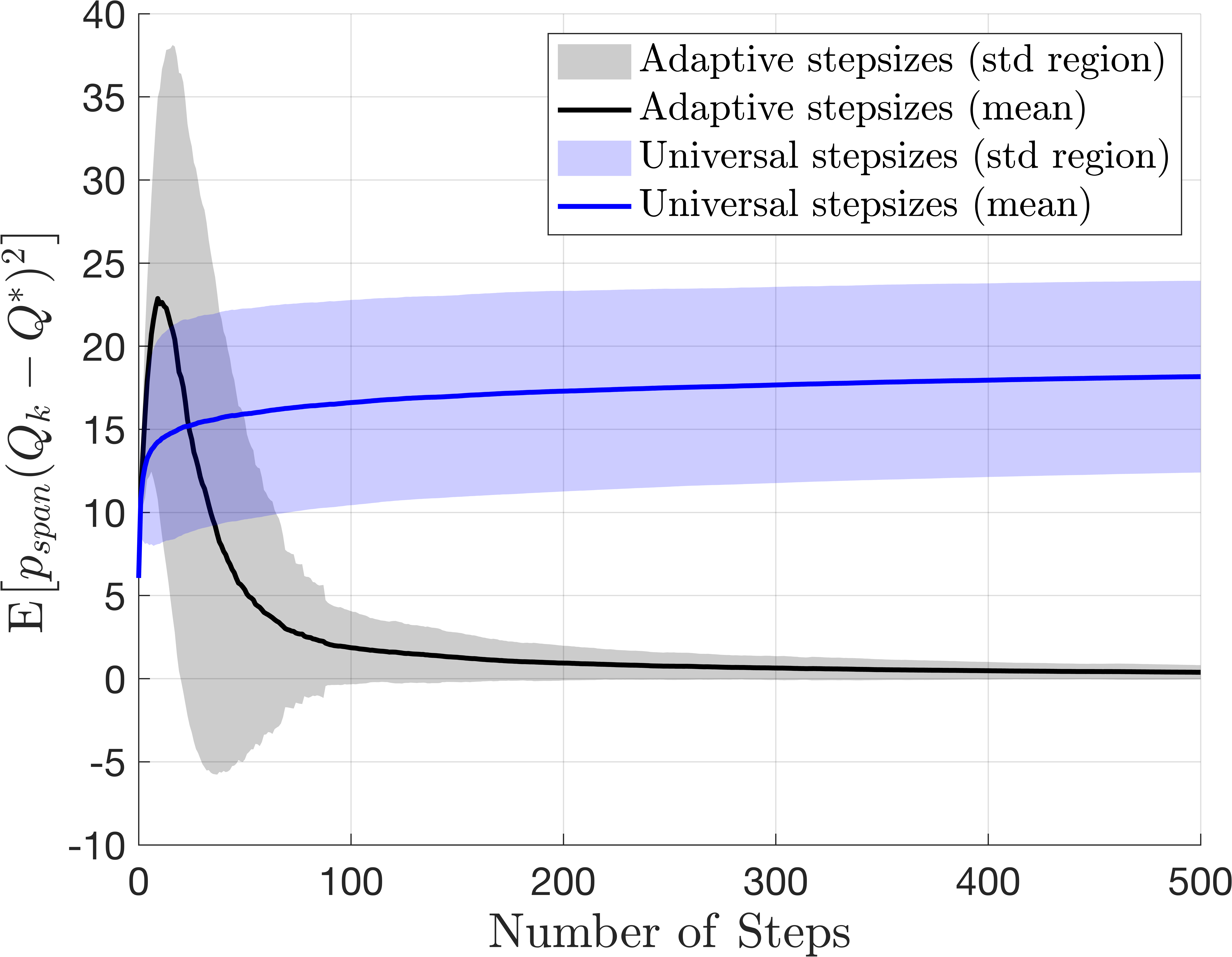}
        \caption{Convergence to $Q^*$ in $\mysp(\cdot)$}
        \label{fig:qlearning1}
    \end{minipage}
    \hfill
    \begin{minipage}{0.45\textwidth}
        \centering
        \includegraphics[width=\linewidth]{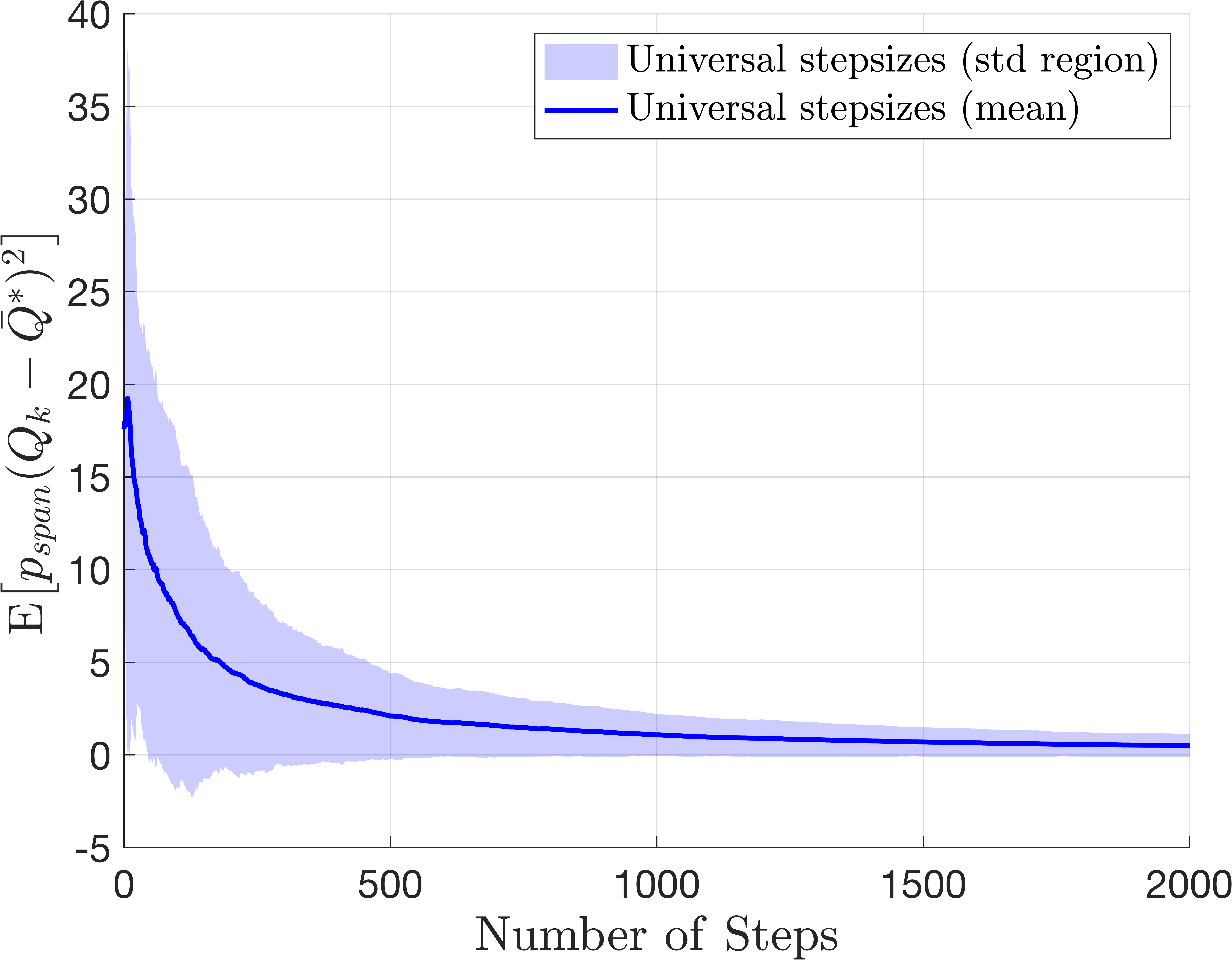}
        \caption{Convergence to $\bar{Q}^*$ in $\mysp(\cdot)$}
        \label{fig:qlearning2}
   \end{minipage}
\end{figure}

\begin{figure}[ht]
    \centering
    \begin{minipage}{0.45\textwidth}
        \centering
        \includegraphics[width=\linewidth]{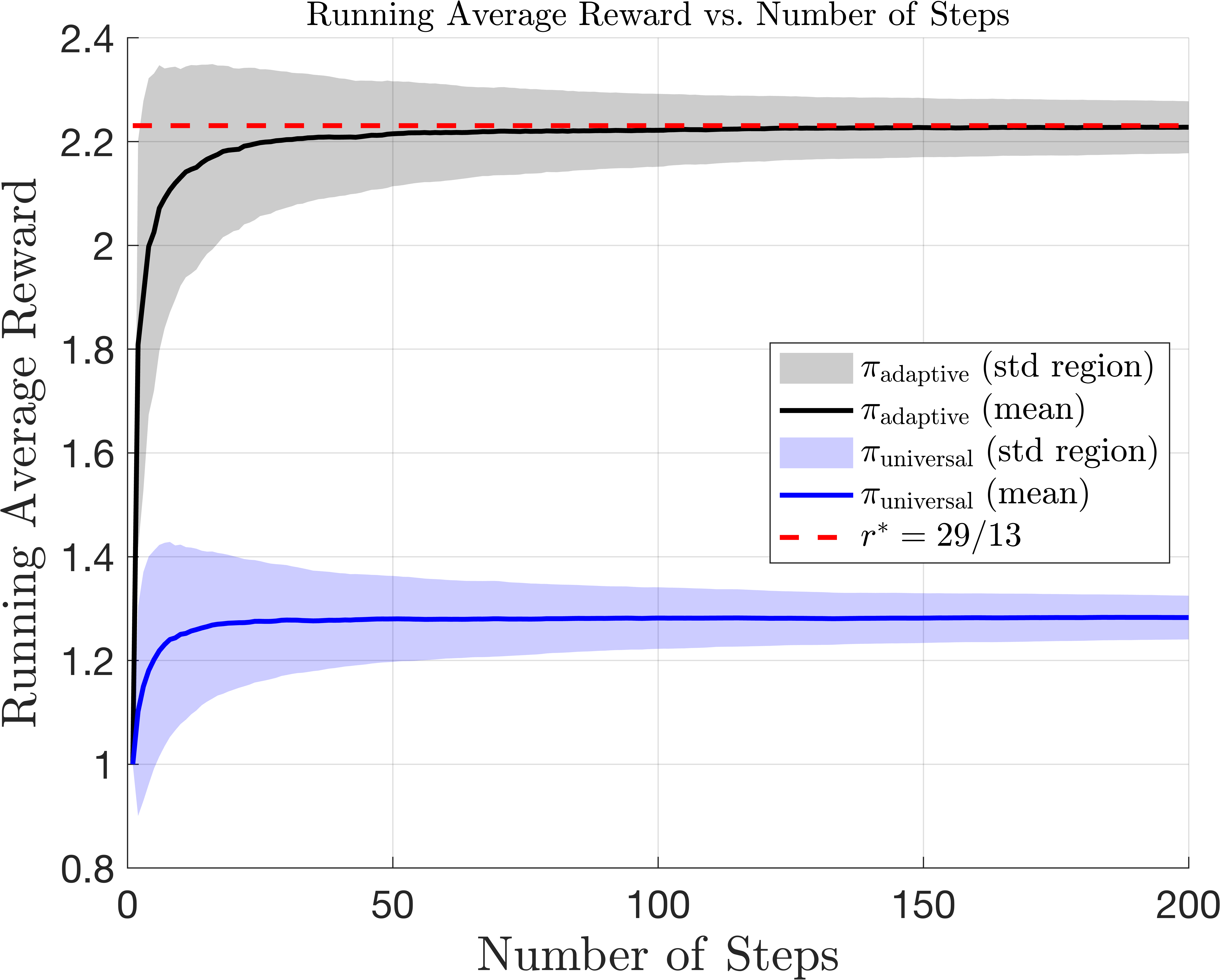}
        \caption{Performance of the Output Policies}
        \label{fig:qlearning3}
    \end{minipage}
    \hfill
    \begin{minipage}{0.45\textwidth}
        \centering
        \includegraphics[width=\linewidth]{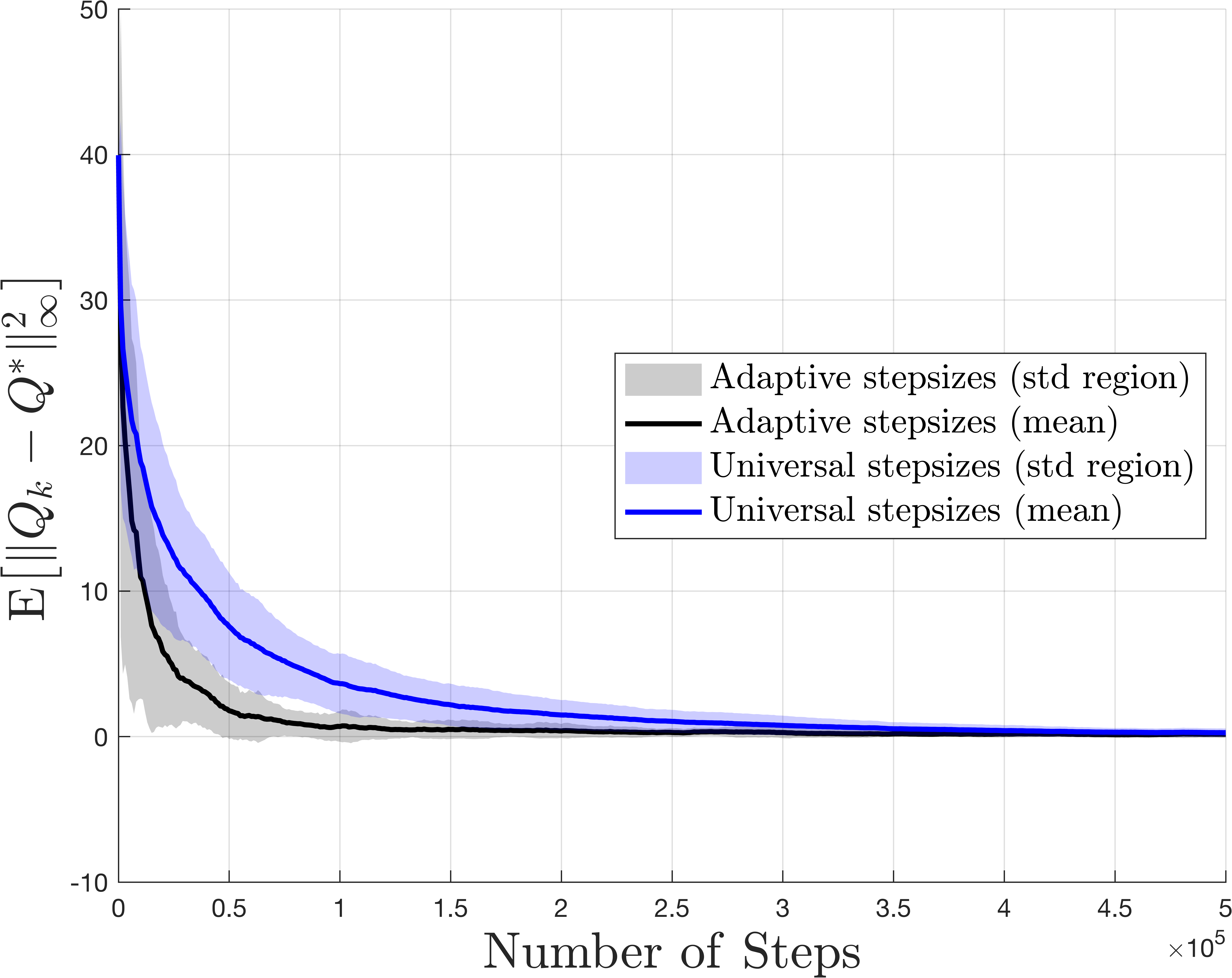}
        \caption{Discounted $Q$-Learning}
        \label{fig:qlearning4}
   \end{minipage}
\end{figure}

In our first simulation, we run $Q$-learning with universal stepsizes (cf. Equation  (\ref{algo:illustration})) and $Q$-learning with adaptive stepsizes (cf. Algorithm \ref{alg:QL-set}) on the MDP described above, where we use the same behavior policy $\pi$ defined as $\pi(a_1\mid s_1)=1/5$, $\pi(a_2\mid s_1)=4/5$, $\pi(a_1\mid s_2)=4/5$, and $\pi(a_2\mid s_2)=1/5$. It is clear that the Markov chain $\{S_k\}$ induced by $\pi$ is uniformly ergodic because all entries of the transition matrix are strictly positive. From Figure \ref{fig:qlearning1}, we observe that $Q$-learning with adaptive stepsizes converges to $Q^*$ in $\mysp(\cdot)$, whereas $Q$-learning with universal stepsizes fails to converge to $Q^*$. 

In our second simulation, we compute a solution $\bar{Q}^*$ to the asynchronous Bellman equation $\mysp(\bar{\mathcal{H}}(Q)-Q)=0$ using the seminorm fixed-point iteration \cite{chen2025non} and then plot $\mathbb{E}[\mysp(Q_k-\bar{Q}^*)^2]$ for $Q$-learning with universal stepsizes. From Figure \ref{fig:qlearning2}, we see that $Q$-learning with universal stepsizes actually converges to $\bar{Q}^*$ in $\mysp(\cdot)$, verifying Proposition \ref{prop:universal-convergence}.

Although $Q$-learning with universal stepsizes does not converge to $Q^*$ under $\mysp(\cdot)$, the algorithm remains acceptable if the policy greedily induced from $\bar{Q}^*$ is optimal. Therefore, in our third simulation, we compare the policies $\pi_{\text{adaptive}}$ and $\pi_{\text{universal}}$, which are greedily induced from the final iterates of $Q$-learning with adaptive and universal stepsizes, respectively. In both cases, we plot the running average of the rewards, $\mathbb{E}[\sum_{i=1}^k \mathcal{R}(S_i, A_i)]/k$,
as a function of $k$. From Figure \ref{fig:qlearning3}, we see that $\pi_{\text{adaptive}}$ is actually optimal as it converges to the optimal value $r^*$, whereas $\pi_{\text{universal}}$ is far from being optimal. This result further verifies the necessity of using adaptive stepsizes in average-reward $Q$-learning.

In our last simulation, we consider the exact same MDP but with a discount factor $\gamma = 0.99$. Figure \ref{fig:qlearning4} shows that in the discounted setting (even when $\gamma$ is close to one), both $Q$-learning with universal stepsizes and $Q$-learning with adaptive stepsizes converge to the optimal $Q$-function $Q_\gamma^*$. In addition, the convergence rate for $Q$-learning with adaptive stepsizes appears to be faster. A theoretical investigation of this phenomenon is an interesting future direction for this work.

\subsection{Convergence Behavior of Q-Learning under Different Adaptive Stepsizes}

In this section, we present numerical simulations for Q-learning under different choices of adaptive stepsizes, as discussed in Section \ref{subsec:adaptive_stepsizes}. 
\paragraph{The MDP.}
We construct an MDP with $|\mathcal{S}| = 20$ and $|\mathcal{A}| = 10$, where both the reward function and the transition probabilities are generated randomly. The behavior policy $\pi$ is chosen to be uniform, i.e., $\pi(a \mid s) = 1/|\mathcal{A}|$ for every $s \in \mathcal{S}$ and $a \in \mathcal{A}$. To ensure that Assumption \ref{as:seminorm-contraction} is satisfied, it is enough to require $\max_{(s,a),(s',a')} \|p(\cdot \mid s,a) - p(\cdot \mid s',a')\|_{\text{TV}} < 1$, which is equivalent to $\min_{(s,a),(s',a')} \sum_{s'' \in \mathcal{S}} \min\{ p(s'' \mid s, a), p(s'' \mid s', a') \} > 0$ \cite[Proposition 6.6.1]{puterman2014markov}. This condition is clearly satisfied by the MDP we constructed, since $p(s' \mid s, a) > 0$ for every $s, s' \in \mathcal{S}$ and $a \in \mathcal{A}$. As for Assumption \ref{as:MC}, because $\pi(a \mid s) > 0$ and $p(s' \mid s,a) > 0$ for every $s, s' \in \mathcal{S}$ and $a \in \mathcal{A}$, the Markov chain $\{(S_k,A_k)\}$ induced by $\pi$ is irreducible and aperiodic.

\begin{figure}[ht]
    \centering
    \begin{minipage}{0.45\textwidth}
        \centering
        \includegraphics[width=\linewidth]{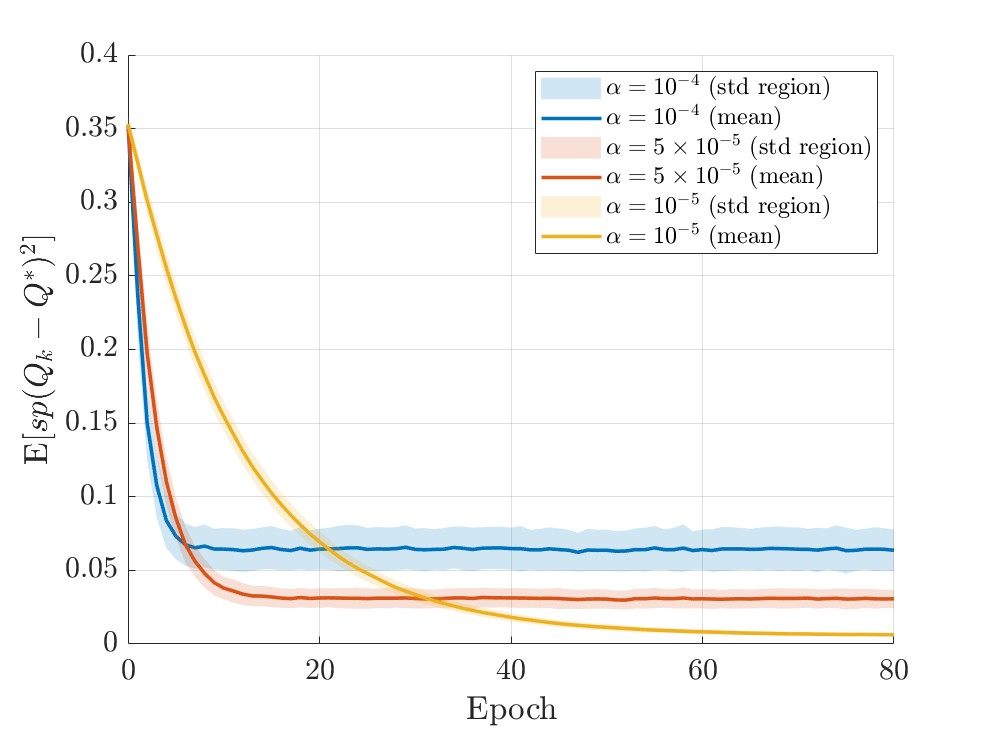}
        \caption{Adaptive Stepsize $\alpha_k(s,a) = \frac{\alpha (k + h)}{N_k(s,a) + h}$}
        \label{fig:const-step}
    \end{minipage}
    \hfill
    \begin{minipage}{0.45\textwidth}
        \centering
        \includegraphics[width=\linewidth]{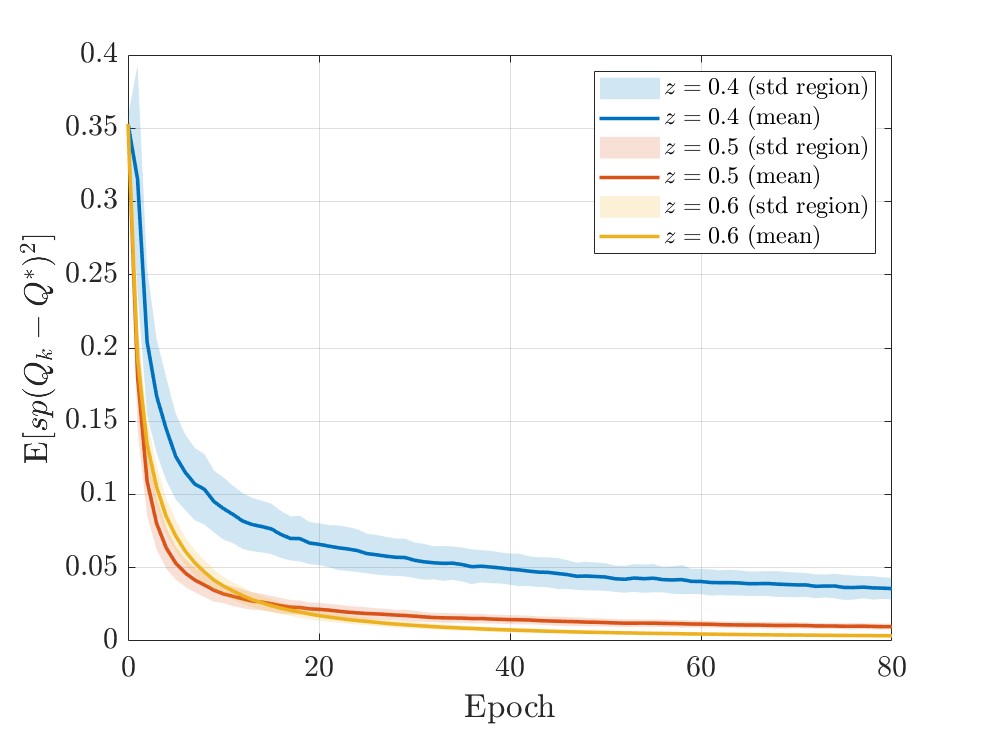}
        \caption{Adaptive Stepsize $\alpha_k(s,a) = \frac{\alpha (k + h)^{1 - z}}{N_k(s,a) + h}$}
        \label{fig:poly-step}
   \end{minipage}
\end{figure}

\paragraph{Adaptive Stepsize \boldmath{$\alpha_k(s,a) = [\alpha (k + h)] / (N_k(s,a) + h)$}.}
With this choice of stepsize, we study the convergence of $Q_k$ to $Q^*$ under the span seminorm for different values of $\alpha$. In particular, we choose $\alpha \in \{10^{-4},\, 5\times 10^{-5},\, 10^{-5}\}$. For each value of $\alpha$, we run Algorithm~\ref{alg:QL-set} for $400000$ steps and average the results over $500$ trajectories. Finally, we plot the epoch-wise errors, where each epoch consists of $5000$ iterations, as shown in Figure~\ref{fig:const-step}.
It can be observed from Figure~\ref{fig:const-step} that, for each $\alpha$, the error decreases as the number of iterations increases, but only to a bounded region. This behavior is consistent with our discussion in Section~\ref{subsec:adaptive_stepsizes}. In addition, the plots clearly illustrate the trade-off between bias and variance. A larger $\alpha$ leads to larger steps, resulting in a faster decay of the initial bias but at the cost of higher variance. In contrast, a smaller $\alpha$ produces a slower decay of the initial bias due to the smaller steps, but achieves a lower variance.

\paragraph{Adaptive Stepsize \boldmath{$\alpha_k(s,a) = [\alpha (k + h)^{1 - z}] / (N_k(s,a) + h)$}. }
For this stepsize schedule, we illustrate the influence of different values of $z$ on the convergence of the error. We choose $z \in \{0.4, 0.5, 0.6\}$ while keeping $\alpha$ fixed ($\alpha = 0.01$). For each $z$, the number of trajectories, the number of iterations per trajectory, and the epoch length are kept the same as in the previous stepsize setting. Figure~\ref{fig:poly-step} shows that the expected error measured in the span seminorm gradually decays to zero for every $z$, in contrast to the behavior observed in Figure~\ref{fig:const-step}. Moreover, the bias–variance trade-off is again evident. A smaller $z$ reduces the initial bias more quickly because it yields larger steps, but the resulting variance takes longer to decay. Conversely, a larger $z$ requires more iterations to diminish the initial bias, but exhibits a faster decay of the variance.

\end{document}